%% file: vdr_survey.tex
\documentclass[10pt]{article} 
\usepackage[preprint]{tmlr}
\input{math_commands.tex}

\input{packages}
\usepackage{hyperref}
\usepackage{microtype}
\usepackage{graphicx}
\usepackage[utf8]{inputenc} 
\usepackage{booktabs}       
\usepackage{needspace}      
\usepackage{nicefrac}       
\usepackage{microtype}      
\usepackage{xcolor}
\definecolor{bluecite}{HTML}{0071BC}
\usepackage[toc]{appendix}
\usepackage{etoc}
\usepackage{minitoc}

\title{Unlocking Multimodal Document Intelligence: \\From Current Triumphs to Future Frontiers of Visual\\Document Retrieval in the Era of Large Language Models}

\author{\name Yibo Yan \email yanyibo70@gmail.com \\
      \addr Hong Kong University of Science and Technology (Guangzhou)\\
      Alibaba Cloud Computing\\
      Hong Kong University of Science and Technology
      \AND
      \name Jiahao Huo \\
      \addr Hong Kong University of Science and Technology (Guangzhou)\\
      Alibaba Cloud Computing\\
      University of Illinois Chicago
      \AND
      \name Guanbo Feng \\
      \addr Hong Kong University of Science and Technology (Guangzhou)
      \AND
      \name Mingdong Ou\thanks{Project Lead.} \\
      \addr Alibaba Cloud Computing
      \AND
      \name Yi Cao \\
      \addr Alibaba Cloud Computing
      \AND
      \name Xin Zou \\
      \addr Hong Kong University of Science and Technology (Guangzhou)\\
      Hong Kong University of Science and Technology
      \AND
      \name Shuliang Liu \\
      \addr Hong Kong University of Science and Technology (Guangzhou)\\
      Hong Kong University of Science and Technology
      \AND
      \name Yuanhuiyi Lyu \\
      \addr Hong Kong University of Science and Technology (Guangzhou)\\
      Hong Kong University of Science and Technology
      \AND
      \name Yu Huang \\
      \addr Hong Kong University of Science and Technology (Guangzhou)\\
      Alibaba Cloud Computing
      \AND
      \name Jungang Li \\
      \addr Hong Kong University of Science and Technology (Guangzhou)
      \AND
      \name Kening Zheng \\
      \addr University of Illinois Chicago
      \AND
      \name Xu Zheng \\
      \addr Hong Kong University of Science and Technology (Guangzhou)\\
      Hong Kong University of Science and Technology
      \AND
      \name Philip S. Yu \\
      \addr University of Illinois Chicago
      \AND
      \name James Kwok \\
      \addr Hong Kong University of Science and Technology
      \AND
      \name Xuming Hu\thanks{Corresponding author.} \email xuminghu@hkust-gz.edu.cn \\
      \addr Hong Kong University of Science and Technology (Guangzhou)\\
      Hong Kong University of Science and Technology}

\hypersetup{
  pdftitle={Unlocking Multimodal Document Intelligence: From Current Triumphs to Future Frontiers of Visual Document Retrieval in the Era of Large Language Models},
  pdfauthor={Yibo Yan, Jiahao Huo, Guanbo Feng, Mingdong Ou, Yi Cao, Xin Zou, Shuliang Liu, Yuanhuiyi Lyu, Yu Huang, Jungang Li, Kening Zheng, Xu Zheng, Philip S. Yu, James Kwok, Xuming Hu}
}

\def\month{September}
\def\year{2026}
\def\openreview{\url{https://openreview.net/forum?id=XXXX}} 

\begin{document}

\maketitle

\begin{abstract}
With the rapid proliferation of multimodal information, Visual Document Retrieval (VDR) has emerged as a critical frontier in bridging the gap between unstructured visually rich data and precise information acquisition. Unlike traditional natural image retrieval, visual documents exhibit unique characteristics defined by dense textual content, intricate layouts, and fine-grained semantic dependencies. This paper surveys the VDR landscape \textbf{as a retrieval problem in its own right}, rather than as the front end of a generation pipeline, and does so specifically through the lens of the Multimodal Large Language Model (MLLM) era. We begin by examining the benchmark landscape, including the recent turn toward reasoning-intensive evaluation, and then dive into the methodological evolution along two orthogonal axes: \textit{what a retriever is}, spanning multimodal \textit{embedding models} and \textit{reranker models}, and \textit{how it is deployed}, from single-stage retrieval through \textit{Retrieval-Augmented Generation} (RAG) to \textit{Agentic systems}. Cutting across both, we analyse the representation--efficiency trade-off that late interaction imposes. We ground these categories in leaderboard evidence on where the empirical frontier actually lies, and close by identifying persistent challenges and outlining promising future directions for multimodal document intelligence.
\end{abstract}

\input{sections/1-introduction}

\input{sections/2-benchmark}
\input{sections/3-methodology}
\input{sections/4-challenge}

\input{sections/5-conclusion}




\clearpage
\bibliography{vdr_survey}
\bibliographystyle{tmlr}

\appendix

\input{sections/Appendix}
\end{document}

%% file: math_commands.tex
\usepackage{amsmath,amsfonts,bm}

\def\eqref#1{equation~\ref{#1}}

\def\1{\bm{1}}

\DeclareMathAlphabet{\mathsfit}{\encodingdefault}{\sfdefault}{m}{sl}
\SetMathAlphabet{\mathsfit}{bold}{\encodingdefault}{\sfdefault}{bx}{n}



%% file: packages.tex
\usepackage{graphicx}
\usepackage{subcaption}
\usepackage{multirow}
\usepackage{tikz}
\newcommand\encircle[2][]{%
    \tikz[baseline=(char.base)]
    \node[
        fill=blue!20,
        inner sep=2pt, 
        rectangle, 
        rounded corners=1.5mm, 
        #1, 
        name=char 
    ] {#2};%
}
\usepackage[dvipsnames]{xcolor} 
\newcommand{\breakslash}{/\hspace{0pt}}

\definecolor{lightergray}{RGB}{230,230,230}
\definecolor{DarkRed}{RGB}{130,25,0}
\definecolor{PurpleRed}{RGB}{204,0,102}
\definecolor{DarkGreen}{RGB}{30,130,30}
\definecolor{DarkBlue}{RGB}{0,0,250}
\definecolor{DarkYellow}{RGB}{255,128,0}
\definecolor{light-gray}{gray}{0.95}
\definecolor{lightgreen}{RGB}{231,255,219}
\definecolor{lightred}{RGB}{252,231,234}
\definecolor{lightyellow}{RGB}{250,253,191}
\definecolor{lightpurple}{RGB}{229,204,255}
\definecolor{lightblue}{RGB}{229,246,254}

\usepackage{amsmath}

\usepackage{paralist}

\usepackage{microtype}
\usepackage{booktabs} %

\definecolor{darkgreen}{rgb}{0.0, 0.5, 0.0} 

\newcommand{\comt}[1]{#1}

\renewcommand{\comt}[1]{}

\usepackage[dvipsnames,table]{xcolor}

\usepackage{tabularx}
\usepackage{multirow} 
\usepackage{multicol}
\definecolor{myblue}{RGB}{235,235,250}

\usepackage{xspace}
\usepackage{pifont}
\usepackage{arydshln}
\usepackage{enumitem}
\usepackage{wrapfig}
\usepackage{array}
\usepackage{epigraph}

\definecolor{lightpink}{RGB}{204, 231, 207} 
\definecolor{lightblue}{RGB}{210, 220, 250} 
\definecolor{lightgray}{RGB}{237, 237, 237} 

\definecolor{superlightred}{rgb}{0.99, 0.92, 0.92}
\definecolor{darkgreen}{RGB}{50,100,0}
\definecolor{darkred}{RGB}{200, 0, 0}
\newcommand{\halfcheck}{\textcolor{gray!65}{\ding{52}}}

\usepackage{forest}
\usetikzlibrary{fit,positioning}
\definecolor{taxM}{RGB}{0,0,250}      
\definecolor{taxD}{RGB}{30,130,30}    
\definecolor{taxT}{RGB}{130,25,0}     
\definecolor{taxE}{RGB}{255,128,0}    

\usepackage{makecell}
\usepackage{colortbl}

\usepackage{hyperref}
\usepackage[capitalize]{cleveref}
\usepackage[most]{tcolorbox}
\usepackage{wrapfig}
\usepackage{graphicx,xcolor,float}
\usepackage{subcaption}
\usepackage{threeparttable}
\usepackage{algorithm}
\usepackage{colortbl}
\usepackage{color}
\usepackage{multirow}
\usepackage{tabularx}
\usepackage{float}
\usepackage{graphicx}
\usepackage{booktabs}
\usepackage{arydshln}
\usepackage{enumitem}
\usepackage{wrapfig}
\usepackage{caption}
\usepackage{graphicx}
\usepackage{makecell}
\usepackage{float} 
\usepackage{makecell}
\usepackage{tabularx}
\usepackage{twemojis}
\usepackage{amssymb,mathrsfs,amsmath}
\usepackage[export]{adjustbox}
\usepackage{xcolor}
\usepackage{xspace}
\usepackage{shadowtext}

\hypersetup{
    colorlinks=true,
    linkcolor=red,
    citecolor=cyan,
    filecolor=magenta,      
    urlcolor=magenta,
    }

\definecolor{bittersweet}{rgb}{1.0, 0.44, 0.37}
\definecolor{mygreen}{rgb}{0.29, 0.7, 0.48}
\definecolor{my_green}{RGB}{51,102,0}
\definecolor{my_yellow}{RGB}{255,165,0}
\definecolor{my_red}{RGB}{204, 0, 0}
\definecolor{my_orange}{RGB}{232, 132, 23}

\definecolor{mygray}{gray}{0.4}
\hypersetup{
    colorlinks=true,
    linkcolor=red,
    citecolor=cyan,
    filecolor=magenta,      
    urlcolor=magenta,
    }
\usepackage{xcolor} 
\usepackage{xcolor}

\usepackage[font=small,labelfont=bf]{caption}  
\usepackage{makecell}
\usepackage{tabulary}
\definecolor{ada_green}{rgb}{0,205,205}
\definecolor{glt_red}{rgb}{109,205,255}

\definecolor{backred}{RGB}{255, 190, 190}
\definecolor{backblue}{RGB}{210, 230, 250}
\definecolor{backgrey}{RGB}{220, 220, 220}

\definecolor{shadecolor}{RGB}{237,237,237}



%% file: sections/1-introduction.tex
\section{Introduction}
\label{sec:introduction}
\begin{wrapfigure}{r}{0.5\textwidth}
\centering
\includegraphics[width=0.50\textwidth]{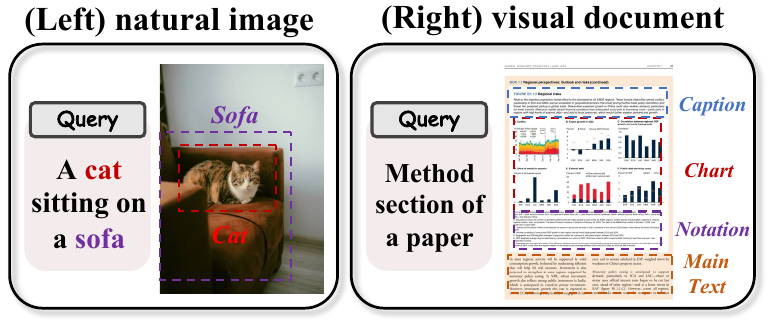}
\caption{Natural-image retrieval (\textit{left}) versus visual-document retrieval (\textit{right}), the focus of this survey.}
\label{fig:task_img_comparison}
\end{wrapfigure}
\begin{figure}[ht!]
  \centering 

  \begin{subfigure}[b]{0.43\textwidth}
    \centering
    \includegraphics[width=\textwidth]{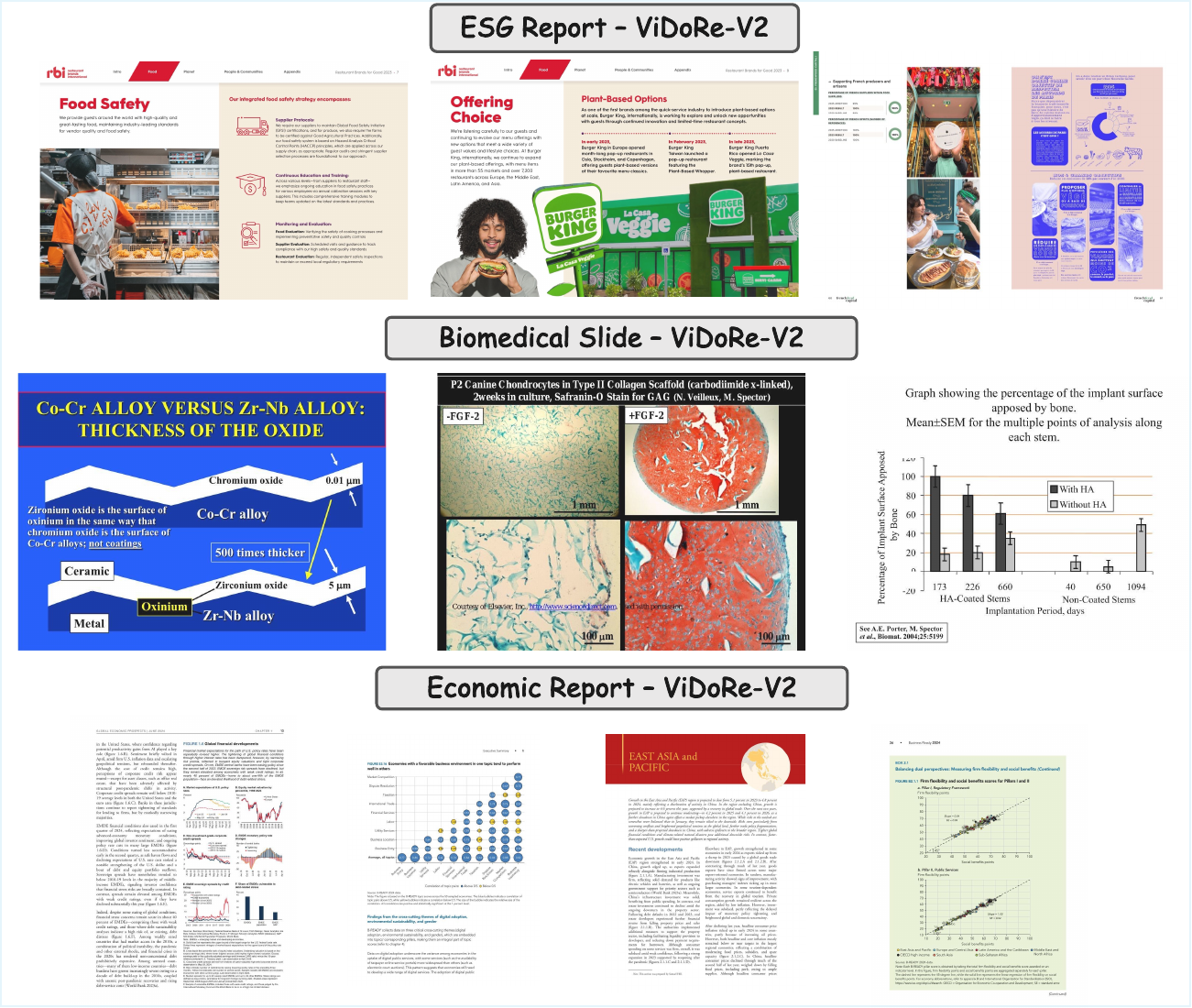}
    \caption{Illustrative examples of ViDoRe-V2 \citep{mace2025vidorev2}.}
    \label{fig:sub_vidorev2}
  \end{subfigure}
  \hfill 
  \begin{subfigure}[b]{0.55\textwidth}
    \centering
    \includegraphics[width=\textwidth]{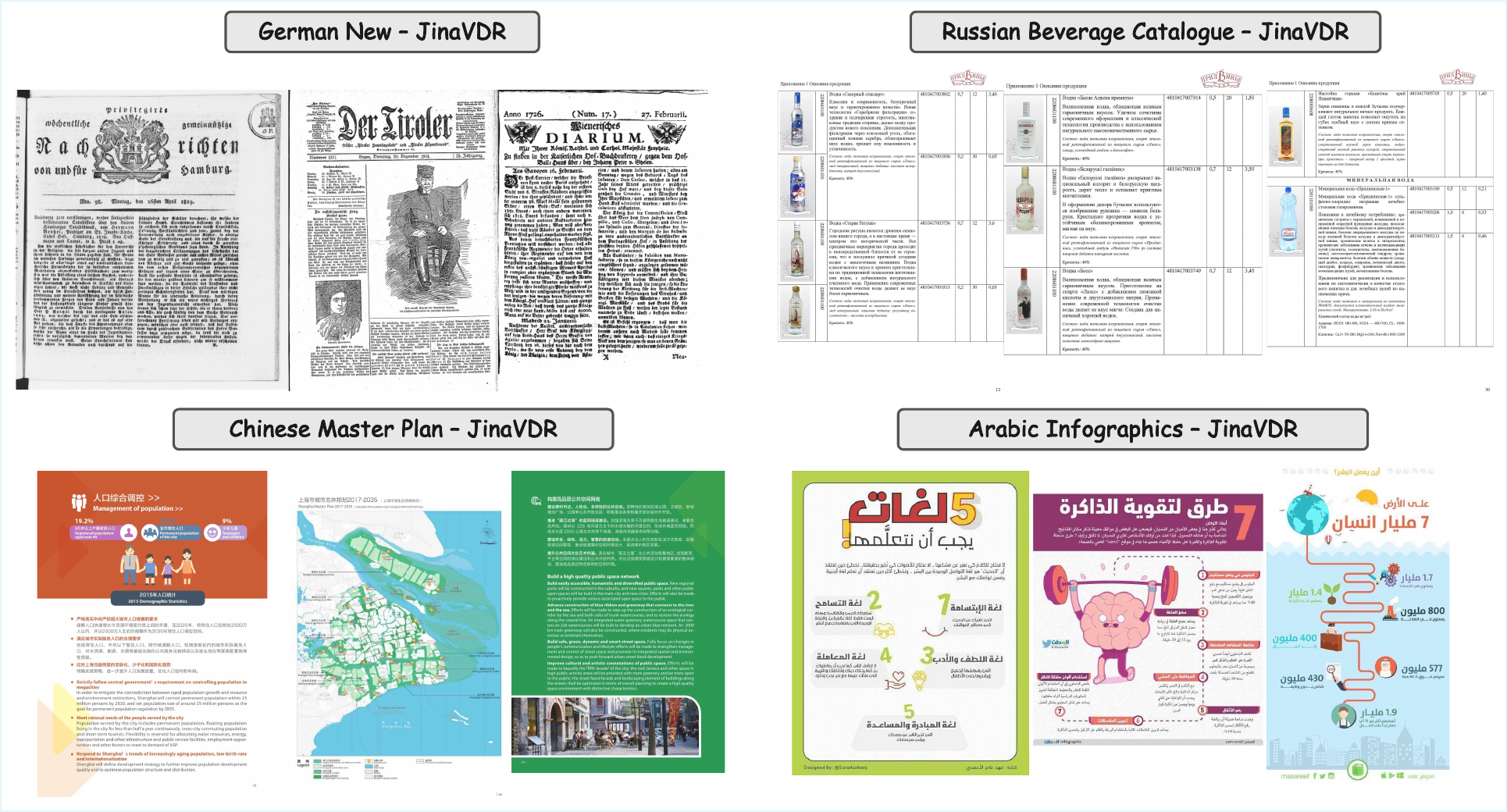}
    \caption{Illustrative examples of JinaVDR \citep{gunther2025jina}.}
    \label{fig:sub_jinavdr}
  \end{subfigure}

  \caption{Illustrative examples from two representative datasets, ViDoRe-V2 (a) and JinaVDR (b).}
  \label{fig:examples_combined}
  \vspace{-3mm} 
\end{figure}

Multimodal retrieval, the task of retrieving relevant multimodal information from a large-scale collection using queries that span multiple modalities like text and vision, has become a cornerstone of modern information retrieval \citep{mei2025survey,zheng2025retrieval}. 
Historically, research in this domain has predominantly focused on natural image retrieval, targeting datasets of photographs and web images where the primary goal is to match objects, scenes, or holistic visual concepts \citep{wu2024retrieval,arslan2024survey}. 
However, both academia and industry have begun to turn their attention to a distinct yet ubiquitous data type: \textbf{visual documents}\footnote{They are also commonly referred to as \textit{visually rich documents}, \textit{document images}, \textit{etc}. We use ``visual documents'' as a unifying term. See the illustrative examples from representative VDR datasets in Figure \ref{fig:examples_combined}.}. 
These documents, ranging from scanned PDFs and business reports to invoices and academic papers, are characterized by a dense interplay of textual content, complex layouts, and graphical elements \citep{tang2023unifying,li2024enhancing}. 
This shift necessitates an exploration of retrieval techniques tailored to the unique structure and information density of visual documents, pushing the boundaries of what multimodal systems can achieve.

The pivot towards \textbf{Visual Document Retrieval} (VDR) is driven by three fundamental differences that distinguish visual documents from natural images, as illustrated in Figure \ref{fig:task_img_comparison}. 
\ding{182} \textbf{Information modality and density}: unlike natural images which convey semantic meaning through holistic scenes, visual documents are hybrid entities where meaning is co-determined by rich textual information and a structured spatial layout. 
The information is dense, hierarchical, and multi-modal by nature. 
\ding{183} \textbf{Semantic granularity}: retrieval in natural images often targets high-level concepts (\textit{e.g.,} "a cat sitting on a sofa"), whereas VDR demands a much finer-grained understanding. 
Users may query for specific facts embedded within a table, a particular sentence in a paragraph, or information contingent on its document-level position (\textit{e.g.,} "the methodology section of a paper"). 
\ding{184} \textbf{User intent and task complexity}: VDR is geared towards precise information-seeking, question answering, and evidence-based reasoning, rather than conceptual or aesthetic matching. 

\begin{wraptable}{r}{0.5\textwidth}
    \centering
    \small
    \caption{Coverage of prior surveys relative to ours. Instead of tagging each work with a single \underline{R}etrieval/\underline{U}nderstanding label, a dichotomy that misrepresents retrieval-oriented surveys that happen to be framed around RAG, we score four \textit{checkable} axes of the retrieval stack: dedicated treatment of \textbf{Emb}edding models, of \textbf{Rer}ankers as a first-class stage, of index \textbf{Eff}iciency (pruning\breakslash merging\breakslash compression), and of \textbf{Rea}soning-intensive retrieval. \ding{52}~=~dedicated section or table; \halfcheck~=~discussed without dedicated treatment; blank~=~absent. Each judgement is reproducible from the cited work's table of contents.}
    \label{tab:survey_comparison}
    \resizebox{\linewidth}{!}{
    \begin{tabular}{cccccccc}
    \toprule
    \multirow{2}{*}{\textbf{Survey}} & \multirow{2}{*}{\textbf{Venue}} & \multirow{2}{*}{\textbf{Scope}} & \multicolumn{4}{c}{\textbf{Retrieval-Stack Coverage}} & \multirow{2}{*}{\textbf{Agent}} \\

    \cmidrule(lr){4-7}

    & & & \textbf{Emb} & \textbf{Rer} & \textbf{Eff} & \textbf{Rea} & \\
    \midrule

    \cite{alaei2016brief} & IJCNN'16 & IR4Doc &  &  &  &  &  \\

    \cite{zhou2017recent} & arxiv'17 & IR4Img &  &  &  &  &  \\

    \cite{subramani2020survey} & NeurIPS Workshop'20 & DL4Doc &  &  &  &  &  \\

    \cite{cui2021document} & ICDAR'21 & IR4Doc &  &  &  &  &  \\

    \cite{hameed2021content} & CE'21 & IR4Img &  &  &  &  &  \\

    \cite{sassioui2023visually} & WINCOM'23 & IR4Doc &  &  &  &  &  \\

    \cite{zhao2023retrieving} & EMNLP Findings'23 & LLM4Img & \halfcheck &  &  &  &  \\

    \cite{ding2024deep} & arxiv'24 & DL4Doc & \halfcheck &  &  &  &  \\

    \cite{zhang2025composed} & arxiv'25 & LLM4Img & \ding{52} & \halfcheck &  & \halfcheck &  \\

    \cite{ding2025survey} & IJCAI Tutorial'25 & LLM4Doc & \halfcheck &  &  &  &  \\

    \cite{zhang2025roles} & AACL-IJCNLP'25 & LLM4Doc & \ding{52} & \halfcheck & \halfcheck &  &  \\

    \cite{abootorabi2025ask} & ACL Findings'25 & LLM4MM & \ding{52} & \halfcheck & \halfcheck & \halfcheck & \ding{52} \\

    \cite{huang2024pixels} & IEEE TKDE'25 & LLM4Doc & \halfcheck &  &  &  & \ding{52} \\

    \cite{gao2025scaling} & ACL'26 & LLM4Doc & \ding{52} & \halfcheck & \ding{52} & \halfcheck & \ding{52} \\

    \cite{10.1145/3768156} & ACM TOIS'26 & LLM4Doc & \halfcheck &  &  & \halfcheck & \ding{52} \\

    \cite{rombach2025deep} & ACM Computing Survey'26 & DL4Doc & \halfcheck &  &  &  &  \\

    \midrule
    \textbf{Ours} & - & \textbf{LLM4Doc} & \ding{52} & \ding{52} & \ding{52} & \ding{52} & \ding{52} \\
    \bottomrule
    \end{tabular}}
\end{wraptable}
Furthermore, as the general capabilities of Multimodal Large Language Models (MLLMs) advance \citep{song2025bridge,yan2025position,yan2025survey}, the VDR field is increasingly focusing on their integration. This includes the development of MLLM-based embedding and reranker models to enhance semantic matching \citep{tao2024llms,zhang2024word,wang2024cross}. Beyond that, there is active exploration into leveraging these models within more sophisticated frameworks like Retrieval-Augmented Generation (RAG) pipelines \citep{gao2023retrieval,cheng2025survey,gan2025retrieval} and Agentic systems \citep{singh2025agentic} to tackle complex document-based settings.

\textbf{Scope.} Prior reviews approach this material from adjacent vantage points (Table~\ref{tab:survey_comparison}). Earlier work concentrated on classical document information retrieval~\citep{alaei2016brief} and deep learning for document understanding~\citep{subramani2020survey, sassioui2023visually, ding2024deep}, or on retrieval for natural images~\citep{zhou2017recent, hameed2021content}. Surveys written after the rise of MLLMs largely retain a generation-centric or understanding-centric framing~\citep{huang2024pixels, rombach2025deep, ding2025survey, 10.1145/3768156} or target natural-image retrieval~\citep{zhao2023retrieving, zhang2025composed}.

Three recent works are close enough that we state the difference explicitly rather than rely on a checkmark. \citet{abootorabi2025ask} survey multimodal RAG broadly; retrieval appears there as one component of a generation pipeline, and visually rich documents are one modality among many, so the document-specific retrieval stack (layout-aware multi-vector representations, index compression, document rerankers) is not the organising concern. \citet{gao2025scaling} is genuinely retrieval-oriented and covers embedding models, the ColPali family, and compression; its framing runs from multimodal RAG toward document understanding, and we differ in emphasis rather than in kind, treating retrieval as the object of study instead of the front end of a generator. Closest of all is \citet{zhang2025roles}, a VDR retrieval survey with substantially overlapping scope; relative to it we add a dedicated treatment of rerankers as a first-class stage with its ranking-paradigm taxonomy, the reasoning-intensive turn in benchmarks, an empirical reading of the current leaderboard frontier, and the efficiency literature that has appeared since.

Our contribution is therefore not to be first to mention VDR, but to organise it as a retrieval problem in its own right: we \textbf{treat embedding, reranking, and agentic deployment along explicit axes, and pair each with the benchmark and efficiency evidence needed to compare methods}\footnote{This version covers literature up to \textbf{25 August, 2026}, with updates scheduled every two months for the fast-growing research community.}.

\textbf{Structure.} We begin from a \textbf{benchmark perspective}, systematically organizing the field by examining task formulations, basic settings, and dataset characteristics such as multilingual support and the growing emphasis on reasoning-intensive queries. We then turn to a \textbf{methodology-centric analysis}.

Here we make our organising principle explicit, because a flat list of ``paradigms'' would conflate two questions that vary independently. We use \textbf{two orthogonal axes and one cross-cutting dimension}. The first axis asks \textit{what a retriever is}, that is, its representation and scoring mechanism, and runs from single-vector bi-encoders through multi-vector late interaction ($\vartriangleright$ \Cref{sec:embedding}) to cross-encoder joint scoring ($\vartriangleright$ \Cref{sec:reranker}) and, most recently, generative formulations. The second axis asks \textit{how a retriever is deployed}: single-stage retrieval, retrieve-then-rerank cascades, static RAG pipelines, or iterative agentic loops ($\vartriangleright$ \Cref{sec:rag_pipeline}). These axes are genuinely independent: a multi-vector embedder can be deployed single-stage or inside an agent, and an agentic system can be assembled from any underlying retriever. We note explicitly that the second axis is \textit{downstream} of the first, which is why \Cref{sec:rag_pipeline} builds on rather than parallels the two preceding sections. Cutting across both is the \textbf{resource--quality trade-off}: the representation choice that wins on accuracy also determines index size and query-time cost, a tension we quantify in \Cref{sec:embedding} and pursue in \Cref{app:challenge_performance_efficiency_dilemma}.

Finally, we conclude by discussing the challenges and outlining \textbf{future frontiers}, aiming to provide valuable insights and inspire subsequent research in the multimodal document intelligence community.

%% file: sections/2-benchmark.tex
\section{Benchmark Perspective}
\label{sec:benchmark}

This section provides a systematic review of the VDR evaluation landscape. We first establish a formal mathematical definition  ($\vartriangleright$ \Cref{sec:benchmark_formulation}), and then analyze the current trends in terms of data scale and metrics ($\vartriangleright$ \Cref{sec:benchmark_basic_setting}), followed by a discussion on the emerging frontiers of multilingual support ($\vartriangleright$ \Cref{sec:benchmark_multilingual_support}) and reasoning-intensive retrieval ($\vartriangleright$ \Cref{sec:benchmark_reasoning_setting}), which reflect the shift from keyword matching to complex document intelligence.

\subsection{Formulation}
\label{sec:benchmark_formulation}

VDR aims to identify the most relevant document images from a large-scale corpus based on a given query. Formally, let $\mathcal{C} = \{d_n\}_{n=1}^{|\mathcal{C}|}$ be a corpus of document pages, where each $d_n \in \mathcal{I}$ is a visually rich document image. Given a query $q$, the task is to produce a ranked list of documents from $\mathcal{C}$ such that the top-ranked items maximize a relevance score $s(q, d)$.

\textbf{Input and Modalities.} In the standard VDR setting, the query is typically a natural language string $q \in \mathcal{T}$ (text-to-image retrieval). However, the generalized VDR setting extends the query space to multimodal inputs, including images or interleaved text-and-image sequences $q \in \{\mathcal{T} \cup \mathcal{I}\}^*$. Unlike traditional OCR-based retrieval that treats documents as plain text, VDR models process the document $d$ directly in the visual domain, often representing it as a set of patch-level embeddings to preserve layout and graphical information.

\textbf{Mathematical Representation.} Following the late-interaction paradigm pioneered by ColPali~\citep{faysse2024colpali}, let $\mathbf{U}(q) = \{\mathbf{u}_i\}_{i=1}^{n_q}$ denote the query-side embeddings (token-level for the standard text-query setting) and $\mathbf{V}(d) = \{\mathbf{v}_j\}_{j=1}^{n_d}$ the patch-level document embeddings, with $\mathbf{u}_i, \mathbf{v}_j \in \mathbb{R}^{h}$. Relevance can then be computed with a late-interaction operation such as \texttt{MaxSim}: $s(q, d) = \sum_{i=1}^{n_q} \max_{j=1}^{n_d} \mathbf{u}_i^\top \mathbf{v}_j.$

\textbf{Evaluation Metrics.} This section provides detailed mathematical formulations for the primary retrieval metrics used in the benchmarks discussed in this survey. For a given set of test queries $\mathcal{Q}$, the goal of a retrieval system is to return a ranked list of documents from a corpus $\mathcal{C}$ for each query $q \in \mathcal{Q}$.

\paragraph{\ding{182} Recall@k}

Recall at k (Recall@k) measures the fraction of relevant documents that are successfully retrieved within the top-k results. It is a metric of coverage, evaluating how well the system is able to find all ground-truth documents. For a given query $q$, let $\mathcal{D}_q^+$ be the set of ground-truth relevant documents and $\pi_k(q)$ be the ranked list of the top-$k$ retrieved documents. The overall Recall@k is the average score across all queries in $\mathcal{Q}$.

Recall@k is calculated as:
\begin{equation}
    \text{Recall}@k = \frac{1}{|\mathcal{Q}|} \sum_{q \in \mathcal{Q}} \frac{|\mathcal{D}_q^+ \cap \pi_k(q)|}{|\mathcal{D}_q^+|}
\end{equation}
where $|\mathcal{Q}|$ is the total number of queries; $|\mathcal{D}_q^+|$ is the total number of relevant documents for query $q$; and $|\mathcal{D}_q^+ \cap \pi_k(q)|$ is the number of relevant documents found in the top-$k$ retrieved list.

\paragraph{\ding{183} Mean Reciprocal Rank (MRR)}

Mean Reciprocal Rank (MRR) evaluates the ranking of the \textit{first} correct document retrieved. It is particularly useful for tasks where finding a single relevant item quickly is the primary goal. For each query $q$, the reciprocal rank is the inverse of the rank of the first relevant document. If no relevant document is retrieved, the reciprocal rank is 0.

MRR is calculated as:
\begin{equation}
    \text{MRR} = \frac{1}{|\mathcal{Q}|} \sum_{q \in \mathcal{Q}} \frac{1}{\text{rank}_q}
\end{equation}
where $\text{rank}_q$ is the position of the first relevant document for query $q$, with $\text{rank}_q=\infty$ when no relevant document is retrieved.

\paragraph{\ding{184} Mean Average Precision (mAP)}
Mean Average Precision (mAP) provides a comprehensive evaluation of a ranked list by considering both precision and recall. It rewards retrieving relevant documents at higher ranks. The Average Precision (AP) for a single query is first calculated by averaging the precision at each relevant document's position. mAP is then the mean of these AP scores over all queries.

The Average Precision for a single query $q$ is:
\begin{equation}
    \text{AP}_q = \frac{1}{|\mathcal{D}_q^+|} \sum_{r=1}^{|\mathcal{C}|} \operatorname{Prec}_q(r)\,\operatorname{rel}_q(r)
\end{equation}
And mAP is the mean of all AP scores:
\begin{equation}
    \text{mAP} = \frac{1}{|\mathcal{Q}|} \sum_{q \in \mathcal{Q}} \text{AP}_q
\end{equation}
where $|\mathcal{C}|$ is the total number of documents in the corpus; $\operatorname{Prec}_q(r)$ is the precision for query $q$ at rank $r$; and $\operatorname{rel}_q(r)$ is 1 if the document at rank $r$ is relevant to $q$, and 0 otherwise.

\paragraph{\ding{185} Normalized Discounted Cumulative Gain (nDCG@k)}
Normalized Discounted Cumulative Gain (nDCG@k) is a measure of ranking quality that accounts for the graded relevance of documents. It assigns higher scores to more relevant documents placed at top ranks, with a logarithmic discount for documents at lower ranks. The score is normalized by the ideal ranking to fall between 0 and 1.

The DCG@k is first calculated as:
\begin{equation}
    \text{DCG}@k(q) = \sum_{r=1}^{k} \frac{2^{g_{q,r}}-1}{\log_2(r+1)}
\end{equation}
The nDCG@k is then:
\begin{equation}
    \text{nDCG}@k = \frac{1}{|\mathcal{Q}|}\sum_{q\in\mathcal{Q}}\frac{\text{DCG}@k(q)}{\text{IDCG}@k(q)}
\end{equation}
where $k$ is the number of results being evaluated; $g_{q,r}$ is the graded relevance of the document at rank $r$ for query $q$; and $\text{IDCG}@k(q)$ is the DCG of the ideal ranking for $q$ up to rank $k$.

\paragraph{\ding{186} Hit Rate (HR@k)}
As used in benchmarks like \texttt{Double-Bench}~\citep{shen2025we}, Hit Rate at k (HR@k) is a binary success metric that evaluates whether \textit{at least one} relevant document is found within the top-k retrieved results. It is particularly useful for assessing whether the retrieval system can find any correct evidence, which is a prerequisite for downstream tasks. For multi-hop queries, a "hit" is only registered if evidence for \textit{all} hops is found within the top-k results.

For a single-hop query $q$, the HR@k is:
\begin{equation}
    \text{HR}@k(q) = \mathbb{I}(\pi_k(q) \cap \mathcal{D}_q^+ \neq \emptyset)
\end{equation}
The overall HR@k is the average across all queries:
\begin{equation}
    \text{HR}@k = \frac{1}{|\mathcal{Q}|} \sum_{q \in \mathcal{Q}} \text{HR}@k(q)
\end{equation}
where $\mathbb{I}(\cdot)$ is the indicator function, which is 1 if the condition is true and 0 otherwise; $\pi_k(q)$ is the top-$k$ retrieved list and $\mathcal{D}_q^+$ is the set of ground-truth relevant documents.

\paragraph{\ding{187} Averaged Normalized Longest Common Subsequence (ANLCS)}
Used in \texttt{VisDoMBench}~\citep{suri2025visdom} to evaluate evidence extraction, Averaged Normalized Longest Common Subsequence (ANLCS) measures the textual similarity between retrieved content and ground-truth evidence. Instead of evaluating ranking, it assesses the quality of the retrieved content itself. First, the Normalized Longest Common Subsequence (NLCS) is calculated for a pair of text strings, measuring their shared content normalized by their lengths.

The NLCS between a ground-truth evidence string $x_{\mathrm{gt}}$ and a retrieved chunk string $x_{\mathrm{ret}}$ is:
\begin{equation}
    \text{NLCS}(x_{\mathrm{gt}}, x_{\mathrm{ret}}) = \frac{2\,|\text{LCS}(x_{\mathrm{gt}}, x_{\mathrm{ret}})|}{|x_{\mathrm{gt}}| + |x_{\mathrm{ret}}|}
\end{equation}
The ANLCS score for a query is then computed by finding the best-matching retrieved chunk for each ground-truth evidence chunk and averaging these scores. The final metric is the average over all queries.
\begin{equation}
\begin{split}
    \text{ANLCS}@k = \frac{1}{|\mathcal{Q}|} \sum_{q \in \mathcal{Q}} \left( \frac{1}{|\mathcal{G}_q|} \sum_{g \in \mathcal{G}_q}\max_{x \in \mathcal{X}_k(q)} \text{NLCS}(g, x) \right)
\end{split}
\end{equation}
where $\text{LCS}(x_{\mathrm{gt}}, x_{\mathrm{ret}})$ is the Longest Common Subsequence between the two strings; $|x|$ denotes the length of string $x$; $\mathcal{G}_q$ is the set of ground-truth evidence texts for query $q$; and $\mathcal{X}_k(q)$ is the set of retrieved text chunks in the top-$k$ results for query $q$.


\subsection{Basic Setting}
\label{sec:benchmark_basic_setting}

\textbf{Escalating Research Momentum.} Over the past two years, VDR has transitioned from a niche task to a central focus in both industry and academia. As shown in Table~\ref{tab:benchmark_comparison}, the majority of specialized VDR benchmarks, such as the ViDoRe series~\citep{faysse2024colpali} and Real-MM-RAG~\citep{wasserman2025real}, emerged in 2024 and 2025. This surge is driven by the realization that (OCR-based) text-only retrieval fails to capture the visual nuances of documents, such as tables, charts, and spatial hierarchies~\citep{zhang2025ocr,most2025lost}.

\textbf{Diverse Dataset Scales.} Current benchmarks exhibit a wide range of data magnitudes~\citep{cao2025toward}. While expert-annotated sets like SeaDoc~\citep{xiao2025scaling} and M4DocBench~\citep{dong2025doc} focus on specialized samples, recent large-scale efforts have pushed boundaries. NL-DIR~\citep{guo2025towards} provides 205k queries, and Jina-VDR~\citep{gunther2025jina} utilizes over 70k documents, reflecting a trend toward scaling both query volume and corpus diversity.

\textbf{Standardized Evaluation.} nDCG and Recall remain the primary metrics for performance measurement. However, as VDR is increasingly integrated into RAG, some benchmarks (\textit{e.g.,} EVisRAG~\citep{sun2025visrag} and MMLongBench-Doc~\citep{ma2024mmlongbench}) incorporate downstream Accuracy and F1 scores to evaluate how retrieval quality directly impacts final multimodal generation.
\vspace{-2mm}
\subsection{Multilingual Support}
\label{sec:benchmark_multilingual_support}

\begin{table*}[!t] 
    \centering
    \small
    \resizebox{\linewidth}{!}{
\begin{tabular}{ll cc lllll l} 
    \toprule
    \multirow{2}{*}{\makecell{\textbf{Category}}} & 
    \multirow{2}{*}{\makecell{\textbf{Benchmark}}} & 
    \multicolumn{2}{c}{\textbf{Publication}} & 
    \multicolumn{5}{c}{\textbf{Dataset}} & 
    \multirow{2}{*}{\makecell{\textbf{Resource}}} \\
    
    \cmidrule(lr){3-4} \cmidrule(lr){5-9}
    
    & & 
    \multicolumn{1}{c}{\textbf{Team}} & 
    \multicolumn{1}{c}{\textbf{Venue}} & 
    \multicolumn{1}{c}{\textbf{\#Query*}} & 
    \multicolumn{1}{c}{\textbf{\#Corpus*}} & 
    \multicolumn{1}{c}{\textbf{Multilingual}} & 
    \multicolumn{1}{c}{\textbf{Reasoning}} & 
    \multicolumn{1}{c}{\textbf{Retrieval Metric}} & 
    \\
    \midrule

    VDR & 
    MRMR \citep{zhang2025mrmr} & 
    NTU & 
    ICLR'26 & 
    1.50k & 
    42.80k & 
    en & 
    \ding{52} & 
    nDCG / HR & 
    \begin{adjustbox}{valign=m}
        \href{https://github.com/rebeccaz4/MRMR}{
            \includegraphics[height=3ex]{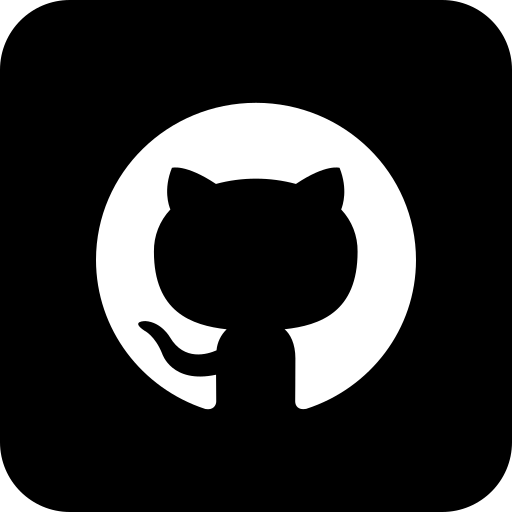}
        }
        \hspace{0.05em}
        \href{https://huggingface.co/MRMRbenchmark}{
            \includegraphics[height=3ex]{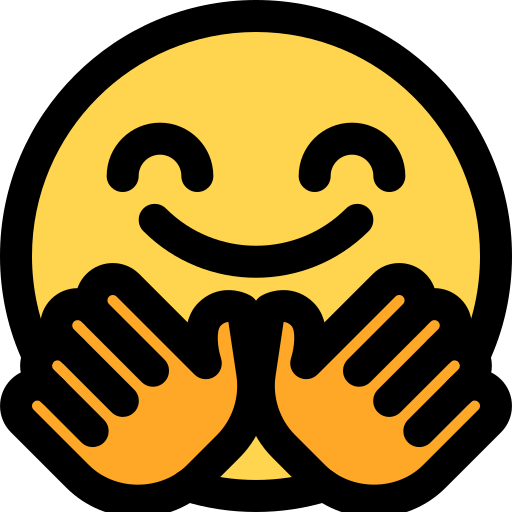}
        }
        \hspace{0.05em}
        \href{https://arxiv.org/pdf/2510.09510}{
            \includegraphics[height=3ex]{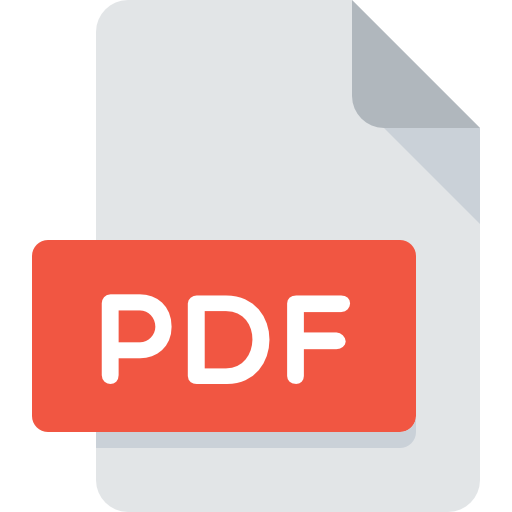}
        }
    \end{adjustbox} \\

    VDR & 
    Invoice Haystack \citep{kanagalingam2026invoice} & 
    Melbourne & 
    ECCV'26 & 
    200 & 
    1.50k & 
    en & 
    - & 
    Recall & 
    \begin{adjustbox}{valign=m}
        \href{https://arxiv.org/pdf/2606.25343}{
            \includegraphics[height=3ex]{icons/icon_pdf.png}
        }
    \end{adjustbox} \\
\bottomrule    
    \end{tabular}
    }
    \caption{Comparison of VDR and general multimodal retrieval benchmarks. In the \textit{Resource} column, we denote the corresponding GitHub codebase, Hugging Face repository, and paper (\textit{e.g.,} arXiv paper, technical report, or blog) as \includegraphics[height=2ex]{icons/icon_github.png} / \includegraphics[height=2ex]{icons/icon_huggingface.png} / \includegraphics[height=2ex]{icons/icon_pdf.png}. * indicates that \#query and \#corpus correspond to the retrieval-related \#evaluation samples and \#candidates, respectively, for general multimodal retrieval benchmarks.} 
    \label{tab:benchmark_comparison}
    \vspace{-6mm}
\end{table*}

\begin{figure*}[t!]
    \centering
    \includegraphics[width=\linewidth]{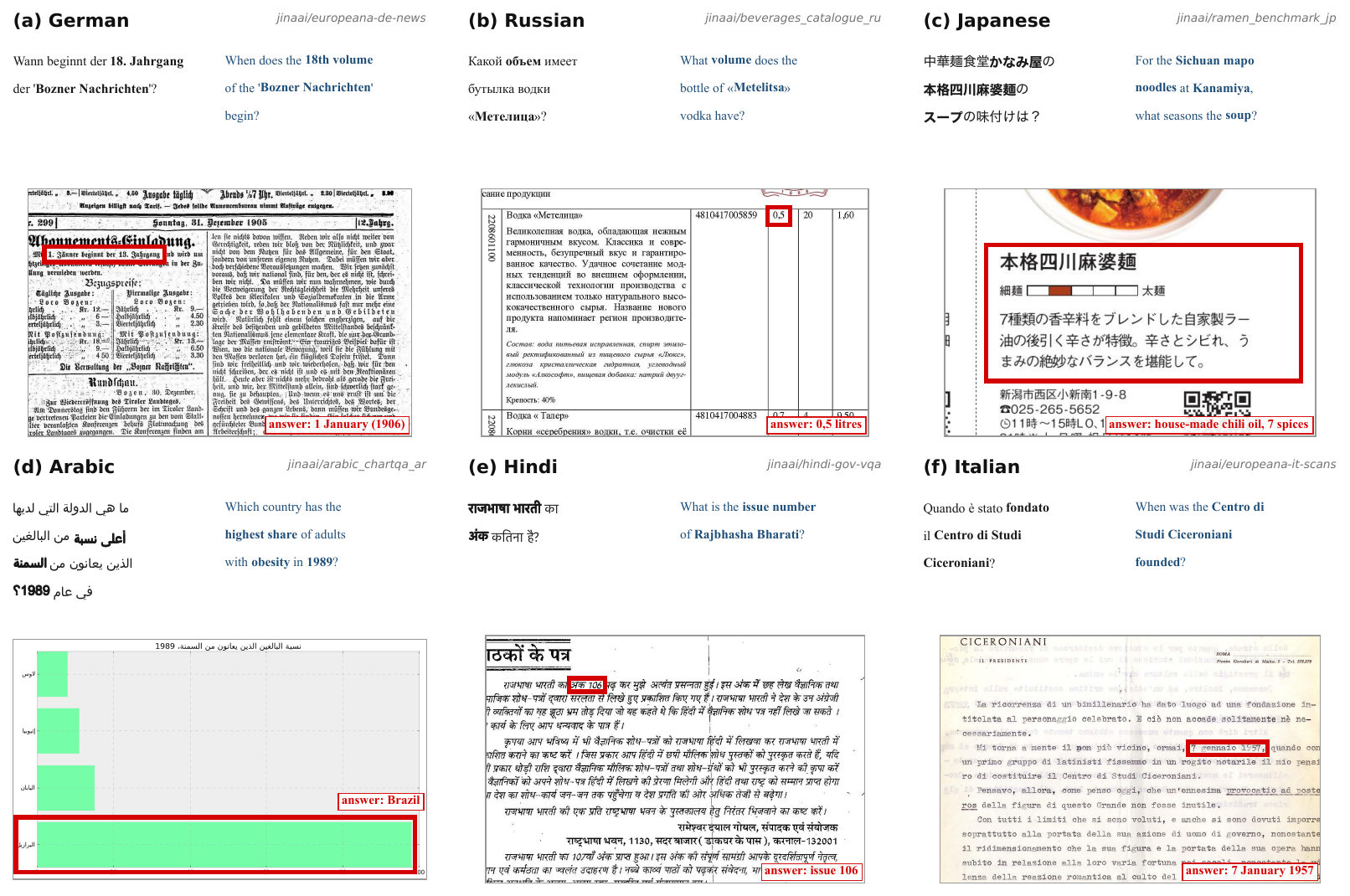}
    \caption{\textbf{What multilingual VDR actually asks of a retriever.} One case from each of six non-English JinaVDR subsets \citep{gunther2025jina}. Each panel pairs the query in its original script (\textit{left}) with an English gloss (\textit{right}) on the same line, and outlines in \textcolor{red}{\textbf{red}} the page region that answers it, annotated with the answer itself. Key entities, dates and quantities, the tokens that must survive cross-script alignment, are set in bold in \textit{both} the original and the gloss, so the two readings can be compared term by term. The panels span four writing systems and both text directions, and the evidence takes a different form in each: a masthead notice (a), a single table cell (b), a shop card (c), a bar in a chart (d), an issue number in running prose (e), and a date inside a letter body (f). A retriever therefore needs script-level reading, layout parsing and numeric grounding at once, which is why the reranking gap discussed in \Cref{sec:reranker} is not merely a matter of adding languages.}
    \label{fig:multilingual_cases}
\end{figure*}

The majority of early VDR benchmarks are predominantly English-centric. Following advances in multilingual text-embedding evaluation~\citep{enevoldsen2025mmteb,zhang2023miracl}, recent VDR work has begun to bridge this linguistic gap. Jina-VDR~\citep{gunther2025jina} and Nayana-IR~\citep{kolavi2025m3dr} represent a significant shift, supporting 20 and 22 languages, respectively. Similarly, MIRACL-VISION~\citep{osmulski2025miracl} introduces a large-scale multilingual corpus covering 18 languages. 
This movement highlights the necessity for VDR models to move beyond English-specific representations and develop cross-lingual semantic alignment capabilities, particularly for visually rich documents. Figure~\ref{fig:multilingual_cases} makes concrete what that demand involves. Across the six subsets the queries are not translations of one another but are grounded in locally specific artefacts, a Habsburg-era subscription notice, a Belarusian price list, a Niigata ramen catalogue, an Arabic-labelled chart, a Hindi house journal and an Italian letterhead, and the answer span is a different structural object in each. Two consequences follow for benchmark design. First, reporting a single averaged multilingual score conflates script recognition with layout understanding, since a model can fail panel (b) for want of table parsing while failing panel (e) for want of Devanagari reading. Second, right-to-left scripts such as the Arabic of panel (d) alter the reading order that patch-level models implicitly assume, an effect that no amount of additional training data in left-to-right languages will surface.

\begin{figure}[!t]
    \centering
    \includegraphics[width=0.72\linewidth]{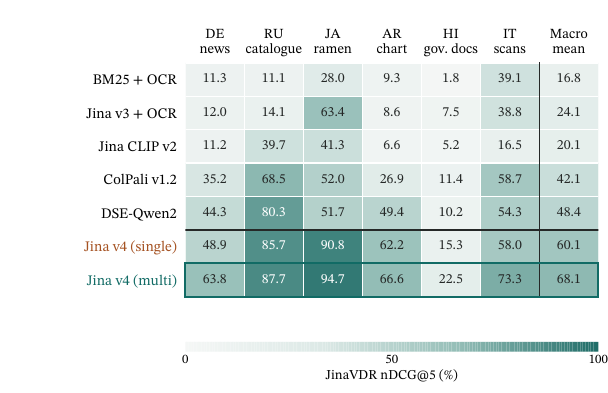}
    \caption{\textbf{Model performance on the six subsets illustrated in Figure~\ref{fig:multilingual_cases}.} Results from Jina-VDR \citep{gunther2025jina}: cells show nDCG@5 (\%) on a fixed 0--100 scale; bold values mark the per-column maximum, and the outlined row wins every column. ``Macro mean'' is the unweighted average over only these six subsets, rather than JinaVDR's all-task average. Each language code is paired with its document type because language, domain and visual format co-vary.}
    \label{fig:jinavdr_multilingual_performance}
\end{figure}

\Needspace{6\baselineskip}
\textbf{Results expose substantial task-level heterogeneity.}
Figure~\ref{fig:jinavdr_multilingual_performance} compares Jina-VDR results on the same six subsets visualized qualitatively above. Jina v4 (multi) ranks first on every subset and reaches a six-task macro-average of 68.09, compared with 60.14 for its single-vector mode and 48.35 for DSE-Qwen2, the strongest non-Jina-v4 baseline on this slice. Because the two Jina v4 modes share a backbone, their comparison controls model capacity more closely than a cross-model contrast: late interaction adds 7.96 nDCG points on average. The gain is nevertheless far from uniform: it is largest on the two Europeana scan tasks, Italian (+15.28) and German (+14.87), but narrows to +2.05 on the Russian catalogue and +3.88 on the Japanese ramen catalogue. This pattern is consistent with, although it does not by itself prove, the hypothesis that patch-level matching is especially useful when a small answer-bearing region must be localized within degraded or layout-heavy pages. The older systems also reverse order across subsets: DSE-Qwen2 is the strongest non-Jina-v4 model for German, Russian and Arabic; Jina v3 + OCR leads that group for Japanese; and ColPali does so for Hindi and Italian. Nor should these scores be interpreted as an isolated ranking of language difficulty: the best score ranges from only 22.49 on Hindi government documents to 94.65 on the Japanese catalogue, while each column simultaneously changes the script, domain, corpus size, scan quality and evidence structure. A single multilingual average therefore hides both architectural sensitivity and task-specific failure modes, and the comparison supports a claim about heterogeneity across \emph{language--document pairs}, rather than a causal claim that one language is intrinsically harder than another.

\subsection{Reasoning-Intensive Setting}
\label{sec:benchmark_reasoning_setting}


The rapid advancement of MLLMs has expanded information-retrieval evaluation beyond surface-level semantic matching toward tasks that require explicit reasoning \citep{bi2025reasoning,su2025thinking,lin2025mind,yan2024errorradar}. \textbf{Text-centric retrieval has already developed reasoning-aware capabilities} through benchmarks\footnote{Instruction-following retrieval (\textit{e.g.,} FollowIR \citep{weller2024followir} and mFollowIR \citep{weller2025mfollowir}), long-context retrieval (\textit{e.g.,} LongEmbed \citep{zhu2024longembed}) and code retrieval benchmarks (\textit{e.g.,} CoIR \citep{li2024coircomprehensivebenchmarkcode} and CodeRAG-Bench \citep{wang2024coderagbenchretrievalaugmentcode}) are not within the scope of reasoning-intensive retrieval in this survey.} (\textit{e.g.,} BRIGHT \citep{su2024bright}, RAR-b \citep{xiao2024rar}, ATEB \citep{han2025ateb}, R2MED \citep{li2025r2med}), embedding models (\textit{e.g.,} ReasonIR \citep{shao2025reasonir}, ReasonEmbed \citep{chen2025reasonembed}, RaDeR \citep{das2025rader}, DIVER-Retriever \citep{long2025diver}, RITE \citep{liu2025exploring}, REAPER \citep{joshi2024reaper}), and rerankers (\textit{e.g.,} Reason-to-Rank \citep{ji2024reasoningrank}, InteRank \citep{samarinas2025distillation}, Rank1 \citep{weller2025rank1}, Rank-K \citep{yang2025rank}, TFRank \citep{fan2025tfrank}, TS-SetRank \citep{huang2025contextual}, InsertRank \citep{seetharaman2025insertrank}, RGS \citep{xu2025beyond}). VDR is now undergoing the same transition: dense textual information combined with intricate visual layouts (\textit{e.g.,} charts, tables, and diagrams) demands logical deduction beyond traditional OCR-based matching \citep{duan2025docopilot,liao2025doclayllm,nacson2025docvlm,jiang2025docrelation}. 

Recent pioneering efforts have begun to formalize this reasoning-intensive frontier through three distinct perspectives:

\textbf{\ding{182} Vision-Centric Logic and Abstract Reasoning.} 
As highlighted by \textbf{MR2-Bench}~\citep{zhou2025mr}, current VDR models often suffer from "shallow matching," where they succeed by identifying object-text correlations but fail at logical, spatial, or causal inference. MR2-Bench introduces vision-centric tasks such as \textit{Visual Puzzle} (solving Raven-style matrices) and \textit{Visual Illustration Search} (\textit{e.g.,} matching a mathematical formula to its corresponding geometric proof). Their findings reveal a massive "reasoning gap": models achieving high scores on general multimodal benchmarks (\textit{e.g.,} MMEB) exhibit a significant performance drop when required to solve abstract visual analogies or multi-image relational scenarios.

\textbf{\ding{183} Expert-Level Multidisciplinary and Contradiction Reasoning.} 
Moving beyond general knowledge, \textbf{MRMR}~\citep{zhang2025mrmr} introduces \textit{Contradiction Retrieval}, a novel task requiring models to identify rules or requirements that conflict with a given case description (\textit{e.g.,} identifying a traffic violation in a visual scene based on a text-based rulebook). MRMR focuses on expert domains like medicine and engineering, where visually rich documents (\textit{e.g.,} pathological slides) require specialized interpretation. Figure~\ref{fig:mrmr_reasoning_cases} makes the distinction among its three task families concrete: Knowledge grounds a visual symbol in expert facts, Theorem infers and instantiates the rule needed to solve a problem, whereas Contradiction verifies a polarity conflict between visible evidence and a candidate statement. Consequently, success requires an intermediate inference that is not literally shared by the query and relevant document, rather than surface-level image--text overlap alone.

\begingroup
\raggedbottom
\begin{figure}[H]
    \centering
    \includegraphics[width=0.90\textwidth]{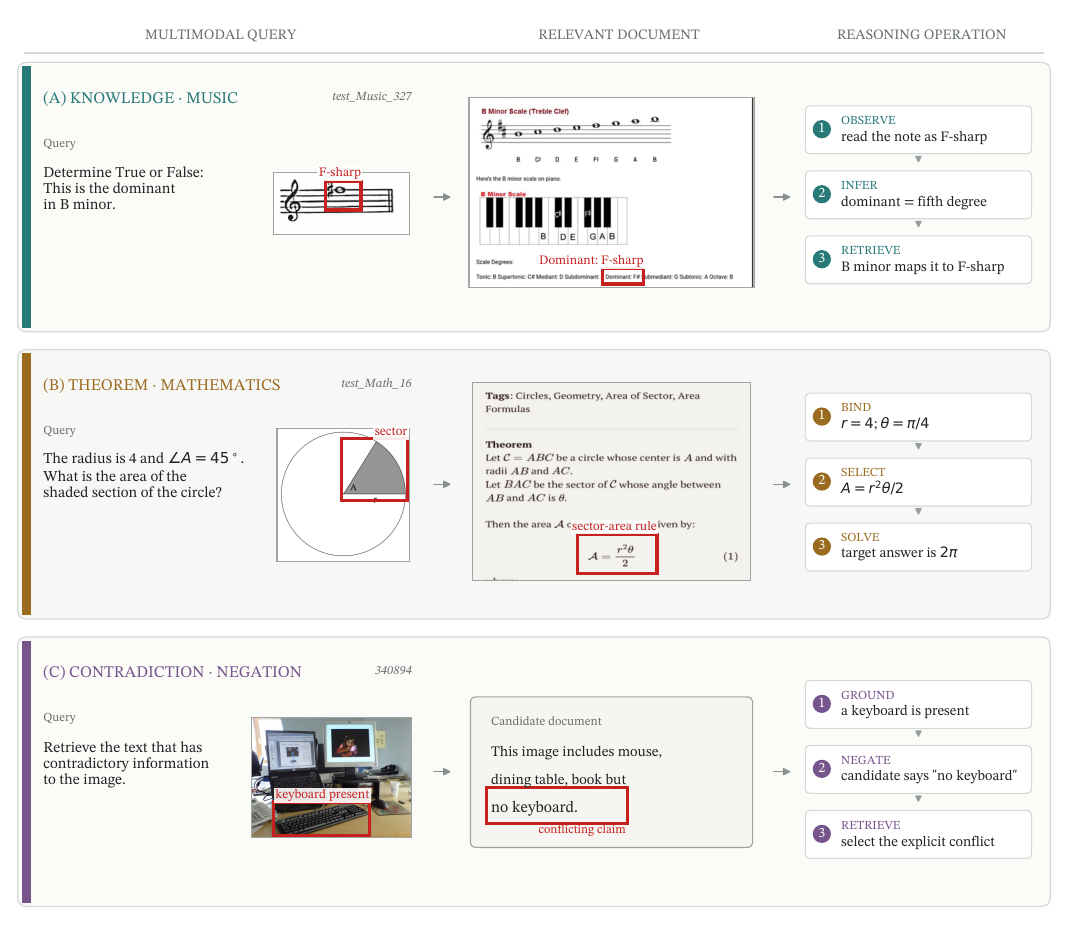}
    \caption{\textbf{Three reasoning operations in MRMR.} Each row links a multimodal query to its relevant document through an explicit inference; \textcolor{red}{\textbf{red}} boxes mark the decisive cues. Knowledge grounds F-sharp in B minor, Theorem applies the sector-area rule, and Contradiction detects the conflict between a visible keyboard and ``no keyboard.'' Cases are from MRMR~\citep{zhang2025mrmr}.}
    \label{fig:mrmr_reasoning_cases}
\end{figure}

Figure~\ref{fig:mrmr_table3_performance} further shows that Qwen3-Embedding with image captions obtains the best overall score (54.1) and leads 9 of the 11 subtasks, indicating that language-model-mediated visual interpretation remains stronger than direct multimodal alignment on this benchmark. Even the strongest native multimodal configuration, Ops-MM-Embedding with merged images, falls from a 67.4 Knowledge mean to 37.4 on Theorem and 36.6 on Contradiction; moreover, 15 of the 16 configurations remain at or below 20 Hit@1 on Negation, below its 25\% random baseline.

\begin{figure}[H]
    \centering
    \includegraphics[width=0.92\textwidth]{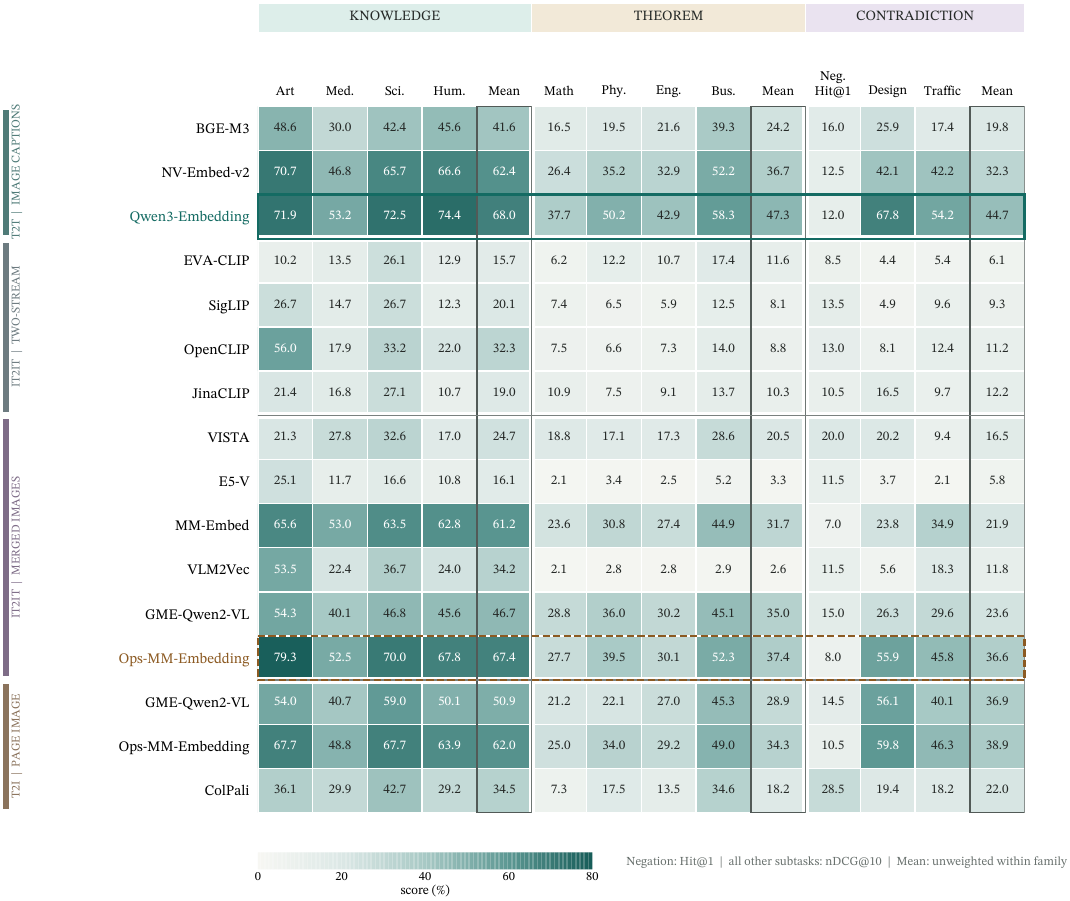}
    \caption{\textbf{MRMR performance by task family and subtask.} Results from MRMR \citep{zhang2025mrmr}: the matrix retains all 16 configurations and 11 subtask scores, with unweighted family means. Negation uses Hit@1 (25\% chance), while the other columns use nDCG@10. Bold values mark maxima; the solid and dashed outlines identify the best overall and strongest native multimodal configurations, respectively.}
    \label{fig:mrmr_table3_performance}
\end{figure}

\textbf{\ding{184} Agentic, Multi-hop, and Process-Oriented Reasoning.} 
In complex enterprise scenarios, \textbf{M4DocBench}~\citep{dong2025doc} redefines reasoning as an iterative, agentic process. It introduces the \textit{M4 framework} (Multi-modal, Multi-hop, Multi-document, and Multi-turn), focusing on deep research tasks where evidence is scattered across dozens of documents. Unlike single-shot retrieval, M4DocBench evaluates a system's ability to perform \textit{Query Decomposition} and iterative refinement. It underscores that deep document intelligence requires "Strategic Planning", the ability to filter relevant documents from noisy collections and adaptively select the optimal retrieval granularity (\textit{e.g.,} chunk, page, or summary) based on the evolving state of the research workflow.

\textbf{Future Frontiers.} 
Despite these early triumphs, several reasoning challenges remain largely unexplored. Future VDR research must address \textbf{implicit retrieval intents}, where queries involve fuzzy constraints that can only be resolved through world-knowledge synthesis \citep{zhang2025imprag,wei2024instructrag}. Furthermore, the development of \textbf{active retrieval agents}, capable of self-correcting their search path when initial reasoning leads to dead ends, will be paramount to unlocking the full potential of multimodal document intelligence in open-domain, large-scale scenarios \citep{zhu2024retrieval,liu2025hm,zhang2025toward}.

\endgroup

%% file: sections/3-methodology.tex
\section{Methodology Perspective}
\label{sec:methodology}

This section deconstructs the technical trends that underpin modern VDR, following the two axes declared in \Cref{sec:introduction}.

\Cref{sec:embedding} and \Cref{sec:reranker} traverse the first axis, \textit{what a retriever is}: embedding models that score a query against a precomputed index (Table \ref{tab:vdr_overview}), and rerankers that jointly encode a query--candidate pair to trade throughput for precision. The two are separated by where interaction happens, at index time versus at query time, which is what fixes their asymmetric cost profiles. \Cref{sec:rag_pipeline} then traverses the second axis, \textit{how a retriever is deployed}, from static RAG pipelines to iterative agentic loops. This second axis presupposes the first: an agentic system is assembled \textit{from} the components of \Cref{sec:embedding} and \Cref{sec:reranker} rather than competing with them, so the reader should expect composition here, not a third alternative.

\subsection{Embedding Models}
\label{sec:embedding}

\begin{table*}[!t] 
    \centering
    \small
    \resizebox{\linewidth}{!}{
\begin{tabular}{l l ll llllll l l} 
    \toprule
    \multirow{2}{*}{\makecell{\textbf{Category}}} & 
    \multirow{2}{*}{\makecell{\textbf{Embedding}}} &
    \multicolumn{2}{c}{\textbf{Publication}} & 
    \multicolumn{6}{c}{\textbf{Model}} & 
    \multirow{2}{*}{\makecell{\textbf{Novelty}}} & 
    \multirow{2}{*}{\makecell{\textbf{Resource}}} \\
    
    \cmidrule(lr){3-4} \cmidrule(lr){5-10}
    
    & & 
    \multicolumn{1}{c}{\textbf{Team}} & 
    \multicolumn{1}{c}{\textbf{Venue}} & 
    \multicolumn{1}{c}{\textbf{Params.}} & 
    \multicolumn{1}{c}{\textbf{Backbone}} & 
    \multicolumn{1}{c}{\textbf{Multilingual}} & 
    \multicolumn{1}{c}{\textbf{Multi-Vec}} &
    \multicolumn{1}{c}{\textbf{Training-Free}} &
    \multicolumn{1}{c}{\textbf{Generative}} &
    & \\
    \midrule

VDR & 
Nemotron ColEmbed V2 \citep{moreira2026nemotron} &
NVIDIA &
ECIR'26 &
4B/8B & 
Qwen3-VL/Llama3.2 & 
en & 
\ding{52} & 
- & 
- & 
\encircle[fill=DarkBlue, text=white]{M} \encircle[fill=DarkGreen, text=white]{D} & 
\begin{adjustbox}{valign=m}
    \href{https://huggingface.co/collections/nvidia/nemotron-colembed-v2}{\includegraphics[height=3ex]{icons/icon_huggingface.png}} \hspace{0.05em} 
    \href{https://arxiv.org/pdf/2602.03992}{\includegraphics[height=3ex]{icons/icon_pdf.png}}
\end{adjustbox} \\

VDR &
Argus-Retriever \citep{abdallah2026argus} &
Innsbruck \& Chungbuk &
arxiv'26 &
2B/4B/9B &
Qwen3-VL &
en &
\ding{52} &
- &
- &
\encircle[fill=DarkBlue, text=white]{M} & 
\begin{adjustbox}{valign=m}
    \href{https://arxiv.org/pdf/2606.04300}{\includegraphics[height=3ex]{icons/icon_pdf.png}}
\end{adjustbox} \\
\bottomrule    
    \end{tabular}
    }
    \vspace{-2mm}
    \caption{Comparison of VDR and general multimodal embedding models. In the \textit{Novelty} column, we denote \underline{M}odel-/\underline{D}ata-/\underline{T}raining-/\underline{E}fficiency-level contributions as \encircle[fill=DarkBlue, text=white]{M} / \encircle[fill=DarkGreen, text=white]{D} / \encircle[fill=DarkRed, text=white]{T} / \encircle[fill=DarkYellow, text=white]{E}. In the \textit{Resource} column, we denote the corresponding GitHub codebase, Hugging Face repository, and paper (\textit{e.g.,} arXiv paper, technical report, or blog) as \includegraphics[height=2ex]{icons/icon_github.png} / \includegraphics[height=2ex]{icons/icon_huggingface.png} / \includegraphics[height=2ex]{icons/icon_pdf.png}.}  
    \label{tab:vdr_overview}
\end{table*}

\subsubsection{Formulation}
In the context of VDR, an embedding model, denoted by an encoder $\mathcal{E}(\cdot)$, processes multimodal inputs to produce vector representations. Let $T_q = \{\tau_i^q\}_{i=1}^{n_q}$ be the token sequence for a query $q$ and $P_d = \{p_j^d\}_{j=1}^{n_d}$ the image-patch sequence for a document page $d$. The objective is to learn an encoder that aligns their representations. While early models produced a single vector for each item \citep{zhang2024gme,liu2025lamra,jiang2024vlm2vec}, recent VDR models adopt a multi-vector paradigm, yielding the embedding sets $\mathbf{U}(q)=\mathcal{E}(T_q)$ and $\mathbf{V}(d)=\mathcal{E}(P_d)$ introduced in \Cref{sec:benchmark_formulation}. Their relevance score $s(q,d)$ is often computed using late interaction such as MaxSim \citep{faysse2024colpali}.

\subsubsection{Common VDR Training Sets.} 
The primary distinction between specialized VDR models and general multimodal embedding models lies in their training data. 
VDR models are fine-tuned extensively on visual document datasets to master the fine-grained semantic and layout-aware understanding required for this domain. 
We discuss common VDR training sets as follows.

\paragraph{\ding{182} colpali-train-set.} This dataset serves as the training data for the ColPali model\footnote{\url{https://huggingface.co/datasets/vidore/colpali_train_set}} and represents a hybrid approach to data collection \citep{faysse2024colpali}. It comprises approximately 127,000 query-image pairs, strategically combining established academic benchmarks (63\%), such as DocVQA \citep{mathew2021docvqa}, InfoVQA \citep{mathew2022infographicvqa}, TAT-DQA \citep{zhu2022tatdqa} and arXivQA \citep{li2024multimodalarxiv}, with a custom synthetic dataset (37\%). The synthetic component was created from a diverse collection of web-crawled PDF documents, with pseudo-questions generated by Claude-3 Sonnet. The dataset is intentionally English-only, designed to facilitate research into zero-shot cross-lingual generalization capabilities of VDR models.

\paragraph{\ding{183} VisRAG-Ret-Train-Synthetic-data.} As the synthetic training component for the VisRAG model\footnote{\url{https://huggingface.co/datasets/openbmb/VisRAG-Ret-Train-Synthetic-data}} \citep{yu2024visrag,sun2025visrag}, this dataset is composed entirely of VLM-generated data, totaling around 239k query-document pairs. The corpus was constructed from web-crawled PDFs spanning diverse domains, including college-level textbooks\footnote{\url{https://openstax.org/}}, academic papers from premier conferences like ICML'23 and NeurIPS'23, and product manuals\footnote{\url{https://www.manualslib.com/}}. A powerful VLM, GPT-4o, was leveraged to generate pseudo-queries for these document pages, creating a large-scale resource tailored for training retrieval models on a variety of document layouts and topics.

\paragraph{\ding{184} vdr-multilingual-train.} This dataset\footnote{\url{https://huggingface.co/datasets/llamaindex/vdr-multilingual-train}} marks a significant step towards multilingual VDR, containing nearly 500,000 query-image samples across five languages (English, Spanish, Italian, German, and French). Its construction involved a highly sophisticated pipeline. First, a diverse corpus of \textasciitilde50k documents was scraped from the internet using topic-based search queries for each language. A key innovation was the use of layout analysis to sample pages, ensuring an even distribution of text-only, visual-only, and mixed-modality pages. Synthetic queries were then generated using powerful VLMs (Gemini-1.5-Pro and Qwen2-VL-72B) with an advanced prompting technique that distinguished between general and specific questions to improve query quality. The dataset underwent rigorous cleaning, filtering, and hard-negative mining to enhance its utility for training robust retrieval models.

\paragraph{\ding{185} VDR\_MEGA\_MultiDomain\_DocRetrieval.} This dataset\footnote{\url{https://huggingface.co/datasets/racineai/VDR_MEGA_MultiDomain_DocRetrieval}} represents the largest and most comprehensive resource to date, amalgamating approximately 1.09 million examples across five languages. It functions as a meta-dataset, strategically fusing the three previously mentioned datasets (\texttt{colpali-train-set}, \texttt{VisRAG-Ret-Train-Synthetic-data}, and \texttt{vdr-multilingual-train}) with several new, domain-specific collections. These additions cover specialized fields such as military\footnote{\url{https://huggingface.co/datasets/racineai/VDR_Military}}, energy\footnote{\url{https://huggingface.co/datasets/racineai/VDR_Energy}}, hydrogen technology\footnote{\url{https://huggingface.co/datasets/racineai/VDR_Hydrogen}}, and geotechnical engineering\footnote{\url{https://huggingface.co/datasets/racineai/VDR_Geotechnie}}. By unifying multiple datasets, it provides unparalleled scale and diversity in both language and subject matter, making it an ideal resource for training highly generalized and robust VDR systems.

\paragraph{\ding{186} docmatix.} This is an interleaved multimodal pre-training dataset\footnote{\url{https://huggingface.co/datasets/moca-embed/docmatix}} created for the modality-aware continual pre-training of MoCa models \citep{chen2025moca}. It is adapted from the original Docmatix dataset \citep{laurençon2024building}, a massive-scale Document Visual Question Answering resource containing approximately 2.4 million images and 9.5 million question-answer pairs, which was initially used to fine-tune the Idefics3 model. The adaptation process involves transforming the original question-answer pairs by concatenating document screenshots with their corresponding texts, thereby creating an interleaved format suitable for continuous pre-training.

\paragraph{Common Characteristics.} The current generation of VDR training sets reveals several unifying trends. Firstly, there is a clear paradigm shift towards \textbf{leveraging synthetic data generation at scale}. All major datasets heavily rely on powerful VLMs to create pseudo-queries for large, unannotated document corpora, effectively bypassing the bottleneck of manual annotation. Secondly, these corpora are primarily built from \textbf{web-crawled PDF documents}, which provides a rich diversity of layouts, domains, and styles that mirror real-world scenarios. Thirdly, there is a growing emphasis on \textbf{sophisticated data curation}, including techniques like layout-aware page sampling and automated hard-negative mining, to improve training efficiency and model performance. Finally, a clear trajectory exists towards \textbf{increasing scale and multilingualism}, evolving from English-only datasets to massive, multi-language compilations that enable the development of globally competent models.

\paragraph{Future Directions.} Looking ahead, the optimization of VDR training sets can be advanced in several key directions. The most critical frontier is moving \textbf{beyond simple semantic matching towards reasoning-intensive data}. Future datasets should include query-document pairs that necessitate multi-hop, logical, or causal reasoning to find the correct answer, mirroring the challenges posed by benchmarks like MRMR and M4DocBench. Secondly, there is a need for \textbf{enhanced data authenticity and complexity}. While VLM-generated queries are scalable, they can lack the nuance and "messiness" of real user intent. Future work could explore mining queries from anonymized user logs or using agentic workflows to simulate more realistic information-seeking behaviors. Lastly, training sets could benefit from \textbf{finer-grained structural annotations}. Instead of just matching a query to a page, future datasets could provide explicit links to specific sub-page elements (\textit{e.g.,} a single row in a table, a data point in a chart, or a specific paragraph), which would be invaluable for training models that can perform precise, element-level evidence retrieval.

\subsubsection{Technical Trends}
\paragraph{\textbf{Model Choices.}}
The field has witnessed a clear trend in both model scale and architecture. While early models were based on smaller backbones like BERT~\citep{nussbaum2024nomic}, current VDR embeddings are predominantly built on powerful MLLMs, such as PaliGemma~\citep{beyer2024paligemma} and Qwen-VL series~\citep{bai2025qwen2}. Parameter counts have also escalated, with SOTA models typically ranging from 2 to 8 billion parameters, a significant increase that enables more nuanced comprehension of complex document structures.

\paragraph{\textbf{Multilingual Support.}}
As document intelligence applications become increasingly global, multilingual capability has emerged as a critical focus in VDR. 
Unlike general-domain image retrieval where this aspect is less emphasized, the prevalence of multilingual business reports, academic papers, and official forms necessitates cross-lingual understanding. Recent models like jina-embeddings-v4~\citep{gunther2025jina} and Nemoretriever~\citep{xu2025llama} reflect this shift, offering support for over 20 languages and demonstrating strong performance on multilingual VDR benchmarks such as Jina-VDR \citep{gunther2025jina} and Nayana-IR \citep{kolavi2025m3dr}. Complementing these broad-coverage models, a language-specialised strand is emerging: KoVRE \citep{choi2026kovre} trains a compact Korean VDR encoder, and is released alongside the KoViDoRe benchmark \citep{choi2026kovidore}, an instance of the co-design of model and evaluation that \Cref{sec:reranker} argues is precisely what multilingual reranking still lacks.

\begin{wrapfigure}{r}{0.46\textwidth}
\centering
\includegraphics[width=\linewidth]{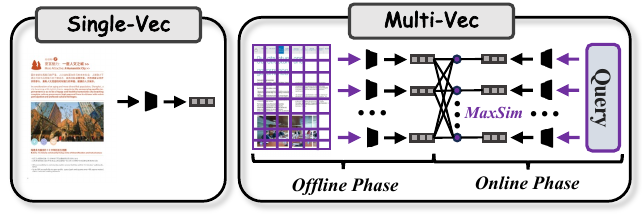}
\caption{Single-vec (\textit{left}) and multi-vec (\textit{right}) VDR.}
\label{fig:multi_vector_illustration}
\end{wrapfigure}
\paragraph{\textbf{Multi-Vector Representation.}}
The multi-vector paradigm, popularized in the VDR domain by ColPali \citep{faysse2024colpali}, has become a dominant approach for fine-grained retrieval \citep{Li2025MIRAGERS,Yang2026UnifiedAE,Chaffin2026ColBERTZeroTP,zhu2026mure}, as shown in Figure \ref{fig:multi_vector_illustration}. 
Instead of compressing an entire document or query into a single embedding, this method generates a set of token-level or patch-level embeddings. 
This granular representation is particularly effective for VDR because it enables "late interaction" matching \citep{khattab2020colbert}, where specific phrases in a query can be precisely aligned with corresponding visual or textual regions in a document, thus preserving local details often lost in single-vector representations. 
ColPali \citep{faysse2024colpali} adapts PaliGemma to produce multi-vector outputs by treating the embeddings of individual image patches as distinct vectors for late interaction.
PreFLMR \citep{lin2024preflmr} generates its multi-vector representations by concatenating embeddings from text tokens with both global and patch-level visual features, which are further refined through cross-attention to be query-aware.
Addressing efficiency, ColModernVBERT \citep{teiletche2025modernvbert} demonstrates that a compact, bidirectional language encoder can be effectively aligned with a vision encoder to generate granular embeddings; while MetaEmbed \citep{xiao2025metaembed} introduces a fixed set of learnable "Meta Tokens" whose final hidden states serve as a compact and scalable multi-vector representation, enabling flexible trade-offs between retrieval quality and efficiency at test-time.
\begin{figure*}[t!]
    \centering
    \includegraphics[width=\linewidth]{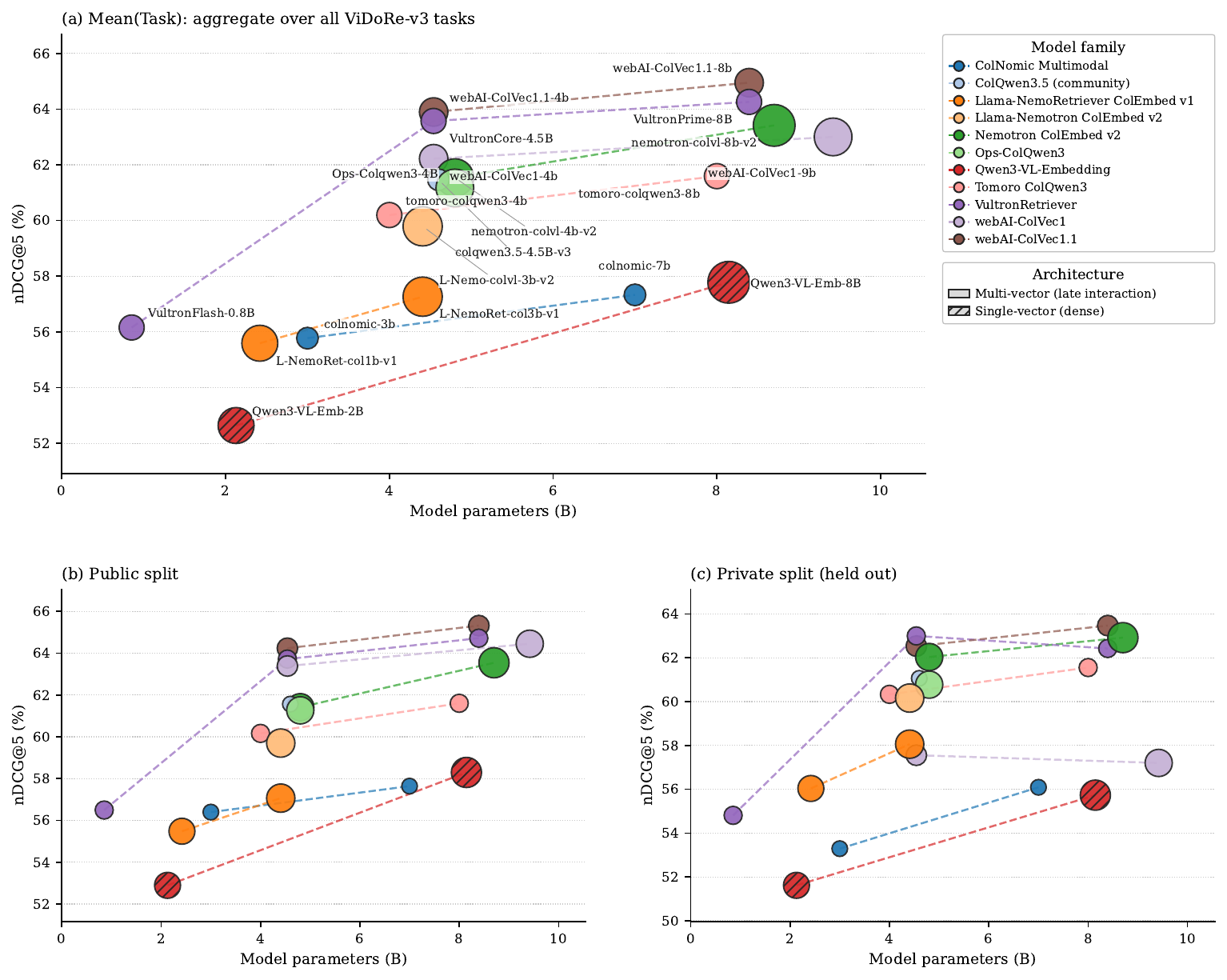}
    \caption{\textbf{Why late interaction became the default VDR architecture.} Parameter count against retrieval quality for the top-20 entries of the public ViDoRe-v3 leaderboard, using the latest results available as of 25 Aug, 2026. Panel (a) aggregates all tasks; panels (b) and (c) repeat the analysis on the public and held-out private splits, and share the legends of panel (a). Colour groups a model family, with dashed lines joining its released scales; bubble area is proportional to embedding dimension, the per-vector cost that late interaction multiplies by the number of retained patches; hatching marks the only single-vector (dense) models in the top-20. The dense frontier sits below the multi-vector frontier at every scale in all three panels, and the gap does not close as parameters grow.}
    \label{fig:vidore_v3_bubble}
\end{figure*}

The empirical case for this paradigm is now unusually clear-cut, and Figure \ref{fig:vidore_v3_bubble} makes it concrete.
Of the top-20 systems on the public ViDoRe-v3 leaderboard, \textbf{18 are multi-vector late-interaction models}; the only single-vector entries are the two Qwen3-VL-Embedding checkpoints, and the stronger of them ranks only 14th.
Three readings of the figure are worth drawing out.
\ding{182} \textit{The gap is not a scale artefact.} The best dense model (Qwen3-VL-Embedding-8B, 8.14B parameters) reaches 57.78, while a late-interaction model at \textit{fewer} parameters (Nemotron ColEmbed-VL-4B-v2, 4.80B) reaches 61.54, a 3.76-point advantage at roughly 40\% fewer parameters.
\ding{183} \textit{Late interaction buys parameter efficiency.} VultronRetriever-Flash at 0.85B lands within 1.6 points of the 8.14B dense model, a nearly ten-fold parameter reduction, indicating that fine-grained matching recovers signal that a single pooled vector discards regardless of encoder capacity.
\ding{184} \textit{Within-family scaling is real but secondary.} The dashed trajectories climb monotonically, yet their slopes are visibly shallower than the vertical offset between the dense and multi-vector frontiers: choosing the interaction mechanism matters more than adding parameters.
We caution that these are leaderboard-reported numbers under each submitter's own inference configuration rather than a controlled study, so the magnitudes should be read as indicative; the ordering, however, is stable across the public and private splits, as panels (b) and (c) of Figure \ref{fig:vidore_v3_bubble} confirm.

\textbf{The split-level view.} Separating the two splits matters for two reasons. First, it is a contamination check: the public split overlaps corpora that several submitters may have trained on, whereas the private split does not. Second, it tests whether the multi-vector advantage just claimed is an artefact of the public data. It is not, since the ordering of the dense and multi-vector frontiers is preserved on the private split, where the best single-vector model reaches 55.73 against 62.04 for a smaller late-interaction model. The comparison does, however, expose a per-model robustness signal that the aggregate hides. Most systems lose only a fraction of a point moving from public to private, but a few degrade sharply: webAI-ColVec1-9b falls from 64.45 to 57.20, a 7.25-point drop, while its successor webAI-ColVec1.1-8b loses only 1.85 points. Such asymmetric drops are a useful diagnostic for generalisation, and we suggest that future VDR leaderboards report the public--private delta as a first-class metric rather than a derived one.
This gap is precisely what motivates the compression literature of \Cref{app:challenge_performance_efficiency_dilemma}: late interaction is winning on quality while carrying a storage cost that single-vector models do not pay.

A parallel line of work attacks the same limitation from the supervision side rather than the architecture side: \citet{tilli2026beyond} argue that patch-level late interaction degenerates into a ``bag of patches'' that discards global page layout, and recover that structure by distilling a textual layout descriptor into the retriever during training. Hybrid designs go further and reintroduce explicit lexical signal alongside the dense one: DocRetriever \citep{hu2026docretriever} pairs a layout-aware \textit{sparse} embedding with a generalisable few-shot reranker, retaining OCR-free operation while restoring the exact-match behaviour that dense vectors blur.

More recently, ColParse \citep{yan2026beyond} introduces a layout-informed paradigm that leverages document parsing models to extract a compact set of semantically meaningful sub-images. By fusing these local layout-aware embeddings with a global page-level vector, ColParse achieves superior retrieval performance while significantly reducing storage requirements. Meanwhile, CausalEmbed \citep{huo2026causalembed} reframes the problem through an auto-regressive generative lens. It treats multi-vector embedding construction as a sequential latent space generation task, enabling flexible test-time scaling, where retrieval precision can be dynamically adjusted by modifying the token budget at inference, thereby achieving high-fidelity representations with only a fraction of the conventional token count.

\paragraph{\textbf{Training Paradigm Exploration.}}
The predominant training method for VDR embedding models is end-to-end supervised fine-tuning using a contrastive loss, such as InfoNCE \citep{oord2018representation}. 
This objective pulls positive query-document pairs closer together in the embedding space while pushing negative pairs apart. 
Beyond this standard approach, two novel paradigms are gaining traction.

The first paradigm reduces the dependence on annotated \textit{multimodal} supervision. Here it is worth separating two claims that are frequently conflated, since they imply different deployment costs.

E5-V \citep{jiang2024e5} is best described as \textit{single-modality training}, not as a training-free method. Its mechanism is prompt-based: carefully designed prompts drive the MLLM to compress either an image or a sentence into one representation, which bridges the modality gap and, as the authors report, already yields strong embeddings \textit{even without fine-tuning}. Its actual proposal, however, is to then train the model \textbf{exclusively on text pairs}. This is what delivers the headline result: outperforming conventional training on image--text pairs while cutting training cost by roughly 95\% and removing the need to collect multimodal training data altogether. The lesson generalises beyond E5-V: because the prompt already aligns the modalities, the supervision needed to sharpen the representation need not itself be multimodal.

Genuinely training-free retrieval is a stronger and rarer proposition, exemplified by FreeRet \citep{zhu2025freeret}: a plug-and-play framework that drives off-the-shelf MLLMs for both embedding-based search and multiple-choice reranking with \textit{no} parameter updates at all. The practical distinction is that single-modality training still requires a fine-tuning budget and a text corpus but yields a fixed, cheap-to-serve encoder, whereas a training-free pipeline eliminates that budget entirely and pays for it with higher inference-time compute.

The second paradigm explores how to harness the \textbf{generative capabilities} of MLLMs to enhance retrieval. 
Early works such as CAFe \citep{yu2025cafe} and VladVA \citep{ouali2025vladva} introduced hybrid training frameworks; they jointly optimize a contrastive loss for discriminative power with an autoregressive, next-token prediction loss to preserve and leverage the model's inherent generative abilities. 
More recent approaches explicitly integrate reasoning as a precursor to embedding. 
For example, RGE \citep{liu2025reasoning} and TTE \citep{cui2025think} both propose a "think-then-embed" process, where the model first generates an explicit rationale or reasoning trace, and the final embedding is conditioned on this generated context to capture more nuanced semantics. 
Moving one step beyond embedding construction, Retrv-R1 \citep{zhu2025retrvr1} employs reinforcement learning to optimize a step-by-step reasoning process that directly selects among candidates; we therefore treat it as an adjacent generative-retrieval formulation rather than include it in the embedding-model taxonomy below.
Similarly, TTE-v2 \citep{cui2025reason} extends this paradigm with a cascaded retrieval-reranking framework, using joint query-candidate reasoning to refine the initial results and leveraging the reranker as a teacher for hard negative mining. 
Embed-RL \citep{jiang2026embedrl} proposes an Embedder-Guided Reinforcement Learning (EG-RL) framework to optimize a reasoner to produce evidential traceability CoTs that better align with the embedding task. 
MMEmb-R1 \citep{Wang2026MMEmbR1RM} introduces pair-aware reasoning selection to identify beneficial rationales and uses reinforcement learning to adaptively invoke reasoning only when necessary. 
In a different direction, PLUME \citep{He2026PLUMELR} internalizes reasoning into a compact latent rollout of continuous hidden states, replacing expensive explicit CoT generation while preserving the benefits of multi-step computation for efficiency.

\subsubsection{Technical Innovations}
\label{sec:embedding_innovations}

Having placed embedding models on the representation axis, we now ask a different question: \textit{where does a given paper actually intervene?} Two models may share the late-interaction mechanism yet contribute nothing comparable: one may add an architectural component, another only a data recipe. We therefore group contributions into four levels (\textbf{model}, \textbf{data}, \textbf{training} and \textbf{efficiency}) and subdivide each into the concrete strategies the literature has converged on. Figure \ref{fig:embedding_taxonomy} lays out the resulting taxonomy with representative work at the leaves; these are the same four levels annotated per row in Table \ref{tab:vdr_overview}, so the figure and the table are two views of one scheme. We stress that this is a taxonomy of \textit{contribution type}, not a third representation axis: a single model can and often does appear under several levels, and the assignment below reflects each paper's primary claim.

\begin{figure*}[t!]
\centering
\resizebox{\textwidth}{!}{%
\begin{forest}
  for tree={
    grow'=east, parent anchor=east, child anchor=west, anchor=west,
    align=left, edge={-, semithick, draw=gray!65}, l sep=6mm, s sep=1.1mm,
    edge path={\noexpand\path[\forestoption{edge}]
      (!u.parent anchor) -- +(3mm,0) |- (.child anchor)\forestoption{edge label};},
    font=\footnotesize, rounded corners=1.6pt, draw, inner sep=2.4pt,
  },
  where level=0{fill=gray!14, draw=gray!60, font=\small\bfseries, align=center}{},
  where level=1{font=\footnotesize\bfseries, text width=26mm, minimum height=6.4mm}{},
  where level=2{font=\scriptsize\bfseries, text width=33mm, fill=white}{},
  where level=3{font=\scriptsize, text width=94mm, fill=white, draw=gray!45}{},
  [{VDR Embedding\\Innovations}
    [{Model-level}, draw=taxM, fill=taxM!12, text=taxM
      [{Late-interaction\\adaptation}, draw=taxM!70
        [{\textbf{ColPali}~\citep{faysse2024colpali}, \textbf{PreFLMR}~\citep{lin2024preflmr},\\ \textbf{ColModernVBERT}~\citep{teiletche2025modernvbert}}]
      ]
      [{Layout- \& structure-\\informed encoding}, draw=taxM!70
        [{\textbf{ColParse}~\citep{yan2026beyond}, \textbf{ColChunk}~\citep{yan2026colchunk},\\ \textbf{VaRS-Doc}~\citep{wang2026varsdoc}, \textbf{Argus}~\citep{abdallah2026argus}}]
      ]
      [{Compact token\\budgeting}, draw=taxM!70
        [{\textbf{MetaEmbed}~\citep{xiao2025metaembed}, \textbf{CoMa}~\citep{li2025compression},\\ \textbf{CausalEmbed}~\citep{huo2026causalembed}}]
      ]
    ]
    [{Data-level}, draw=taxD, fill=taxD!12, text=taxD
      [{Unified task\\formulation}, draw=taxD!70
        [{\textbf{VLM2Vec}~\citep{jiang2024vlm2vec}, \textbf{UniIR}~\citep{wei2024uniir},\\ \textbf{VLM2Vec-V2}~\citep{meng2025vlm2vec}}]
      ]
      [{Large-scale pair\\synthesis}, draw=taxD!70
        [{\textbf{MegaPairs}~\citep{zhou2025megapairs}, \textbf{mmE5}~\citep{chen2025mme5},\\ \textbf{GME}~\citep{zhang2024gme}}]
      ]
      [{Quality-aware hard\\negative mining}, draw=taxD!70
        [{\textbf{UniME-V2}~\citep{gu2025unimev2}, \textbf{B3}~\citep{thirukovalluru2025breaking}}]
      ]
    ]
    [{Training-level}, draw=taxT, fill=taxT!12, text=taxT
      [{Single-modality \&\\multi-task objectives}, draw=taxT!70
        [{\textbf{E5-V}~\citep{jiang2024e5}, \textbf{jina-emb-v4}~\citep{gunther2025jina},\\ \textbf{MM-Embed}~\citep{lin2024mm}, \textbf{MoCa}~\citep{chen2025moca}}]
      ]
      [{Reasoning-conditioned\\embedding}, draw=taxT!70
        [{\textbf{RGE}~\citep{liu2025reasoning}, \textbf{TTE}~\citep{cui2025think},\\ \textbf{MMEmb-R1}~\citep{Wang2026MMEmbR1RM}, \textbf{PLUME}~\citep{He2026PLUMELR}}]
      ]
      [{RL-optimised\\retrieval policy}, draw=taxT!70
        [{\textbf{Embed-RL}~\citep{jiang2026embedrl}, \textbf{UME-R1}~\citep{lan2025ume},\\ \textbf{MMEmb-R1}~\citep{Wang2026MMEmbR1RM}}]
      ]
    ]
    [{Efficiency-level}, draw=taxE, fill=taxE!12, text=taxE
      [{Compact backbones\\\& distillation}, draw=taxE!70
        [{\textbf{ColFlor}~\citep{masry2025colflor}, \textbf{NanoVDR}~\citep{liu2026nanovdr},\\ \textbf{DistilVDR}~\citep{liu2026distilvdr}, \textbf{Unveil}~\citep{sun2025unveil}}]
      ]
      [{Elastic \& nested\\representations}, draw=taxE!70
        [{\textbf{MRL}~\citep{kusupati2022matryoshka}, \textbf{MM-Matryoshka}~\citep{xiang2026mmmatryoshka},\\ \textbf{SMEC}~\citep{zhang2025smec}}]
      ]
      [{Sparse \& post-hoc\\index reduction}, draw=taxE!70
        [{\textbf{DocPruner}~\citep{yan2025docpruner}, \textbf{Light-ColPali}~\citep{ma2025towards},\\ \textbf{MarginMerge}~\citep{mahdizadeh2026marginmerge}, \textbf{AnchorFold}~\citep{zuo2026anchorfold}}]
      ]
    ]
  ]
\end{forest}%
}
\caption{\textbf{Taxonomy of VDR embedding-model innovations.} Contributions are grouped by \textit{where} they intervene (\textcolor{taxM}{\textbf{model}}, \textcolor{taxD}{\textbf{data}}, \textcolor{taxT}{\textbf{training}}, or \textcolor{taxE!85!black}{\textbf{efficiency}}), matching the M\,/\,D\,/\,T\,/\,E annotations in the \textit{Novelty} column of Table \ref{tab:vdr_overview}. Each level is refined into the strategies the field has actually converged on, with representative work at the leaves. This axis is orthogonal to the representation axis of \Cref{sec:embedding}: it classifies \textit{contribution type}, so a model may legitimately appear under more than one level.}
\label{fig:embedding_taxonomy}
\end{figure*}

\paragraph{\ding{182} Model-level---changing the inference-time representation path.}
This level covers changes to the encoder, the units it emits, or the mechanism by which query and page interact; the defining test is that the model computes a different representation even if its data and loss are held fixed. \textit{Late-interaction adaptation} first made visual patches directly retrievable: ColPali \citep{faysse2024colpali} projects PaliGemma patch states and applies token-to-patch MaxSim, preserving local evidence without an OCR bottleneck. \textit{Layout- and structure-informed encoding} instead changes the unit being matched. ColParse \citep{yan2026beyond}, for example, replaces an undifferentiated patch grid with parsed semantic sub-images and combines their local vectors with a global page vector, so sections, tables and figures remain distinct retrieval units. \textit{Compact token budgeting} controls how many such units reach the index: MetaEmbed \citep{xiao2025metaembed} summarizes a page into a fixed set of learnable meta tokens, whereas CausalEmbed \citep{huo2026causalembed} autoregressively produces a variable number of latent vectors, turning representation fidelity into a test-time token-budget decision.

\paragraph{\ding{183} Data-level---changing which relevance relations are learned.}
Data-level innovations leave the serving architecture largely intact but reshape the tasks, positive pairs and confusing negatives from which its geometry is learned. Under \textit{unified task formulation}, VLM2Vec \citep{jiang2024vlm2vec} casts heterogeneous multimodal tasks as instruction-conditioned retrieval in MMEB, while UniIR \citep{wei2024uniir} places image--text and cross-modal search in one retrieval interface; both broaden transfer by making formerly separate tasks share a ranking objective. \textit{Large-scale pair synthesis} expands coverage beyond manually labelled corpora: MegaPairs \citep{zhou2025megapairs} mines heterogeneous visual-semantic and visual-pattern relations before synthesising query--candidate pairs, so scale is coupled to relationship diversity rather than simple duplication. Finally, \textit{quality-aware hard-negative mining} determines which near-misses should shape the boundary. UniME-V2 \citep{gu2025unimev2} uses an MLLM judge to assign graded semantic supervision to retrieved candidates, whereas B3 \citep{thirukovalluru2025breaking} builds a graph over the corpus and constructs batches rich in mutual hard negatives; the former improves label quality and the latter improves negative density, so they belong to the same data-level scope but are not the same mechanism.

\paragraph{\ding{184} Training-level---changing how supervision is converted into an encoder.}
This level concerns objectives, schedules and optimisation policies rather than the examples themselves. \textit{Single-modality and multi-task objectives} relax the standard one-loss/one-output recipe: E5-V \citep{jiang2024e5} uses prompting to bridge modalities and then fine-tunes only on text pairs, while jina-embeddings-v4 \citep{gunther2025jina} jointly learns single-vector and late-interaction outputs with task-specific adapters. \textit{Reasoning-conditioned embedding} inserts computation before compression: RGE \citep{liu2025reasoning} generates an explicit rationale and conditions the embedding on it, whereas PLUME \citep{He2026PLUMELR} replaces readable chains of thought with compact continuous latent rollouts. \textit{RL-optimised embedding policies} supervise the reasoning or control decision rather than only the final similarity: Embed-RL \citep{jiang2026embedrl} rewards evidential traces that improve a downstream embedder, UME-R1 \citep{lan2025ume} optimizes reasoning-driven embedding generation, and MMEmb-R1 \citep{Wang2026MMEmbR1RM} learns when reasoning is beneficial. These methods therefore remain within the embedding scope while optimizing when and how retrieval reasoning is performed.

\paragraph{\ding{185} Efficiency-level---changing the deployable cost envelope.}
Efficiency-level work treats model depth, vector width and vector count as explicit resources, seeking lower indexing or query-time cost without redefining relevance. \textit{Compact backbones and distillation} reduce encoder cost: NanoVDR \citep{liu2026nanovdr} distils a 2B visual retriever into a roughly 70M text encoder, while DistilVDR \citep{liu2026distilvdr} uses dual students to retain end-to-end visual retrieval in a smaller model. \textit{Elastic and nested representations} train one checkpoint for several budgets; MM-Matryoshka \citep{xiang2026mmmatryoshka} nests both vector prefixes and intermediate encoder layers, whereas SMEC \citep{zhang2025smec} stabilises dimension-wise Matryoshka learning through a sequential schedule. \textit{Sparse and post-hoc index reduction} operates after encoding: DocPruner \citep{yan2025docpruner} removes low-importance patch vectors, while AnchorFold \citep{zuo2026anchorfold} selects attention-central anchors and folds non-anchor evidence into them. The last distinction matters operationally: pruning saves count by discarding vectors, whereas folding tries to preserve their contribution inside a smaller index.

\subsection{Reranker Models}
\label{sec:reranker}

Table~\ref{tab:reranker_comparison} summarizes the architectures, ranking paradigms and language coverage of representative multimodal document rerankers before we examine their formulation and emerging trade-offs.

\begin{table*}[!t] 
    \centering
    \small
    \resizebox{\linewidth}{!}{
    \begin{tabular}{l ll llll l l} 
    \toprule
    \multirow{2}{*}{\makecell{\textbf{Reranker}}} & 
    \multicolumn{2}{c}{\textbf{Publication}} & 
    \multicolumn{4}{c}{\textbf{Model}} & 
    \multirow{2}{*}{\makecell{\textbf{Benchmark*}}} & 
    \multirow{2}{*}{\makecell{\textbf{Resource}}} \\
    
    \cmidrule(lr){2-3} \cmidrule(lr){4-7}
    
    & \multicolumn{1}{c}{\textbf{Team}} & \multicolumn{1}{c}{\textbf{Venue}} & \multicolumn{1}{c}{\textbf{Para.}} & \multicolumn{1}{c}{\textbf{Backbone}} & \multicolumn{1}{c}{\textbf{Ranking}} & \multicolumn{1}{c}{\textbf{Multilingual}} & & \\
    \midrule

    ZipRerank \citep{sun2026ziprerank} &
    Magellan Tech. RI &
    ICML'26 &
    8B &
    Qwen3-VL &
    \encircle[fill=DarkBlue, text=white]{LI} &
    en &
    MMDocIR &
    \begin{adjustbox}{valign=m}
        \href{https://arxiv.org/pdf/2605.11864}{
            \includegraphics[height=3ex]{icons/icon_pdf.png}
        }
    \end{adjustbox} \\

    miniReranker \citep{fan2026minireranker} &
    EIT &
    arxiv'26 &
    2B &
    Qwen3-VL &
    \encircle[fill=DarkBlue, text=white]{PO} &
    en &
    MMDocIR &
    \begin{adjustbox}{valign=m}
        \href{https://arxiv.org/pdf/2606.10759}{
            \includegraphics[height=3ex]{icons/icon_pdf.png}
        }
    \end{adjustbox} \\

    Lychee-rerank-mm \citep{dai2025supervised} &
    HIT &
    ICLR'26 &
    7B & 
    Qwen2.5-VL & 
    \encircle[fill=DarkBlue, text=white]{PO} & 
    en & 
    MRB/MRMR & 
    \begin{adjustbox}{valign=m}
        \href{https://github.com/vec-ai/lychee-rerank-mm}{
            \includegraphics[height=3ex]{icons/icon_github.png}
        }
        \hspace{0.05em} 
        \href{https://huggingface.co/vec-ai/lychee-rerank-mm}{
            \includegraphics[height=3ex]{icons/icon_huggingface.png}
        }
        \hspace{0.05em} 
        \href{https://www.arxiv.org/pdf/2510.14824}{
            \includegraphics[height=3ex]{icons/icon_pdf.png}
        }
    \end{adjustbox} \\
    
    UniME-V2-Reranker \citep{gu2025unimev2} & 
    MiroMind AI & 
    AAAI'26 & 
    7B & 
    Qwen2.5-VL & 
    \encircle[fill=DarkBlue, text=white]{PA} \encircle[fill=DarkBlue, text=white]{LI} & 
    en & 
    MMEB & 
    \begin{adjustbox}{valign=m}
        \href{https://github.com/GaryGuTC/UniME-v2}{
            \includegraphics[height=3ex]{icons/icon_github.png}
        }
        \hspace{0.05em} 
        \href{https://huggingface.co/TianchengGu/UniME-V2-reranker-Qwen25VL-7B}{
            \includegraphics[height=3ex]{icons/icon_huggingface.png}
        }
        \hspace{0.05em} 
        \href{https://arxiv.org/pdf/2510.13515}{
            \includegraphics[height=3ex]{icons/icon_pdf.png}
        }
    \end{adjustbox} \\

    Qwen3-VL-Reranker \citep{li2026qwen3vlembed} & 
    Alibaba & 
    arxiv'26 & 
    2B/8B & 
    Qwen3-VL & 
    \encircle[fill=DarkBlue, text=white]{PO} & 
    >30 & 
    MMEB-v2/JinaVDR/ViDoRe-v3 & 
    \begin{adjustbox}{valign=m}
        \href{https://github.com/QwenLM/Qwen3-VL-Embedding}{
            \includegraphics[height=3ex]{icons/icon_github.png}
        }
        \hspace{0.05em} 
        \href{https://huggingface.co/collections/Qwen/qwen3-vl-reranker}{
            \includegraphics[height=3ex]{icons/icon_huggingface.png}
        }
        \hspace{0.05em} 
        \href{https://arxiv.org/pdf/2601.04720}{
            \includegraphics[height=3ex]{icons/icon_pdf.png}
        }
    \end{adjustbox} \\

    llama-nemotron-rerank-vl \citep{moreira2026nemotron} &
    NVIDIA &
    ECIR'26 &
    3B & 
    Llama 3.2 & 
    \encircle[fill=DarkBlue, text=white]{LI} & 
    en & 
    MIRACL-Vision/ViDoRe-v1/v2/v3 & 
    \begin{adjustbox}{valign=m}
        \href{https://huggingface.co/nvidia/llama-nemotron-rerank-vl-1b-v2}{
            \includegraphics[height=3ex]{icons/icon_huggingface.png}
        }
        \hspace{0.05em} 
        \href{https://arxiv.org/pdf/2602.03992}{
            \includegraphics[height=3ex]{icons/icon_pdf.png}
        }
    \end{adjustbox} \\

    Rank-Nexus \citep{Cai2026WhenVM} & 
    Malaya & 
    arxiv'26 & 
    2B & 
    Qwen3-VL/InternVL3 & 
    \encircle[fill=DarkBlue, text=white]{LI} & 
    en & 
    MMDocIR & 
    \begin{adjustbox}{valign=m}
        \href{https://arxiv.org/pdf/2601.20623}{
            \includegraphics[height=3ex]{icons/icon_pdf.png}
        }
    \end{adjustbox} \\

    LamRA-Rank \citep{liu2025lamra} & 
    SJTU \& Xiaohongshu & 
    CVPR'25 & 
    7B & 
    Qwen2.5-VL & 
    \encircle[fill=DarkBlue, text=white]{PO} \encircle[fill=DarkBlue, text=white]{LI} & 
    en & 
    M-BEIR & 
    \begin{adjustbox}{valign=m}
        \href{https://github.com/Code-kunkun/LamRA}{
            \includegraphics[height=3ex]{icons/icon_github.png}
        }
        \hspace{0.05em} 
        \href{https://huggingface.co/code-kunkun/LamRA-Rank}{
            \includegraphics[height=3ex]{icons/icon_huggingface.png}
        }
        \hspace{0.05em} 
        \href{https://arxiv.org/pdf/2412.01720}{
            \includegraphics[height=3ex]{icons/icon_pdf.png}
        }
    \end{adjustbox} \\

    DocReRank \citep{wasserman2025docrerank} & 
    WIS & 
    EMNLP'25 & 
    2B & 
    Qwen2-VL & 
    \encircle[fill=DarkBlue, text=white]{PO} & 
    en & 
    ViDoRe-v2/Real-MM-RAG & 
    \begin{adjustbox}{valign=m}
        \href{https://huggingface.co/DocReRank/DocReRank-Reranker}{
            \includegraphics[height=3ex]{icons/icon_huggingface.png}
        }
        \hspace{0.05em} 
        \href{https://arxiv.org/pdf/2505.22584}{
            \includegraphics[height=3ex]{icons/icon_pdf.png}
        }
    \end{adjustbox} \\

    RagVL \citep{chen2024mllm} & 
    IDEA & 
    EMNLP Findings'25 & 
    1B/2B/4B/13B & 
    Qwen-VL/InternVL/LLaVA-v1.5 & 
    \encircle[fill=DarkBlue, text=white]{PO} & 
    en & 
    - & 
    \begin{adjustbox}{valign=m}
        \href{https://github.com/DataArcTech/RagVL}{
            \includegraphics[height=3ex]{icons/icon_github.png}
        }
        \hspace{0.05em} 
        \href{https://aclanthology.org/2025.findings-emnlp.432.pdf}{
            \includegraphics[height=3ex]{icons/icon_pdf.png}
        }
    \end{adjustbox} \\

    MM-R5 \citep{xu2025mm} & 
    DP Tech & 
    arxiv'25 & 
    7B & 
    Qwen2.5-VL & 
    \encircle[fill=DarkBlue, text=white]{LI}  & 
    en & 
    MMDocIR & 
    \begin{adjustbox}{valign=m}
        \href{https://github.com/i2vec/MM-R5}{
            \includegraphics[height=3ex]{icons/icon_github.png}
        }
        \hspace{0.05em} 
        \href{https://huggingface.co/i2vec/MM-R5}{
            \includegraphics[height=3ex]{icons/icon_huggingface.png}
        }
        \hspace{0.05em} 
        \href{https://arxiv.org/pdf/2506.12364}{
            \includegraphics[height=3ex]{icons/icon_pdf.png}
        }
    \end{adjustbox} \\

    jina-reranker-m0 \citep{JinaAI2025rerankerm0} & 
    Jina AI & 
    blog'25 & 
    2.4B & 
    Qwen2-VL & 
    \encircle[fill=DarkBlue, text=white]{PO} & 
    29 & 
    ViDoRe-v1/M-BEIR & 
    \begin{adjustbox}{valign=m}
        \href{https://huggingface.co/jinaai/jina-reranker-m0}{
            \includegraphics[height=3ex]{icons/icon_huggingface.png}
        }
        \hspace{0.05em} 
        \href{https://jina.ai/news/jina-reranker-m0-multilingual-multimodal-document-reranker/}{
            \includegraphics[height=3ex]{icons/icon_pdf.png}
        }
    \end{adjustbox} \\

    MonoQwen2-VL-v0.1 \citep{MonoQwen} & 
    LightOn & 
    blog'24 & 
    2B & 
    Qwen2-VL & 
    \encircle[fill=DarkBlue, text=white]{PO}  & 
    en & 
    ViDoRe-v1 & 
    \begin{adjustbox}{valign=m}
        \href{https://huggingface.co/lightonai/MonoQwen2-VL-v0.1}{
            \includegraphics[height=3ex]{icons/icon_huggingface.png}
        }
        \hspace{0.05em} 
        \href{https://www.lighton.ai/lighton-blogs/monoqwen-vision}{
            \includegraphics[height=3ex]{icons/icon_pdf.png}
        }
    \end{adjustbox} \\

    \bottomrule    
    \end{tabular}
    }
    \caption{Comparison of multimodal document rerankers. * indicates evaluation only on VDR-related benchmarks. In the \textit{Ranking} column, we denote \underline{PO}intwise/\underline{PA}irwise/\underline{LI}stwise designs as \encircle[fill=DarkBlue, text=white]{PO} / \encircle[fill=DarkBlue, text=white]{PA} / \encircle[fill=DarkBlue, text=white]{LI}. In the \textit{Resource} column, we denote the corresponding GitHub codebase, Hugging Face repository, and paper (\textit{e.g.,} arXiv paper, technical report, or blog) as \includegraphics[height=2ex]{icons/icon_github.png} / \includegraphics[height=2ex]{icons/icon_huggingface.png} / \includegraphics[height=2ex]{icons/icon_pdf.png}.} 
    \label{tab:reranker_comparison}
    \vspace{-4mm}
\end{table*}

\subsubsection{Formulation}
A reranker model, denoted by $R(\cdot, \cdot)$, performs query-conditioned joint encoding after first-stage retrieval. At the basic pointwise granularity, it takes a query $q$ and a single candidate document $d_c$ as a combined input. By jointly processing the query-token sequence $T_q = \{\tau_i^q\}_{i=1}^{n_q}$ and document-patch sequence $P_{d_c} = \{p_j^{d_c}\}_{j=1}^{n_{d_c}}$, the model can leverage deep cross-attention to capture fine-grained inter-modal dependencies, yielding the scalar score $\rho(q,d_c)=R(T_q,P_{d_c})$. Pairwise and listwise variants extend the jointly encoded input from one candidate to a pair or an entire candidate list, as detailed in \Cref{sec:reranking_paradigms}.
This deep interaction allows the model to make more accurate relevance judgments than the dot-product-based similarity used in a bi-encoder.

\subsubsection{Technical Trends}

\paragraph{Model Choices.}
Similar to embedding models, the trend in rerankers is towards larger, MLLM-based architectures. Models such as UniME-V2-Reranker~\citep{gu2025unimev2} and LamRA-Rank~\citep{liu2025lamra} utilize powerful Qwen2.5-VL with parameter counts reaching 7 billion. The strong multimodal understanding capabilities of MLLMs make them suited for reranking.

\paragraph{Multilingual Support.}
In stark contrast to the rapid adoption of multilingualism in VDR embedding models, this capability remains largely underdeveloped in the reranker space. Of the eleven rerankers in Table \ref{tab:reranker_comparison}, only two report broad language coverage, namely jina-reranker-m0~\citep{JinaAI2025rerankerm0} with 29 languages and Qwen3-VL-Reranker~\citep{li2026qwen3vlembed} with over 30, while the remainder, including DocReRank~\citep{wasserman2025docrerank}, MM-R5~\citep{xu2025mm} and MonoQwen2-VL-v0.1~\citep{MonoQwen}, are English-only. On the embedding side of Table \ref{tab:vdr_overview}, by contrast, multilingual coverage is routine: Nemoretriever ColEmbed spans 18 languages~\citep{xu2025llama}, jina-embeddings-v4 20~\citep{gunther2025jina}, ColNetraEmbed 22~\citep{kolavi2025m3dr}, and mmE5 93~\citep{chen2025mme5}.

\textbf{Why the asymmetry?} Prior work notes the gap but rarely asks what sustains it. We attribute it to four structural causes rather than to a simple lack of effort, and each suggests a different remedy.

\ding{182} \textit{Supervision cost scales differently.} A multilingual embedder can be trained from (query, positive) pairs alone, with in-batch sampling supplying negatives for free; such pairs are cheap to mine or synthesise per language, which is exactly how the datasets of \Cref{sec:embedding} were built. A reranker needs \textit{graded, comparative} judgements (which of these $k$ candidates is more relevant), and for listwise objectives an ordering over the whole list. Annotation cost therefore grows with languages $\times$ candidates rather than with languages alone, and the resulting labels are far harder to synthesise reliably.

\ding{183} \textit{Inference cost multiplies the research overhead.} As formalised in \Cref{sec:reranker}, a cross-encoder requires a forward pass per query--candidate pair, whereas an embedder amortises document encoding into a reusable index. Every additional language thus multiplies not just training but also validation and ablation cost. Faced with a fixed budget, the higher marginal return lies in improving a multilingual first-stage retriever, since first-stage recall upper-bounds what any reranker can recover.

\ding{184} \textit{Cross-lingual alignment is harder in pixel space.} An embedder can inherit cross-lingual transfer from the backbone's shared semantic space under a single contrastive objective. A document reranker must jointly (i) read text rendered as pixels, possibly in a script different from the query's, and (ii) judge relevance. Both must succeed for the score to be meaningful, so errors compound multiplicatively; and step (i) degrades precisely on low-resource and non-Latin scripts (connected Arabic forms, Indic conjuncts, dense CJK glyphs at low resolution), where the backbone's visual text recognition is weakest. This also explains why the two multilingual exceptions are built on backbones with strong multilingual pre-training rather than on document-specific architectures.

\ding{185} \textit{A missing-evaluation feedback loop.} The multilingual benchmarks that do exist, namely Jina-VDR~\citep{gunther2025jina}, Nayana-IR~\citep{kolavi2025m3dr}, MIRACL-VISION~\citep{osmulski2025miracl} and SDS KoPub VDR~\citep{lee2025sds}, report first-stage retrieval metrics and define no standard reranking protocol over their candidate pools. Without a leaderboard slot there is little incentive to train multilingual rerankers, and without such models there is little pressure to add the protocol: the gap is self-reinforcing. The cheapest intervention is consequently not a new model but a reranking track over existing multilingual candidate pools, which would convert a latent gap into a measurable one.

\subsubsection{Reranking Paradigms}
\label{sec:reranking_paradigms}
Reranker models are typically trained using one or a combination of three main paradigms: 
\vspace{-2mm}
\begin{itemize}[leftmargin=*]
    \item[\ding{182}] \textbf{Pointwise:} This approach evaluates each query-document pair $(q, d_i)$ independently, predicting an absolute relevance score. The model is trained using a loss function like Binary Cross-Entropy (BCE) on the predicted score $\rho_i = \rho(q, d_i)$ against a ground-truth label $y_i \in \{0, 1\}$. It can be formulated as $\mathcal{L}_{\text{pointwise}} = \sum_i \text{BCE}(\rho_i, y_i)$.
    \vspace{-2mm}
    \item[\ding{183}] \textbf{Pairwise:} This method learns relative relevance by comparing pairs. Given a query $q$ and a pair of documents $(d_i, d_j)$ where $d_i$ is more relevant than $d_j$, the model is trained with a margin $m>0$ to ensure $\rho(q, d_i) > \rho(q, d_j)$, formulated as $\mathcal{L}_{\text{pairwise}} = \sum_{(i,j):\,y_i>y_j} \max(0, m - (\rho_i - \rho_j))$.
    \vspace{-2mm}
    \item[\ding{184}] \textbf{Listwise:} This paradigm considers the entire list of candidate documents for a query simultaneously. It optimizes a loss function, such as ListNet's softmax cross-entropy, that directly corresponds to ranking metrics like nDCG, learning to predict the optimal ordering of the full list, formulated as $\mathcal{L}_{\text{listwise}} = -\sum_i \eta_i \log \hat{\eta}_i$, where $\eta_i$ and $\hat{\eta}_i$ are the ground-truth and predicted probabilities assigned to candidate $d_i$.
\end{itemize}
\vspace{-2mm}
Recent rerankers have shown that combining multiple loss functions often yields superior performance. 
LamRA-Rank~\citep{liu2025lamra} and UniME-V2-Reranker~\citep{gu2025unimev2} adopt hybrid training strategies that combine listwise optimization, where the model predicts the correct item's position, with pointwise or pairwise objectives that classify the relevance of individual or paired candidates. 
Other approaches refine a single paradigm to great effect; models like Lychee-rerank-mm~\citep{dai2025supervised} and DocReRank~\citep{wasserman2025docrerank} utilize supervised fine-tuning to frame reranking as a pointwise classification task, predicting "yes" or "no" to align with the generative nature of MLLMs. Besides, MM-R5~\citep{xu2025mm} advances the listwise paradigm by incorporating CoT reasoning and leveraging reinforcement learning with a task-specific reward to optimize the ranked output. 

The three paradigms are usually presented as a modelling choice, but they differ just as much in \textit{cost}, and that is what governs which is deployable. Pointwise scoring is embarrassingly parallel, since each pair is independent and a batch of $k$ candidates is one batched forward pass, but it must produce a \textit{calibrated} absolute score, and calibration drifts across domains, which is why pointwise rerankers often need per-corpus threshold tuning. Listwise scoring sidesteps calibration entirely, since only the induced order matters, but it must place all $k$ candidates in a single context window. For visual documents this is punishing: each page costs hundreds to thousands of visual tokens, so a list of ten pages can exceed the context budget outright and grows the attention cost super-linearly. Pairwise scoring sits between the two, trading $O(k^2)$ comparisons for supervision that is easier to elicit than a full ordering.

\begin{figure*}[!t]
    \centering
    \includegraphics[width=0.98\linewidth]{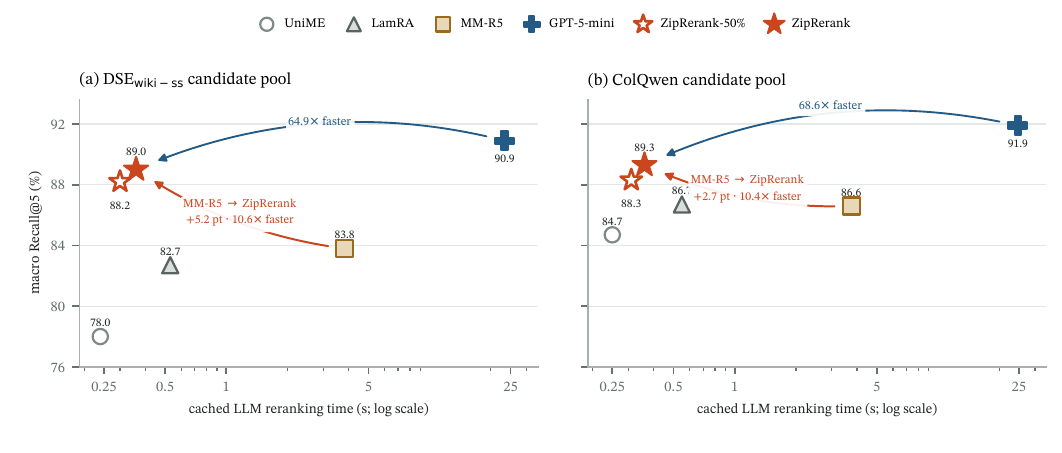}
    \caption{\textbf{The quality--latency frontier of listwise visual-document reranking.} Results from ZipRerank \citep{sun2026ziprerank}; each panel fixes the first-stage candidate pool and compares macro Recall@5 with cached-LLM reranking wall time on a log scale, excluding visual encoding and preprocessing. ZipRerank improves over MM-R5 by 5.2/2.7 points while being 10.6$\times$/10.4$\times$ faster on the DSE$_{\mathrm{wiki-ss}}$/ColQwen pools, respectively; it approaches the stronger GPT-5-mini listwise judge at a small fraction of its latency.}
    \label{fig:ziprerank_tradeoff}
\end{figure*}

This cost asymmetry explains the direction of recent work. ZipRerank \citep{sun2026ziprerank} keeps the listwise formulation but compresses the visual evidence entering the context; Figure~\ref{fig:ziprerank_tradeoff} shows that it reaches 89.0/89.3 macro Recall@5 on the DSE$_{\mathrm{wiki-ss}}$/ColQwen candidate pools, 5.2/2.7 points above MM-R5 while reducing cached reranking time from 3.82/3.74 seconds to 0.36 seconds. It remains within 1.9/2.6 points of GPT-5-mini yet is 64.9$\times$/68.6$\times$ faster, demonstrating that visual-context compression can retain most of the listwise quality without API-model latency. miniReranker \citep{fan2026minireranker} instead accepts the pointwise formulation and attacks its per-pair overhead through visual cache reuse and interaction sparsity. DocRetriever \citep{hu2026docretriever} takes a third route, targeting the \textit{generalisation} weakness of supervised rerankers with reasoning-augmented demonstrations so that a single reranker transfers across domains without per-corpus retraining. The common thread is that the paradigm is increasingly chosen for its cost profile and then repaired, rather than chosen for its statistical elegance.

\subsection{RAG Pipelines \& Agentic Systems}
\label{sec:rag_pipeline}

\subsubsection{Formulation}
The integration of a VDR embedding model $\mathcal{E}(\cdot)$ and a reranker $R(\cdot, \cdot)$ into a RAG pipeline can be formulated as a multi-stage process. Given a query $q$ and a corpus $\mathcal{C}$, the process unfolds as follows:
\vspace{-2mm}
\begin{enumerate}[leftmargin=*]
    \item[\ding{182}] \textbf{First-Stage Retrieval:} The embedding model efficiently retrieves an initial set of $k$ candidate documents $\mathcal{C}_k(q)$ from the corpus according to the relevance function defined in \Cref{sec:benchmark_formulation}: $\mathcal{C}_k(q) = \underset{d \in \mathcal{C}}{\operatorname{Top}_k}\; s(q,d)$.
    \vspace{-2mm}
    \item[\ding{183}] \textbf{Second-Stage Reranking:} The reranker jointly scores or orders the candidate set and returns its top $j \le k$ documents: $\widetilde{\mathcal{C}}_j(q)=\operatorname{Top}_j\!\left(R\bigl(q,\mathcal{C}_k(q)\bigr)\right)$. This notation covers pointwise score sorting as well as pairwise or listwise ordering.
    \vspace{-2mm}
    \item[\ding{184}] \textbf{Augmented Generation:} Finally, a generative MLLM $G(\cdot)$ synthesizes an answer $a$ by conditioning on both the original query and the reranked documents: $a = G\bigl(q,\widetilde{\mathcal{C}}_j(q)\bigr)$.
\end{enumerate}
\vspace{-2mm}
In an \textbf{agentic system}, this process becomes dynamic and iterative. An agent $A$ uses the VDR model as a tool. At each step $t$, based on the query $q$ and internal state (or history) $\sigma_t$, the agent generates an action $\alpha_t = A(q,\sigma_t)$. If $\alpha_t$ issues a retrieval query $q_t$, the VDR system returns an evidence set $\mathcal{Z}_t = \operatorname{Retrieve}(q_t)$, after which the agent updates its state as $\sigma_{t+1}=\operatorname{Update}(\sigma_t,q_t,\mathcal{Z}_t)$ and decides on the next action.

\subsubsection{Current Paradigms and Key Trends}
\label{sec:rag_agent_trends}

The integration of VDR into RAG and agentic systems has moved beyond simple document fetching, evolving into sophisticated frameworks that emulate human-like reasoning and interaction \citep{li2025towards,arya2025advances}. This evolution is characterized by several key trends, progressing from foundational end-to-end pipelines to complex, iterative, and deeply aware reasoning workflows.

\paragraph{\ding{182} Foundational Multimodal RAG Pipelines.} The most fundamental shift has been the move from brittle OCR-based textual RAG to robust, end-to-end multimodal pipelines that process documents as visual inputs. This paradigm preserves critical layout and graphical information often lost in text extraction. A prime example is M3DocRAG~\citep{cho2024m3docrag}, which establishes a flexible framework for multi-page and multi-document question answering by directly retrieving relevant page images for a multimodal generator, forming a foundational approach for subsequent innovations.

\paragraph{\ding{183} Expansion of Interaction Modalities.} Building on the visual-centric pipeline, researchers are expanding the interaction modalities beyond traditional text-based queries to create more natural and accessible interfaces. A pioneering work in this direction is TextlessRAG~\citep{xie2025textlessrag}, which introduces a fully "textless" pipeline that directly processes speech queries and generates spoken answers without any explicit Automatic Speech Recognition (ASR) or Text-to-Speech (TTS) steps, showcasing the potential to significantly broaden the application scenarios of VDR.

\paragraph{\ding{184} Emergence of Agentic and Iterative Reasoning Workflows.} A dominant trend is the replacement of static, linear RAG pipelines with dynamic, agentic systems that perform multi-step, iterative reasoning. These systems decompose complex queries and progressively refine evidence, mimicking human research processes.
\begin{itemize}[leftmargin=*]
    \item \textbf{Task Decomposition and Collaboration:} Many frameworks now employ a "society of agents," where specialized agents collaborate to solve a problem. For instance, MDocAgent~\citep{han2025mdocagent} utilizes a team of five agents (e.g., General, Critical, Text, and Image agents) to synthesize insights from different modalities. Similarly, ViDoRAG~\citep{wang2025vidorag} introduces a coarse-to-fine workflow where a "Seeker" agent hunts for relevant images and an "Inspector" agent provides detailed feedback, enabling iterative evidence refinement.
    \item \textbf{Iterative Deep Research Workflows:} This agentic concept is scaled further in systems designed for "deep research." Doc-Researcher~\citep{dong2025doc} implements a comprehensive multi-agent framework with a "Planner" for query decomposition and a "Searcher-Refiner" loop that iteratively gathers and filters evidence across multiple documents and granularities (\textit{e.g.,} chunks, pages, or summaries).
\end{itemize}

\paragraph{\ding{185} Enhancing Core RAG Components with Advanced Mechanisms.} Beyond structuring the workflow, significant innovation is occurring within the core retrieval and generation components themselves to make them more intelligent and aware.
\begin{itemize}[leftmargin=*]
    \item \textbf{Holistic Knowledge Retrieval and Fusion:} To address the challenge that standard retrieval often misses nuanced information, new methods are designed for more comprehensive knowledge extraction. HKRAG~\citep{tong2025hkrag} explicitly models and retrieves both "salient" and "fine-print" knowledge using a hybrid masking retriever and an uncertainty-guided generator. In a similar vein, VisDoMRAG~\citep{suri2025visdom} runs parallel textual and visual RAG pipelines and then employs a consistency-constrained fusion mechanism to intelligently integrate their outputs.
    \item \textbf{Active Visual Perception and Learning:} The most advanced systems empower agents with the ability to actively interact with retrieved visual content. VRAG-RL~\citep{wang2025vragrl} pioneers this by defining a visual perception action space that allows an agent to perform actions like "crop" and "zoom" on retrieved images. This interactive process is optimized using reinforcement learning, enabling the agent to actively seek out fine-grained details in a coarse-to-fine manner, much like a human analyst.
\end{itemize}

\begin{figure*}[!t]
    \centering
    \includegraphics[width=0.98\linewidth]{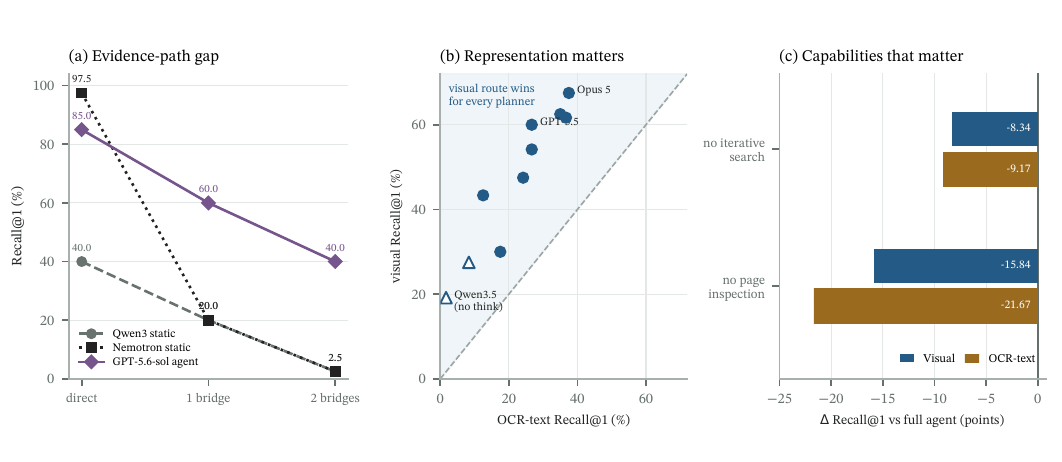}
    \caption{\textbf{What interaction contributes to visual-document retrieval.} Results from VisDocAgentBench \citep{hu2026visdocagentbench}. \textbf{(a)} Recall@1 by evidence-path depth; the GPT-5.6-sol agent and Qwen3 static baseline use the same Qwen3 visual index, whereas Nemotron is a stronger but different static retriever. \textbf{(b)} Visual versus OCR-text Recall@1 across ten planners; marker shape distinguishes open-weight Qwen3.5 planners. \textbf{(c)} Controlled GPT-5.6-sol ablations, expressed as the point loss relative to the full agent. Together, the panels separate the need for iterative evidence acquisition, the value of native visual retrieval, and the contribution of page inspection.}
    \label{fig:visdocagent_findings}
\end{figure*}

\textbf{What does the agentic loop actually buy?} VisDocAgentBench \citep{hu2026visdocagentbench} makes the answer measurable rather than purely architectural. On the shared Qwen3 visual index in Figure~\ref{fig:visdocagent_findings}(a), static Recall@1 falls from 40.0 on direct queries to 2.5 on two-bridge queries, whereas a GPT-5.6-sol agent retains 40.0; the separate Nemotron baseline shows that even 97.5 direct-match recall can still collapse to 2.5 when bridge evidence is required. This is a diagnostic contrast rather than a strict model-controlled comparison, but panel~(c) supplies controlled component evidence: within the same GPT-5.6-sol agent, removing iterative search costs 8.34/9.17 Recall@1 points on the visual/OCR routes, and removing page inspection costs 15.84/21.67.

The representation and planner remain coupled. Every one of the ten planners in Figure~\ref{fig:visdocagent_findings}(b) performs better with visual than OCR-text retrieval, so the modality advantage established in \Cref{sec:embedding} survives inside an agentic loop rather than being washed out by it. At the same time, the wide spread among planners shows that the agent is not a commodity layer over an interchangeable retriever, strengthening the co-design argument of \Cref{app:challenge_interactive_retrieval}.

In summary, the application of VDR in RAG and agentic systems is rapidly maturing from a simple retrieval-and-generation process to highly dynamic, interactive, and collaborative workflows. The field is pushing towards systems that not only find relevant documents but also intelligently reason, synthesize, and interact with multimodal information to solve complex problems \citep{ashraf2025agent,zhang2025see,yan2025mathagent,su2025cafes}.

\subsubsection{The Evolutionary Synergy of VDU and VDR}

The boundary between Visual Document Understanding (VDU) and VDR is rapidly dissolving in the MLLM era. Modern RAG pipelines are evolving into autonomous agentic systems capable of perception, strategic planning, and iterative refinement.

\paragraph{VDU Trends in RAG and Agentic Systems.}
Recent breakthroughs emphasize moving from single-turn OCR-based retrieval to multi-step reasoning-driven navigation. \citep{sourati2025lad} introduces LAD-RAG, which leverages an LLM agent to dynamically interact with a symbolic document graph and neural indices, capturing structural dependencies missed by dense embeddings. \citep{yu2025visual} proposes MACT, a collaborative framework that decomposes document intelligence into planning, execution, and judgment agents, implementing a self-correction loop via procedural scaling. To address evidence sparsity, MHier-RAG \citep{gong2025mhier} facilitates multi-granularity reasoning by retrieving parent pages and document summaries, while MoLoRAG \citep{wu2025molorag} constructs page graphs to navigate logical connections beyond surface-level semantic similarity. For fine-grained localization, DocLens \citep{zhu2025doclens} employs a tool-augmented ``zoom-in'' strategy where agents locate specific visual elements like tables or charts within retrieved pages. Finally, SLEUTH \citep{liu2025resolving} optimizes the input quality through context engineering, utilizing page-screening agents to filter visual noise and construct evidence-dense multimodal contexts.

\paragraph{Comparison between VDR and VDU.} 
Table \ref{tab:vdr_vdu_comparison} illustrates the multi-dimensional differences and the emerging convergence between retrieval and understanding tasks. While VDR focuses on efficient large-scale candidate localization, VDU emphasizes deep reasoning. The integration of the two, as seen in recent agentic frameworks, enables systems to perform ``Retrieval as Understanding.''

\begin{table*}[!h] 
    \centering
    \resizebox{\columnwidth}{!}{
    \begin{tabular}{lccc} 
    \toprule
    \textbf{Dimension} & \textbf{Visual Document Retrieval (VDR)} & \textbf{Visual Document Understanding (VDU)} & \textbf{Convergence Trend} \\
    \midrule
    \textbf{Core Objective} & Candidate page localization & Information extraction \& reasoning & \encircle[fill=DarkBlue, text=white]{R} $\rightarrow$ \encircle[fill=DarkGreen, text=white]{U} Integrated \\
    \textbf{Granularity} & Coarse (Document/Page-level) & Fine (Element/Token-level) & Hierarchical Indexing \\
    \textbf{Compute Pattern} & Broad, shallow (whole corpus) & Narrow, deep (top-$k$) & Adaptive Resource Scaling \\
    \textbf{Key Technique} & Late Interaction, Multi-vector Indexing & Visual CoT, Multimodal Fusion & Agent-driven Seek-then-Verify \\
    \textbf{Main Metric} & nDCG, Recall@$k$ & Accuracy, F1-score & Source Attribution Accuracy \\
    \bottomrule    
    \end{tabular}}
    \caption{Multi-dimensional comparisons between VDR and VDU. Recent agentic systems bridge the gap through iterative reasoning and dynamic filtering.} 
    \label{tab:vdr_vdu_comparison}
\end{table*}

\paragraph{Reflections on VDR: Toward Cognitive Discovery.} 
Future VDR systems should evolve from passive semantic matchers to active cognitive navigators. The integration of reinforcement learning (as in VRAG-RL \citep{wang2025vragrl}) and symbolic-neural fusion (as in LAD-RAG \citep{sourati2025lad}) suggests that the next frontier of multimodal document intelligence lies in the ability to understand the ``logical topology'' of a document. Rather than just returning snippets, future retrievers will actively discover implicit contradictions and synthesize cross-document insights, transforming VDR into a tool for complex decision support in expert domains.

It is worth being precise about what "convergence" does and does not mean here, since the claim is easy to overstate. VDR and VDU are not becoming the same task: their cost structures remain opposed, as Table \ref{tab:vdr_vdu_comparison} records: retrieval must touch the whole corpus cheaply, understanding must touch a handful of pages deeply. What is dissolving is the \textit{interface} between them. In a one-shot pipeline that interface is a ranked list, a lossy summary of everything the retriever knew. In an agentic loop the interface becomes a dialogue, so partial understanding can reshape the next retrieval and retrieval can supply the evidence that understanding lacked. The practical consequence is an evaluation gap: neither nDCG, which ignores whether the evidence sufficed, nor answer accuracy, which cannot say whether a failure was retrieval's or the reader's, measures the quality of that interface. Source-attribution accuracy is the most promising candidate precisely because it is the one metric that both stages can be blamed for.

%% file: sections/4-challenge.tex
\section{Challenges \& Outlook}
\label{sec:challenge}
\begin{figure}[!ht]
    \centering
    \includegraphics[width=0.5 \linewidth]{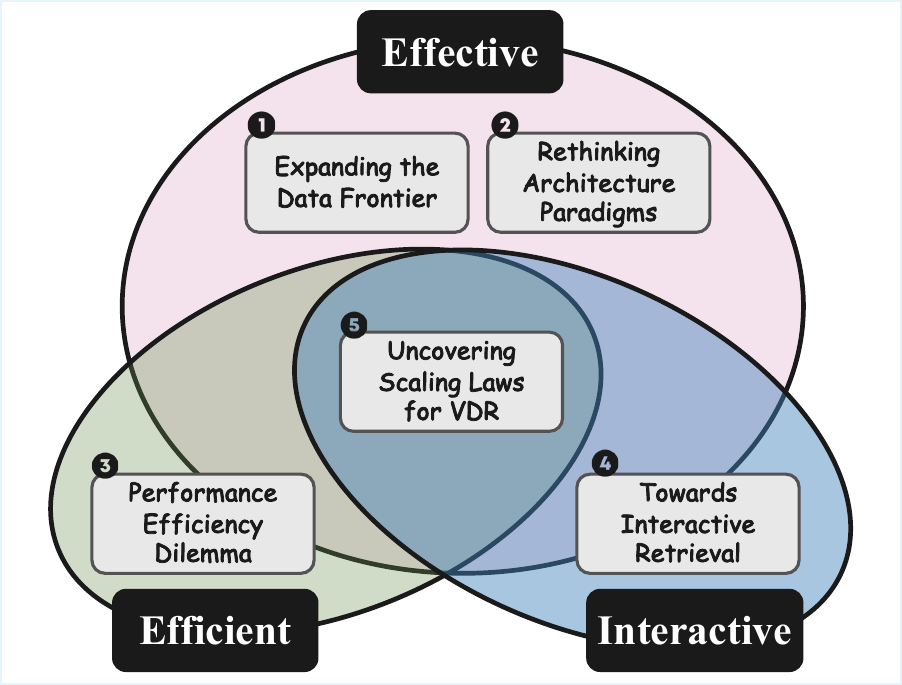}
    \caption{Overview of five future challenges for VDR.}
    \label{fig:challenges_big_pictures}
    \vspace{-6mm}
\end{figure}

\vspace{-2mm}
As shown in Figure \ref{fig:challenges_big_pictures}, building a truly \textit{effective}, \textit{efficient}, and \textit{interactive} VDR system entails persistent challenges. 
The ultimate goal extends beyond simple document matching to enabling nuanced, real-time document intelligence that is scalable and trustworthy. Achieving this vision requires addressing critical bottlenecks across the entire VDR ecosystem. These challenges span from the foundational level of data  ($\vartriangleright$ \Cref{app:challenge_data}) and model architectures ($\vartriangleright$ \Cref{app:challenge_architecture}) to the practical imperatives of performance efficiency  ($\vartriangleright$ \Cref{app:challenge_performance_efficiency_dilemma}) and system interactivity  ($\vartriangleright$ \Cref{app:challenge_interactive_retrieval}), and culminate in the higher-order needs for understanding model scaling behavior  ($\vartriangleright$ \Cref{app:challenge_scaling_law}). The following sections delineate five core challenges and outline promising future directions to navigate these frontiers.
\vspace{-2mm}

\subsection{Expanding the Data Frontier}
\label{app:challenge_data}
The foundation of robust VDR systems is high-quality, diverse, and challenging data, yet the current landscape of benchmarks and training sets presents a significant bottleneck to progress. 
A primary limitation is the lack of diversity in language, domain, and document structure. 
While recent benchmarks like Jina-VDR \citep{gunther2025jina} and MIRACL-VISION \citep{osmulski2025miracl} have made commendable strides in multilingual support, the majority of available resources remain predominantly English-centric and confined to general-domain web documents. 
Furthermore, most benchmarks focus on retrieving single, self-contained pages, failing to capture the complex, multi-hop, and cross-document reasoning required for genuine information synthesis, a gap that pioneering benchmarks like MR2-Bench \citep{zhou2025mr} and MRMR \citep{zhang2025mrmr} are only beginning to explore. 
A more insidious issue lies in data provenance; the heavy reliance on VLM-generated queries for training data, a common practice in datasets, risks creating a hermetic feedback loop where models are evaluated on the same kind of logic they were trained on, inflating performance metrics without guaranteeing real-world generalization. 
This is compounded by the potential for data leakage between large, web-crawled training sets and evaluation benchmarks, which casts a shadow over the true robustness of current models \citep{apicella2025don,lilja2024localization}.

Addressing these data frontiers requires a multi-pronged, forward-looking strategy that moves beyond mere data scaling. 
A primary imperative is the creation of large-scale benchmarks that are \textit{\textbf{not only multilingual but also multi-domain}}, extending into specialized corpora like legal contracts, financial reports, and medical records, which demand expert-level nuance \citep{tang2024we,cao2024recent,chung2025maintaining}. 
To cultivate \textit{\textbf{reasoning-intensive capabilities}}, future data curation can leverage agentic frameworks to automatically generate complex, multi-hop query-evidence chains that span multiple documents, simulating the realistic research workflows envisioned by systems like Doc-Researcher \citep{dong2025doc}.
To break the cycle of synthetic data bias and enhance authenticity, a pivot towards \textit{\textbf{incorporating real human queries}} is essential; this could be achieved by mining anonymized search logs or utilizing human-in-the-loop annotation platforms to capture the ambiguity and complexity of genuine user intent.
Finally, to ensure \textit{\textbf{robust and fair evaluation}}, the community must adopt stricter data governance practices. 
This includes establishing "firewalled" test sets derived from entirely held-out web crawls or proprietary document collections and developing standardized protocols for data contamination detection, thereby fostering a more reliable assessment of model generalization and pushing the field towards true document intelligence \citep{song2024both,xu2024benchmark,samuel2025towards}.

\subsection{Rethinking Architectural Paradigms}
\label{app:challenge_architecture}
The multimodal retrieval field has largely converged on a paradigm centered around contrastively-trained discriminative models, typically bi-encoder architectures that learn to maximize the similarity between positive query-document pairs while minimizing it for negative ones. 
While highly effective for semantic matching, this approach has inherent limitations. 
It primarily treats powerful MLLMs as static feature extractors \citep{mischler2024contextual,jeong2024llm}, underutilizing their sophisticated generative and reasoning capabilities, which can lead to information bottlenecks where fine-grained details or implicit document semantics are lost during compression into a fixed-budget representation. 
Moreover, this "one-size-fits-all" strategy forces a single, monolithic model to act as a generalist for an incredibly diverse array of document structures, from text-heavy reports and dense tables to complex charts and forms, limiting both performance specialization and computational efficiency \citep{gan2025mixture,fan2024towards,lo2025closer}.

To overcome these limitations, future research is pivoting towards two promising architectural shifts: autoregressive retrieval and the integration of Mixture of Experts (MoE). 
The first paradigm \textit{\textbf{reframes retrieval from a purely discriminative task to a generative one}} \citep{deng2025following,xiong2024autoregressive}. For instance, future models could adopt an "embed-via-answering" framework, where embeddings are derived from generating a concise answer to an instruction, as pioneered by InBedder~\citep{peng2024answer}. More advanced paradigms may even involve generating sequences of non-human-readable "soft tokens" that iteratively refine the embedding, a concept introduced by GIRCSE~\citep{tsai2025let} which learns to "speak an embedding language" and shows improved quality with more generation steps at test time. This generative-representational synergy is also validated by frameworks like GritLM~\citep{muennighoff2024generative}, which successfully unifies generative and embedding tasks within a single model, demonstrating that these capabilities can coexist without performance degradation. 
We summarize and compare the representative works from text-only and multimodal generative retrieval in Table \ref{tab:generative_retrieval_comparison_concise}.

\begin{table*}[!ht]
\centering
\resizebox{\textwidth}{!}{%
\begin{tabular}{@{}llllll@{}}
\toprule
\textbf{Representative Works} & \textbf{Modality} & \textbf{Core Paradigm} & \textbf{Generated Content} & \textbf{Embedding Source} & \textbf{Training Objective} \\
\midrule


\textbf{InBedder} \citep{peng2024answer} & Text & Embed-via-Answering & Concise Answer & Hidden state of the 1st generated token & Instruction Tuning (QA) \\

\textbf{GIRCSE} \citep{tsai2025let} & Text & Iterative Refinement & Soft Tokens (Non-readable) & Pooled hidden states of generated soft tokens & Iterative Contrastive Refinement (ICR) \\

\textbf{GritLM} \citep{muennighoff2024generative} & Text & Task Unification & Text Response or Embedding & Mean pooling over input with bidirectional attention & Joint Contrastive + LM Loss \\

\textbf{RGE} \citep{liu2025reasoning} / \textbf{TTE} \citep{cui2025think} & Multimodal & Think-then-Embed & Reasoning Trace (CoT) & Hidden state conditioned on generated reasoning & Joint LM (on trace) + Contrastive Loss \\

\textbf{CAFe} \citep{yu2025cafe} / \textbf{VladVA} \citep{ouali2025vladva}  & Multimodal & Joint Training & N/A (for retrieval) & Last token's hidden state & Joint Contrastive + Autoregressive Loss \\

\textbf{Retrv-R1} \citep{zhu2025retrvr1} & Multimodal & Reasoning-driven Selection & Reasoning trace leading to a selection & Not an embedding model; selects best candidate & SFT + Reinforcement Learning (GRPO) \\

\bottomrule
\end{tabular}%
}
\caption{A multi-dimensional comparison of representative works in \textbf{generative retrieval}. These approaches move beyond static encoding by leveraging the generative capabilities of (M)LLMs. They differ in their core paradigm, the nature of the generated content (from concise answers to reasoning traces), and their training objectives, which range from joint contrastive-generative losses to reinforcement learning.}
\label{tab:generative_retrieval_comparison_concise}
\end{table*}

Concurrently, \textit{\textbf{the MoE architecture offers a path to efficient specialization}}. For instance, different experts can be trained to handle distinct document modalities (\textit{e.g.,} text, tables, audio), a direction explored by Uni-MoE~\citep{li2025unimoe1,li2025unimoe2} and M3-JEPA~\citep{lei2024m3}, which use multi-gate MoE to disentangle modality-specific and shared information. This architecture also unlocks novel embedding sources, as demonstrated by MOEE~\citep{li2024your}, which shows that the expert routing weights themselves can serve as a potent, complementary embedding signal. Together, these architectural innovations promise to evolve VDR from static matching towards more dynamic, specialized, and generative systems.

\subsection{Performance-Efficiency Dilemma}
\label{app:challenge_performance_efficiency_dilemma}

\begin{wrapfigure}{r}{0.50\textwidth}
    \vspace{-3mm}
    \centering
    \includegraphics[width=0.98\linewidth]{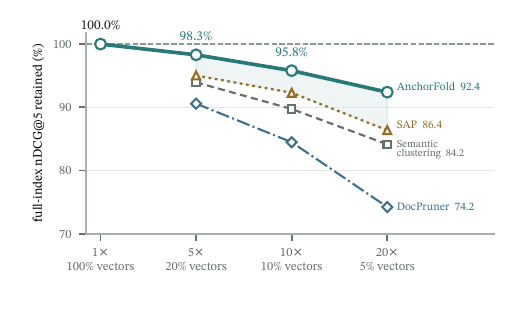}
    \caption{\textbf{Quality retention under index compression.} Results from AnchorFold \citep{zuo2026anchorfold}, averaged across three backbones and the ViDoRe v1/v2 suites. AnchorFold remains above three representative pruning or clustering baselines at every evaluated compression ratio.}
    \label{fig:anchorfold_retention}
    \vspace{-3mm}
\end{wrapfigure}

The high accuracy of state-of-the-art multi-vector VDR models stems directly from their fine-grained representation paradigm, where each document page is encoded into hundreds or even thousands of patch-level embeddings \citep{faysse2024colpali}. While this enables precise query-to-patch matching, it creates a significant performance-efficiency dilemma. The primary issue is the prohibitive storage overhead; for instance, a medium-sized document of just 50 pages can require approximately 10 MB of storage for its embeddings alone \citep{ma2025towards}, rendering the large-scale deployment of such models economically and practically challenging. Furthermore, this abundance of vectors increases online computational costs during the late-interaction scoring stage. Figure~\ref{fig:anchorfold_retention} previews how far a recent structure-aware compressor can move this frontier, motivating embedding reduction techniques that retain retrieval evidence in a much smaller index.

Future solutions to this dilemma are evolving from post-hoc compression to built-in adaptability. The two primary post-hoc strategies are \textit{\textbf{embedding pruning and merging}} \citep{yan2026sculpting,cha2026reinpool,qin2026multi,pony2026colbandit,Liu2026LookIT}. Pruning aims to discard redundant embeddings. While early heuristic-based methods proved unstable \citep{liu2024analysis,lassance2021study,lassance2022learned}, more sophisticated approaches like DocPruner \citep{yan2025docpruner} use attention scores to adaptively prune 50-60\% of patches per document with minimal performance loss. Alternatively, merging or clustering aggregates similar patches, a method that some argue is more suitable as it retains information rather than discarding it. For example, Light-ColPali \citep{ma2025towards} employs hierarchical clustering to merge embeddings, while HEAVEN \citep{kim2025hybrid} uses visually-summarized pages to create a compact first-stage index. Recent work has pushed this in three directions that are worth distinguishing, because they trade different resources. \textit{Learned pooling} keeps the number of vectors fixed but makes the reduction itself trainable, as in the lightweight fine-tuning of \citet{josef2026learn2pool}; \textit{coverage-aware merging} such as MarginMerge \citep{mahdizadeh2026marginmerge} selects which patches to fuse so that the retained set still spans the page rather than clustering redundantly; and \textit{quantisation} such as CrossQ \citep{salla2026crossq} leaves the vector count untouched and instead shrinks each vector's width, which composes with either of the former. AnchorFold \citep{zuo2026anchorfold} illustrates that these are not mutually exclusive, folding attention onto anchor tokens recursively to cut both count and cost. Quantitatively, Figure~\ref{fig:anchorfold_retention} retains 98.3\% of full-index nDCG@5 at 5$\times$ compression and 92.4\% at 20$\times$, while storing only 5\% of the vectors; at the latter budget it remains 6.0 points above the strongest baseline shown. This makes fold depth a promising corpus- or query-adaptive budget that future systems can combine with vector quantisation rather than treating compression methods as mutually exclusive.

\begin{wrapfigure}{r}{0.50\textwidth}
    \vspace{-3mm}
    \centering
    \includegraphics[width=0.98\linewidth]{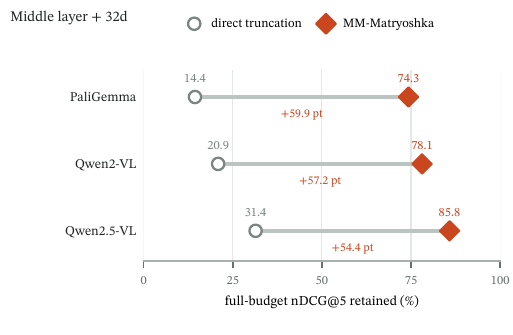}
    \caption{\textbf{Joint depth--width elasticity in MM-Matryoshka.} Results from MM-Matryoshka \citep{xiang2026mmmatryoshka}: ViDoRe-v2 nDCG@5 retained at each backbone's middle layer and 32d, relative to its final-layer 128d reference.}
    \label{fig:mm_matryoshka_retention}
    \vspace{-3mm}
\end{wrapfigure}

A more forward-looking paradigm is \textit{\textbf{Matryoshka Representation Learning (MRL)}} \citep{kusupati2022matryoshka}, which trains embeddings to be inherently flexible. MRL ensures that a single high-dimensional embedding contains a hierarchy of smaller, nested embeddings that are also effective. Flexible Matryoshka embeddings have been adopted by top-performing text embedding models (\textit{e.g.,} Qwen3 Embedding \citep{zhang2025qwen3}, Gemini Embedding \citep{lee2025gemini}, KaLM-Embedding-V2 \citep{zhao2025kalm}, EmbeddingGemma \citep{vera2025embeddinggemma}, jina-embeddings-v3 \citep{sturua2024jina}, jina-colbert-v2 \citep{jha2024jina}) across diverse real-world tasks \citep{lai2024matryoshka,wang2024train,nacar2025enhancing,hanley2025hierarchical,fu2025moon}. MM-Matryoshka \citep{xiang2026mmmatryoshka} transfers this idea to multi-vector VDR by nesting both vector prefixes and encoder layers. At the joint middle-layer-plus-32d budget, Figure~\ref{fig:mm_matryoshka_retention} shows 74.3--85.8\% retention across three backbones, versus only 14.4--31.4\% under direct truncation---a 54.4--59.9-point gain. This makes a single two-dimensionally elastic checkpoint plausible; future systems could select depth and width jointly for each corpus, device, or query instead of maintaining separate retrievers. Formally, at full width $h_{\max}$, the encoder $\mathcal{E}$ produces the multi-vector representations $\mathbf{U}(q)$ and $\mathbf{V}(d)$ introduced in \Cref{sec:benchmark_formulation}, with every constituent vector in $\mathbb{R}^{h_{\max}}$. MRL trains the model on a set of nested widths $\mathcal{H} = \{h_1, h_2, \ldots, h_K\}$, where $h_1 < h_2 < \ldots < h_K = h_{\max}$. For each $h_i$, $\mathbf{U}^{[h_i]}(q)$ and $\mathbf{V}^{[h_i]}(d)$ retain the first $h_i$ components of every query and document vector, respectively. The training objective $\mathcal{L}_{\text{MRL}}$ sums a standard representation loss $\ell(\cdot, \cdot)$ over all nested widths:
\begin{equation}
    \mathcal{L}_{\text{MRL}}(q, d^+) = \sum_{h_i \in \mathcal{H}} \lambda_i\, \ell\left(\mathbf{U}^{[h_i]}(q), \mathbf{V}^{[h_i]}(d^+)\right)
\end{equation}
where $q$ is a query, $d^+$ is a positive document, and $\lambda_i$ are loss weights. This allows practitioners to truncate stored embeddings at inference time, providing a flexible dial to balance performance and efficiency. Building on this, 2DMSE \citep{li20242d} extends this elasticity to model depth, while frameworks like Starbucks \citep{zhuang2024starbucks} and MatTA \citep{verma2025matryoshka} refine the training process to bridge the performance gap with individually trained models. The Matryoshka principle has also been generalized to new granularities and architectures: M3 \citep{cai2024matryoshka} and MQT \citep{hu2024matryoshka} apply it to the number of visual tokens in LVLMs; Matryoshka Re-Ranker \citep{liu2025matryoshka} enables flexible depth and sequence length for re-ranking; and MatMamba \citep{shukla2024matmamba} integrates it into State Space Models. The application of MRL has further expanded to diverse tasks such as improving feature hierarchy in Sparse Autoencoders \citep{bussmann2025learning} and enabling multi-precision models via Matryoshka Quantization \citep{nair2025matryoshka}. Finally, innovations in the training paradigm itself, such as the sequential learning in SMEC \citep{zhang2025smec} and the post-hoc tuning of Matryoshka-Adaptor \citep{yoon2024matryoshka}, are making the creation of these flexible embeddings more efficient and accessible. Besides, recent works also focus on sparse coding \citep{wen2025beyond,guo2026csrv2} and isolation kernel \citep{zhang2026llms} for more lightweight representation learning. 
We compare the representative works above in Table \ref{tab:mrl_comparison}.

\begin{table*}[!h] 
    \centering
    \caption{A multi-dimensional comparison of representative works in \textbf{Matryoshka Representation Learning (MRL)}. The principle of nested, adaptable representations has been extended from embedding dimensions to model depth, token counts, and even quantization bit-widths, addressing a wide range of tasks from efficient retrieval to model interpretability.}
    \label{tab:mrl_comparison}
    \resizebox{\columnwidth}{!}{
    \begin{tabular}{l l l l l} 
    \toprule
    \textbf{Representative Works} & \textbf{Modality} & \textbf{Target Task} & \textbf{Matryoshka On} & \textbf{Core Innovation} \\
    \midrule
    \textbf{MRL} \citep{kusupati2022matryoshka} & Multimodal & Classification / Retrieval & Embedding Dimension & Foundational concept of nested embeddings. \\
    \textbf{2DMSE} \citep{li20242d} & Text & STS / Retrieval & Embedding Dimension \& Model Depth & Extends MRL to two dimensions for greater flexibility. \\
    \textbf{Starbucks} \citep{zhuang2024starbucks} & Text & STS / Retrieval & Embedding Dimension \& Model Depth & Improves 2DMSE to match individually trained models. \\
    \textbf{MatTA} \citep{verma2025matryoshka} & Text & General Tasks & Model Depth \& Width (FFN) & Teacher-TA-Student distillation for elastic student models. \\
    \textbf{M3} \citep{cai2024matryoshka} / \textbf{MQT} \citep{hu2024matryoshka} & Multimodal & VQA / Reasoning & Number of Visual Tokens & Enables adaptive visual granularity in LVLMs. \\
    \textbf{Matryoshka Re-Ranker} \citep{liu2025matryoshka} & Text & Re-ranking & Model Depth \& Sequence Length & Creates flexible re-rankers with configurable architecture. \\
    \textbf{SMEC} \citep{zhang2025smec} & Multimodal & Retrieval / Classification & Embedding Dimension & Sequential training to mitigate gradient variance. \\
    \textbf{Matryoshka-Adaptor} \citep{yoon2024matryoshka} & Multimodal & Retrieval & Embedding Dimension & Post-hoc tuning to impart Matryoshka properties. \\
    \textbf{MM-Matryoshka} \citep{xiang2026mmmatryoshka} & Multimodal & Retrieval & Embedding Dimension \& Encoder Depth & One multi-vector checkpoint with selectable 2D budgets. \\
    \textbf{MatMamba} \citep{shukla2024matmamba} & Multimodal & LM / Classification & Hidden Dimension \& Heads & Generalizes MRL to State Space Model (SSM) architectures. \\
    \textbf{Matryoshka SAE} \citep{bussmann2025learning} & Model Activations & Interpretability & SAE Dictionary Size & Mitigates feature splitting in Sparse Autoencoders. \\
    \textbf{Matryoshka Quantization} \citep{nair2025matryoshka} & Model Weights & Model Compression & Quantization Bit-width & Leverages nested structure of integer data types. \\
    \bottomrule
    \end{tabular}
    }
\end{table*}

\subsection{Towards Interactive Retrieval}
\label{app:challenge_interactive_retrieval}

Integrating VDR with agentic systems holds immense potential, particularly for complex scenarios like \textit{Deep Research}, where a query requires iterative evidence gathering from a vast candidate pool \citep{java2025characterizing,chen2025browsecomp,huang2025deep,zhang2025deep}. The current challenge, however, is that most agentic systems treat VDR models as passive, reactive tools rather than active, strategic partners. These agents typically operate within a predefined, reactive loop of reasoning and acting, which limits their ability to dynamically formulate or adapt a high-level retrieval strategy. For instance, while frameworks like WebThinker~\citep{li2025webthinker} and DeepResearcher~\citep{zheng2025deepresearcher} enable agents to interleave thought with web search actions, they often function as sophisticated \textit{tool-executors} that reactively process information rather than as \textit{research strategists} that proactively plan a multi-step information-seeking campaign. This limitation becomes particularly evident in complex research tasks, such as generating structured reports with dynamic outlines as demonstrated by WebWeaver~\citep{li2025webweaver} or creating text-chart interleaved content as in Multimodal DeepResearcher~\citep{yang2025multimodal}, where the retrieval process itself must be strategically guided to gather diverse and structured evidence.

Future work must focus on the \textit{\textbf{co-design of agents and VDR tools to foster a more organic and strategic synergy}}. A primary direction is to empower agents with explicit retrieval planning capabilities, enabling them to decompose a high-level query into a multi-step, multi-granularity retrieval plan. This moves beyond simple tool use towards strategic orchestration. For example, a planner agent, inspired by the hierarchical structure of Cognitive Kernel-Pro~\citep{fang2025cognitive}, could adaptively select \textit{retrieval granularity} (choosing between document summaries, full pages, or specific visual elements) and delegate these sub-tasks to specialized agents. To handle the inevitable "dead ends" in complex research, agents must also develop self-correction mechanisms. Frameworks like DeepAgent~\citep{li2025deepagent}, with its autonomous memory folding, offer a powerful blueprint by allowing an agent to "take a breath," discard a failed exploration path, and restart from a consolidated memory state. To effectively learn such complex strategies, agents require more fine-grained feedback. This can be achieved through mechanisms like the Atomic Thought Rewards (ATR) proposed in Atom-Searcher~\citep{deng2025atom}, which uses a reasoning reward model to provide process-level supervision on the agent’s retrieval strategy. Finally, for massive-scale research, parallel exploration frameworks like the Research-Synthesis approach in WebResearcher~\citep{qiao2025webresearcher} and Tongyi DeepResearch~\citep{team2025tongyi}, where multiple agents conduct concurrent searches, can transform interactive retrieval into a scalable, collaborative process. The ultimate goal is to evolve the VDR system from a passive tool into an active, \textit{strategic partner} in the knowledge discovery process.

\subsection{Uncovering Scaling Laws for VDR}
\label{app:challenge_scaling_law}

While scaling laws (predictable power-law relationships between model performance, size, and data volume) are well-documented for general-purpose LLMs \citep{alabdulmohsin2022revisiting,zhang2024scaling,xiao2025densing,pearce2024scaling}, their application to the specialized domain of VDR remains a largely unexplored and complex frontier. The core challenge is that the scaling behavior in VDR is not a straightforward extrapolation of model size or data quantity. On the model axis, naively increasing parameter count does not guarantee proportional performance gains. Larger models, without proper fine-tuning, can exhibit greater embedding anisotropy, which paradoxically harms retrieval performance in zero-shot settings by compressing the effective embedding space, even as they capture richer features for transfer tasks \citep{Jiang2024scaling,wu2026tsembed}. This complexity is mirrored on the data axis, where simply increasing the volume of raw documents is insufficient. The effectiveness of VDR models is deeply tied to the quality and diversity of the training pairs. Recent studies powerfully illustrate this, showing that models trained on smaller, high-quality synthetic datasets can significantly outperform those trained on orders-of-magnitude more, but less curated, data. For example, both mmE5 \citep{chen2025mme5} and MegaPairs \citep{zhou2025megapairs} demonstrated that superior performance can be achieved with a fraction of the data if it is diverse and of high quality, underscoring that scaling in VDR is a nuanced interplay between model capacity and data sophistication, rather than a simple numbers game.

To systematically navigate this challenge, future research must pivot from ad-hoc scaling to \textit{\textbf{establishing formal, VDR-specific scaling laws, adapting methodologies from the dense retrieval domain}} \citep{Fang2024scaling}. A foundational step is to adopt continuous and sensitive evaluation metrics, such as contrastive entropy, which can more accurately capture subtle performance changes compared to discrete ranking metrics like nDCG, thus enabling precise power-law curve fitting for parameter count ($N_{\theta}$) and training-set size ($N_{\mathrm{train}}$). Concurrently, advancing the "data" axis of the scaling law requires moving beyond simple data augmentation to sophisticated, document-aware synthetic data generation. This involves leveraging MLLMs not just for query generation but for creating entire ecosystems of diverse training tasks \citep{Wang2024improving}, mining heterogeneous relationships between documents using multiple similarity models (\textit{e.g.,} visual-semantic and visual-pattern correlations) \citep{zhou2025megapairs}, and implementing rigorous quality control through mechanisms like single-pass multi-aspect interpretation and self-refinement \citep{chen2025mme5}. By combining a formal scaling law framework with advanced data synthesis, the VDR community can quantitatively model the trade-offs between investing in larger models versus more extensive data annotation. This will enable more efficient resource allocation and pave the way for building maximally effective VDR systems under practical budget constraints.

%% file: sections/5-conclusion.tex
\section{Conclusion}
\label{conclusion}
This survey systematically discusses the VDR landscape, categorizing the evolution of benchmarks and methodologies, including embedding models, rerankers, and their integration with RAG and agentic systems. 
Current triumphs push the boundaries of fine-grained matching, but the field's trajectory is shifting towards more complex document intelligence. 
Navigating the future frontiers of data complexity, architectural innovation, efficiency, and interactivity will be critical for realizing the full potential of VDR as a cornerstone of multimodal document intelligence.



\section{Broader Impact and Societal Considerations}

While this survey focuses on the technical advancements of VDR in the MLLM era, the deployment of such systems carries significant broader societal implications. As VDR transitions from academic benchmarks to real-world deployment in legal, medical, and enterprise workflows, we must critically evaluate its impact on accountability, equity, and environmental sustainability.

\textbf{Transparency, Accountability, and Trust.} A primary concern in deploying VDR systems, particularly in high-stakes domains, is the "black-box" nature of multimodal reasoning. When retrieval decisions are based on opaque, latent-space alignments, the inability to provide clear, human-interpretable rationales (or citations to evidence) complicates auditability. To ensure responsible deployment, the development of VDR should not only prioritize retrieval accuracy but also emphasize traceable reasoning. Systems that integrate "retrieval-as-reasoning", where the model provides an explicit chain-of-thought or highlights the specific visual/textual evidence, are essential for establishing trust and accountability, especially in regulatory contexts where document decisions must be legally defensible.

\textbf{Mitigating Bias and Promoting Multilingual Equity.} The VDR landscape remains heavily skewed towards English-centric datasets and models. If left unchecked, this "linguistic and cultural bottleneck" threatens to marginalize non-English speaking communities and industries, potentially reinforcing existing global inequities in information access. The surge in multilingual benchmark development is a necessary step, but technical efforts must also address inherent biases within foundation models. Retrieval systems that unknowingly favor dominant cultural or Western-centric document formats risk systematic information omission. We advocate for research that explicitly models fairness across linguistic and structural diversities to ensure that VDR is a globally inclusive technology.

\textbf{Environmental Impact of Large-Scale Intelligence.} The transition from traditional, lightweight OCR-based search to massive-scale VDR models based on multi-vector late-interaction paradigms introduces a significant performance-efficiency dilemma. While these models offer unprecedented semantic depth, they impose substantial computational costs in terms of memory storage, index maintenance, and GPU-intensive inference. In an era where AI-related energy consumption is under intense scrutiny, the pursuit of performance must be balanced with environmental responsibility. Developing sparse, efficient, and Matryoshka-style representations is not merely a technical optimization; it is a critical requirement to ensure that the widespread adoption of multimodal intelligence does not come at the expense of an unsustainable carbon footprint.

\textbf{Societal Safeguards and Data Integrity.} The ability of VDR models to extract fine-grained, structured information from unstructured PDFs, including sensitive personal data, private business intelligence, and restricted intellectual property, poses risks of unintended information disclosure. The shift toward agentic RAG systems that "browse" and synthesize evidence across vast document stores exacerbates the danger of bypassing existing security controls. Therefore, we underscore the necessity of robust data governance, including fine-grained access control and privacy-preserving retrieval architectures, as foundational requirements for future VDR systems.

In summary, the evolution of VDR toward more sophisticated, autonomous, and reasoning-intensive workflows necessitates a proactive approach to societal safeguards. We hope this perspective encourages the community to view technical robustness, interpretability, and ecological sustainability not as peripheral concerns, but as integral pillars of responsible multimodal document intelligence.

%% file: vdr_survey.bib
@inproceedings{alaei2016brief,
  title={A brief review of document image retrieval methods: Recent advances},
  author={Alaei, Fahimeh and Alaei, Alireza and Blumenstein, Michael and Pal, Umapada},
  booktitle={2016 International Joint Conference on Neural Networks (IJCNN)},
  pages={3500--3507},
  year={2016},
  doi={10.1109/IJCNN.2016.7727648},
  url={https://doi.org/10.1109/IJCNN.2016.7727648},
  organization={IEEE}
}

@article{zhou2017recent,
  title={Recent advance in content-based image retrieval: A literature survey},
  author={Zhou, Wengang and Li, Houqiang and Tian, Qi},
  journal={arXiv preprint arXiv:1706.06064},
  year={2017}
}

@article{ding2024deep,
  title={Deep learning based visually rich document content understanding: A survey},
  author={Ding, Yihao and Han, Soyeon Caren and Lee, Jean and Hovy, Eduard},
  journal={arXiv preprint arXiv:2408.01287},
  year={2024}
}

@article{rombach2025deep,
  title={Deep learning based key information extraction from business documents: Systematic literature review},
  author={Rombach, Alexander and Fettke, Peter},
  journal={ACM Computing Surveys},
  volume={58},
  number={2},
  pages={45:1--45:37},
  year={2026},
  doi={10.1145/3749369},
  url={https://doi.org/10.1145/3749369},
  publisher={ACM New York, NY}
}

@article{ding2025survey,
  title={A Survey on MLLM-based Visually Rich Document Understanding: Methods, Challenges, and Emerging Trends},
  author={Ding, Yihao and Luo, Siwen and Dai, Yue and Jiang, Yanbei and Li, Zechuan and Sun, Qiang and Martin, Geoffrey and Liu, Wei and Peng, Yifan},
  journal={arXiv preprint arXiv:2507.09861},
  year={2025}
}

@article{li2025towards,
  title={Towards agentic rag with deep reasoning: A survey of rag-reasoning systems in llms},
  author={Li, Yangning and Zhang, Weizhi and Yang, Yuyao and Huang, Wei-Chieh and Wu, Yaozu and Luo, Junyu and Bei, Yuanchen and Zou, Henry Peng and Luo, Xiao and Zhao, Yusheng and others},
  journal={arXiv preprint arXiv:2507.09477},
  year={2025}
}

@article{arya2025advances,
  title={Advances in Retrieval-Augmented Generation (RAG) and Related Frameworks},
  author={Arya, Sumedha and Gaud, Nirmal},
  journal={IJSAT-International Journal on Science and Technology},
  volume={16},
  number={3},
  year={2025},
  doi={10.71097/IJSAT.V16.I3.7595},
  url={https://doi.org/10.71097/IJSAT.V16.I3.7595},
  publisher={International Research Publication and Journals}
}

@article{hameed2021content,
  title={Content-based image retrieval: A review of recent trends},
  author={Hameed, Ibtihaal M and Abdulhussain, Sadiq H and Mahmmod, Basheera M},
  journal={Cogent Engineering},
  volume={8},
  number={1},
  pages={1927469},
  year={2021},
  doi={10.1080/23311916.2021.1927469},
  url={https://doi.org/10.1080/23311916.2021.1927469},
  publisher={Taylor \& Francis}
}

@article{zhao2023retrieving,
  title={Retrieving multimodal information for augmented generation: A survey},
  author={Zhao, Ruochen and Chen, Hailin and Wang, Weishi and Jiao, Fangkai and Do, Xuan Long and Qin, Chengwei and Ding, Bosheng and Guo, Xiaobao and Li, Minzhi and Li, Xingxuan and others},
  journal={arXiv preprint arXiv:2303.10868},
  year={2023}
}

@inproceedings{sassioui2023visually,
  title={Visually-rich document understanding: concepts, taxonomy and challenges},
  author={Sassioui, Abdellatif and Benouini, Rachid and El Ouargui, Yasser and El-Kamili, Mohamed and Chergui, Meriyem and Ouzzif, Mohammed},
  booktitle={2023 10th International Conference on Wireless Networks and Mobile Communications (WINCOM)},
  pages={1--7},
  year={2023},
  doi={10.1109/WINCOM59760.2023.10322990},
  url={https://doi.org/10.1109/WINCOM59760.2023.10322990},
  organization={IEEE}
}

@article{huang2024pixels,
  title={From pixels to insights: A survey on automatic chart understanding in the era of large foundation models},
  author={Huang, Kung-Hsiang and Chan, Hou Pong and Fung, May and Qiu, Haoyi and Zhou, Mingyang and Joty, Shafiq and Chang, Shih-Fu and Ji, Heng},
  journal={IEEE Transactions on Knowledge and Data Engineering},
  volume={37},
  number={5},
  pages={2550--2568},
  year={2025},
  doi={10.1109/TKDE.2024.3513320},
  url={https://doi.org/10.1109/TKDE.2024.3513320},
  publisher={IEEE}
}

@article{cui2021document,
  title={Document ai: Benchmarks, models and applications},
  author={Cui, Lei and Xu, Yiheng and Lv, Tengchao and Wei, Furu},
  journal={arXiv preprint arXiv:2111.08609},
  year={2021}
}

@article{10.1145/3768156,
author = {Ke, Wenjun and Zheng, Yifan and Li, Yining and Xu, Hengyuan and Nie, Dong and Wang, Peng and He, Yao},
title = {Large Language Models in Document Intelligence: A Comprehensive Survey, Recent Advances, Challenges, and Future Trends},
year = {2026},
issue_date = {January 2026},
publisher = {Association for Computing Machinery},
address = {New York, NY, USA},
volume = {44},
number = {1},
issn = {1046-8188},
url = {https://doi.org/10.1145/3768156},
doi = {10.1145/3768156},
journal = {ACM Trans. Inf. Syst.},
month = nov,
articleno = {18},
numpages = {64}
}

@article{subramani2020survey,
  title={A survey of deep learning approaches for ocr and document understanding},
  author={Subramani, Nishant and Matton, Alexandre and Greaves, Malcolm and Lam, Adrian},
  journal={arXiv preprint arXiv:2011.13534},
  year={2020}
}

@article{zhang2025composed,
  title={Composed multi-modal retrieval: A survey of approaches and applications},
  author={Zhang, Kun and Li, Jingyu and Li, Zhe and Zhang, Jingjing and Li, Fan and Liu, Yandong and Yan, Rui and Jiang, Zihang and Chen, Nan and Zhang, Lei and others},
  journal={arXiv preprint arXiv:2503.01334},
  year={2025}
}

@inproceedings{gao2025scaling,
    title = "Scaling Beyond Context: A Survey of Multimodal Retrieval-Augmented Generation for Document Understanding",
    author = "Gao, Sensen  and
      Zhao, Shanshan  and
      Jiang, Xu  and
      Duan, Lunhao  and
      Chng, Yong Xien  and
      Chen, Qing-Guo  and
      Luo, Weihua  and
      Zhang, Kaifu  and
      Bian, Jia-Wang  and
      Gong, Mingming",
    editor = "Liakata, Maria  and
      Moreira, Viviane P.  and
      Zhang, Jiajun  and
      Jurgens, David",
    booktitle = "Proceedings of the 64th Annual Meeting of the {A}ssociation for {C}omputational {L}inguistics (Volume 1: Long Papers)",
    month = jul,
    year = "2026",
    address = "San Diego, California, United States",
    publisher = "Association for Computational Linguistics",
    url = "https://aclanthology.org/2026.acl-long.204/",
    doi = "10.18653/v1/2026.acl-long.204",
    pages = "4458--4489",
    ISBN = "979-8-89176-390-6"
}

@inproceedings{abootorabi2025ask,
    title = "Ask in Any Modality: A Comprehensive Survey on Multimodal Retrieval-Augmented Generation",
    author = "Abootorabi, Mohammad Mahdi  and
      Zobeiri, Amirhosein  and
      Dehghani, Mahdi  and
      Mohammadkhani, Mohammadali  and
      Mohammadi, Bardia  and
      Ghahroodi, Omid  and
      Baghshah, Mahdieh Soleymani  and
      Asgari, Ehsaneddin",
    editor = "Che, Wanxiang  and
      Nabende, Joyce  and
      Shutova, Ekaterina  and
      Pilehvar, Mohammad Taher",
    booktitle = "Findings of the Association for Computational Linguistics: ACL 2025",
    month = jul,
    year = "2025",
    address = "Vienna, Austria",
    publisher = "Association for Computational Linguistics",
    url = "https://aclanthology.org/2025.findings-acl.861/",
    doi = "10.18653/v1/2025.findings-acl.861",
    pages = "16776--16809",
    ISBN = "979-8-89176-256-5"
}

@article{mei2025survey,
  title={A survey of multimodal retrieval-augmented generation},
  author={Mei, Lang and Mo, Siyu and Yang, Zhihan and Chen, Chong},
  journal={arXiv preprint arXiv:2504.08748},
  year={2025}
}

@article{zheng2025retrieval,
  title={Retrieval augmented generation and understanding in vision: A survey and new outlook},
  author={Zheng, Xu and Weng, Ziqiao and Lyu, Yuanhuiyi and Jiang, Lutao and Xue, Haiwei and Ren, Bin and Paudel, Danda and Sebe, Nicu and Van Gool, Luc and Hu, Xuming},
  journal={arXiv preprint arXiv:2503.18016},
  year={2025}
}

@article{cheng2025survey,
  title={A survey on knowledge-oriented retrieval-augmented generation},
  author={Cheng, Mingyue and Luo, Yucong and Ouyang, Jie and Liu, Qi and Liu, Huijie and Li, Li and Yu, Shuo and Zhang, Bohou and Cao, Jiawei and Ma, Jie and others},
  journal={arXiv preprint arXiv:2503.10677},
  year={2025}
}

@article{gao2023retrieval,
  title={Retrieval-augmented generation for large language models: A survey},
  author={Gao, Yunfan and Xiong, Yun and Gao, Xinyu and Jia, Kangxiang and Pan, Jinliu and Bi, Yuxi and Dai, Yi and Sun, Jiawei and Wang, Meng and Wang, Haofen},
  journal={arXiv preprint arXiv:2312.10997},
  volume={2},
  number={1},
  year={2023}
}

@article{wu2024retrieval,
  title={Retrieval-augmented generation for natural language processing: A survey},
  author={Wu, Shangyu and Xiong, Ying and Cui, Yufei and Wu, Haolun and Chen, Can and Yuan, Ye and Huang, Lianming and Liu, Xue and Kuo, Tei-Wei and Guan, Nan and others},
  journal={arXiv preprint arXiv:2407.13193},
  year={2024}
}

@article{arslan2024survey,
  title={A Survey on RAG with LLMs},
  author={Arslan, Muhammad and Ghanem, Hussam and Munawar, Saba and Cruz, Christophe},
  journal={Procedia computer science},
  volume={246},
  pages={3781--3790},
  year={2024},
  doi={10.1016/j.procs.2024.09.178},
  url={https://doi.org/10.1016/j.procs.2024.09.178},
  publisher={Elsevier}
}

@article{singh2025agentic,
  title={Agentic retrieval-augmented generation: A survey on agentic rag},
  author={Singh, Aditi and Ehtesham, Abul and Kumar, Saket and Khoei, Tala Talaei and Vasilakos, Athanasios V.},
  journal={arXiv preprint arXiv:2501.09136},
  year={2025}
}

@article{gan2025retrieval,
  title={Retrieval Augmented Generation Evaluation in the Era of Large Language Models: A Comprehensive Survey},
  author={Gan, Aoran and Yu, Hao and Zhang, Kai and Liu, Qi and Yan, Wenyu and Huang, Zhenya and Tong, Shiwei and Hu, Guoping},
  journal={arXiv preprint arXiv:2504.14891},
  year={2025}
}

@inproceedings{tang2023unifying,
  title={Unifying vision, text, and layout for universal document processing},
  author={Tang, Zineng and Yang, Ziyi and Wang, Guoxin and Fang, Yuwei and Liu, Yang and Zhu, Chenguang and Zeng, Michael and Zhang, Cha and Bansal, Mohit},
  booktitle={Proceedings of the IEEE/CVF conference on computer vision and pattern recognition},
  pages={19254--19264},
  year={2023},
  doi={10.1109/CVPR52729.2023.01845},
  url={https://doi.org/10.1109/CVPR52729.2023.01845}
}

@inproceedings{li2024enhancing,
  title={Enhancing visual document understanding with contrastive learning in large visual-language models},
  author={Li, Xin and Wu, Yunfei and Jiang, Xinghua and Guo, Zhihao and Gong, Mingming and Cao, Haoyu and Liu, Yinsong and Jiang, Deqiang and Sun, Xing},
  booktitle={Proceedings of the IEEE/CVF Conference on Computer Vision and Pattern Recognition},
  pages={15546--15555},
  year={2024},
  doi={10.1109/CVPR52733.2024.01472},
  url={https://doi.org/10.1109/CVPR52733.2024.01472}
}

@article{song2025bridge,
  title={How to bridge the gap between modalities: Survey on multimodal large language model},
  author={Song, Shezheng and Li, Xiaopeng and Li, Shasha and Zhao, Shan and Yu, Jie and Ma, Jun and Mao, Xiaoguang and Zhang, Weimin and Wang, Meng},
  journal={IEEE Transactions on Knowledge and Data Engineering},
  year={2025},
  doi={10.1109/TKDE.2025.3527978},
  url={https://doi.org/10.1109/TKDE.2025.3527978},
  publisher={IEEE}
}

@article{yan2025position,
  title={Position: Multimodal large language models can significantly advance scientific reasoning},
  author={Yan, Yibo and Wang, Shen and Huo, Jiahao and Ye, Jingheng and Chu, Zhendong and Hu, Xuming and Yu, Philip S and Gomes, Carla and Selman, Bart and Wen, Qingsong},
  journal={arXiv preprint arXiv:2502.02871},
  year={2025}
}

@article{tao2024llms,
  title={Llms are also effective embedding models: An in-depth overview},
  author={Tao, Chongyang and Shen, Tao and Gao, Shen and Zhang, Junshuo and Li, Zhen and Hua, Kai and Hu, Wenpeng and Tao, Zhengwei and Ma, Shuai},
  journal={arXiv preprint arXiv:2412.12591},
  year={2024}
}

@article{zhang2024word,
  title={From word vectors to multimodal embeddings: Techniques, applications, and future directions for large language models},
  author={Zhang, Charles and Peng, Benji and Sun, Xintian and Niu, Qian and Liu, Junyu and Chen, Keyu and Li, Ming and Feng, Pohsun and Bi, Ziqian and Liu, Ming and others},
  journal={arXiv preprint arXiv:2411.05036},
  year={2024}
}

@inproceedings{yan2025survey,
    title = "A Survey of Mathematical Reasoning in the Era of Multimodal Large Language Model: Benchmark, Method {\&} Challenges",
    author = "Yan, Yibo  and
      Su, Jiamin  and
      He, Jianxiang  and
      Fu, Fangteng  and
      Zheng, Xu  and
      Lyu, Yuanhuiyi  and
      Wang, Kun  and
      Wang, Shen  and
      Wen, Qingsong  and
      Hu, Xuming",
    editor = "Che, Wanxiang  and
      Nabende, Joyce  and
      Shutova, Ekaterina  and
      Pilehvar, Mohammad Taher",
    booktitle = "Findings of the Association for Computational Linguistics: ACL 2025",
    month = jul,
    year = "2025",
    address = "Vienna, Austria",
    publisher = "Association for Computational Linguistics",
    url = "https://aclanthology.org/2025.findings-acl.614/",
    doi = "10.18653/v1/2025.findings-acl.614",
    pages = "11798--11827",
    ISBN = "979-8-89176-256-5"
}

@article{wang2024cross,
  title={Cross-modal retrieval: a systematic review of methods and future directions},
  author={Wang, Tianshi and Li, Fengling and Zhu, Lei and Li, Jingjing and Zhang, Zheng and Shen, Heng Tao},
  journal={Proceedings of the IEEE},
  volume={112},
  number={11},
  pages={1716--1754},
  year={2024},
  doi={10.1109/JPROC.2024.3525147},
  url={https://doi.org/10.1109/JPROC.2024.3525147},
  publisher={IEEE}
}

@article{faysse2024colpali,
  title={Colpali: Efficient document retrieval with vision language models},
  author={Faysse, Manuel and Sibille, Hugues and Wu, Tony and Omrani, Bilel and Viaud, Gautier and Hudelot, C{\'e}line and Colombo, Pierre},
  journal={arXiv preprint arXiv:2407.01449},
  year={2024}
}

@article{mace2025vidorev2,
  title={ViDoRe Benchmark V2: Raising the Bar for Visual Retrieval},
  author={Mac{\'e}, Quentin and Loison, Ant{\'o}nio and Faysse, Manuel},
  journal={arXiv preprint arXiv:2505.17166},
  year={2025}
}

@article{yu2024visrag,
  title={Visrag: Vision-based retrieval-augmented generation on multi-modality documents},
  author={Yu, Shi and Tang, Chaoyue and Xu, Bokai and Cui, Junbo and Ran, Junhao and Yan, Yukun and Liu, Zhenghao and Wang, Shuo and Han, Xu and Liu, Zhiyuan and others},
  journal={arXiv preprint arXiv:2410.10594},
  year={2024}
}

@article{sun2025visrag,
  title={{VisRAG2.0}: Mitigating Visual Hallucinations via Evidence-Guided Multi-Image Reasoning in Visual Retrieval-Augmented Generation},
  author={Sun, Yubo and Peng, Chunyi and Yan, Yukun and Yu, Shi and Liu, Zhenghao and Mei, Sen and Chen, Chi and Sun, Maosong},
  journal={arXiv preprint arXiv:2510.09733},
  year={2025}
}

@inproceedings{guo2025towards,
  title={Towards Natural Language-Based Document Image Retrieval: New Dataset and Benchmark},
  author={Guo, Hao and Qin, Xugong and Yang, Jun Jie Ou and Zhang, Peng and Zeng, Gangyan and Li, Yubo and Lin, Hailun},
  booktitle={Proceedings of the Computer Vision and Pattern Recognition Conference},
  pages={29722--29732},
  year={2025},
  doi={10.1109/CVPR52734.2025.02767},
  url={https://doi.org/10.1109/CVPR52734.2025.02767}
}

@article{wang2025vidorag,
  title={Vidorag: Visual document retrieval-augmented generation via dynamic iterative reasoning agents},
  author={Wang, Qiuchen and Ding, Ruixue and Chen, Zehui and Wu, Weiqi and Wang, Shihang and Xie, Pengjun and Zhao, Feng},
  journal={arXiv preprint arXiv:2502.18017},
  year={2025}
}

@article{wasserman2025real,
  title={REAL-MM-RAG: A Real-World Multi-Modal Retrieval Benchmark},
  author={Wasserman, Navve and Pony, Roi and Naparstek, Oshri and Goldfarb, Adi Raz and Schwartz, Eli and Barzelay, Udi and Karlinsky, Leonid},
  journal={arXiv preprint arXiv:2502.12342},
  year={2025}
}

@article{shen2025we,
  title={Are We on the Right Way for Assessing Document Retrieval-Augmented Generation?},
  author={Shen, Wenxuan and Wang, Mingjia and Wang, Yaochen and Chen, Dongping and Yang, Junjie and Wan, Yao and Lin, Weiwei},
  journal={arXiv preprint arXiv:2508.03644},
  year={2025}
}

@article{zhou2025mr,
  title={MR$^2$-Bench: Going Beyond Matching to Reasoning in Multimodal Retrieval},
  author={Zhou, Junjie and Liu, Ze and Xiong, Lei and Yao, Jin-Ge and Wang, Yueze and Xiao, Shitao and Lin, Fenfen and Chen, Miguel Hu and Dou, Zhicheng and Bao, Siqi and others},
  journal={arXiv preprint arXiv:2509.26378},
  year={2025}
}

@article{osmulski2025miracl,
  title={MIRACL-VISION: A Large, multilingual, visual document retrieval benchmark},
  author={Osmulski, Radek and Moreira, Gabriel de Souza P and Ak, Ronay and Xu, Mengyao and Schifferer, Benedikt and Oldridge, Even},
  journal={arXiv preprint arXiv:2505.11651},
  year={2025}
}

@inproceedings{suri2025visdom,
    title = "{V}is{D}o{M}: Multi-Document {QA} with Visually Rich Elements Using Multimodal Retrieval-Augmented Generation",
    author = "Suri, Manan  and
      Mathur, Puneet  and
      Dernoncourt, Franck  and
      Goswami, Kanika  and
      Rossi, Ryan A.  and
      Manocha, Dinesh",
    editor = "Chiruzzo, Luis  and
      Ritter, Alan  and
      Wang, Lu",
    booktitle = "Proceedings of the 2025 Conference of the Nations of the Americas Chapter of the Association for Computational Linguistics: Human Language Technologies (Volume 1: Long Papers)",
    month = apr,
    year = "2025",
    address = "Albuquerque, New Mexico",
    publisher = "Association for Computational Linguistics",
    url = "https://aclanthology.org/2025.naacl-long.310/",
    doi = "10.18653/v1/2025.naacl-long.310",
    pages = "6088--6109",
    ISBN = "979-8-89176-189-6"
}

@inproceedings{zhang2025mrmr,
  title={MRMR: A Realistic and Expert-Level Multidisciplinary Benchmark for Reasoning-Intensive Multimodal Retrieval},
  author={Zhang, Siyue and Gao, Yuan and Zhou, Xiao and Zhao, Yilun and Song, Tingyu and Cohan, Arman and Luu, Anh Tuan and Zhao, Chen},
  booktitle={International Conference on Learning Representations (ICLR)},
  year={2026},
  eprint={2510.09510},
  archivePrefix={arXiv},
  url={https://arxiv.org/abs/2510.09510}
}

@article{kim2025hybrid,
  title={Hybrid-Vector Retrieval for Visually Rich Documents: Combining Single-Vector Efficiency and Multi-Vector Accuracy},
  author={Kim, Juyeon and Lee, Geon and Choi, Dongwon and Kim, Taeuk and Shin, Kijung},
  journal={arXiv preprint arXiv:2510.22215},
  year={2025}
}

@article{dong2025doc,
  title={Doc-Researcher: A Unified System for Multimodal Document Parsing and Deep Research},
  author={Dong, Kuicai and Huang, Shurui and Ye, Fangda and Han, Wei and Zhang, Zhi and Li, Dexun and Li, Wenjun and Yang, Qu and Wang, Gang and Wang, Yichao and others},
  journal={arXiv preprint arXiv:2510.21603},
  year={2025}
}

@article{lee2025sds,
  title={SDS KoPub VDR: A Benchmark Dataset for Visual Document Retrieval in Korean Public Documents},
  author={Lee, Jaehoon and Kim, Sohyun and Park, Wanggeun and Lee, Geon and Kim, Seungkyung and Lee, Minyoung},
  journal={arXiv preprint arXiv:2511.04910},
  year={2025}
}

@article{xiao2025scaling,
  title={Scaling Language-Centric Omnimodal Representation Learning},
  author={Xiao, Chenghao and Chan, Hou Pong and Zhang, Hao and Xu, Weiwen and Aljunied, Mahani and Rong, Yu},
  journal={arXiv preprint arXiv:2510.11693},
  year={2025}
}

@article{kolavi2025m3dr,
  title={M3DR: Towards Universal Multilingual Multimodal Document Retrieval},
  author={Kolavi, Adithya S and Jain, Vyoman},
  journal={arXiv preprint arXiv:2512.03514},
  year={2025}
}

@article{ma2024mmlongbench,
  title={Mmlongbench-doc: Benchmarking long-context document understanding with visualizations},
  author={Ma, Yubo and Zang, Yuhang and Chen, Liangyu and Chen, Meiqi and Jiao, Yizhu and Li, Xinze and Lu, Xinyuan and Liu, Ziyu and Ma, Yan and Dong, Xiaoyi and Zhang, Pan and Pan, Liangming and Jiang, Yu-Gang and Wang, Jiaqi and Cao, Yixin and Sun, Aixin},
  journal={Advances in Neural Information Processing Systems},
  volume={37},
  pages={95963--96010},
  year={2024},
  doi={10.52202/079017-3041},
  url={https://doi.org/10.52202/079017-3041}
}

@article{jiang2024vlm2vec,
  title={Vlm2vec: Training vision-language models for massive multimodal embedding tasks},
  author={Jiang, Ziyan and Meng, Rui and Yang, Xinyi and Yavuz, Semih and Zhou, Yingbo and Chen, Wenhu},
  journal={arXiv preprint arXiv:2410.05160},
  year={2024}
}

@article{meng2025vlm2vec,
  title={Vlm2vec-v2: Advancing multimodal embedding for videos, images, and visual documents},
  author={Meng, Rui and Jiang, Ziyan and Liu, Ye and Su, Mingyi and Yang, Xinyi and Fu, Yuepeng and Qin, Can and Chen, Zeyuan and Xu, Ran and Xiong, Caiming and others},
  journal={arXiv preprint arXiv:2507.04590},
  year={2025}
}

@article{zhang2024gme,
  title={GME: Improving Universal Multimodal Retrieval by Multimodal LLMs},
  author={Zhang, Xin and Zhang, Yanzhao and Xie, Wen and Li, Mingxin and Dai, Ziqi and Long, Dingkun and Xie, Pengjun and Zhang, Meishan and Li, Wenjie and Zhang, Min},
  journal={arXiv preprint arXiv:2412.16855},
  year={2024}
}

@article{lin2024preflmr,
  title={Preflmr: Scaling up fine-grained late-interaction multi-modal retrievers},
  author={Lin, Weizhe and Mei, Jingbiao and Chen, Jinghong and Byrne, Bill},
  journal={arXiv preprint arXiv:2402.08327},
  year={2024}
}

@inproceedings{wei2024uniir,
  title={Uniir: Training and benchmarking universal multimodal information retrievers},
  author={Wei, Cong and Chen, Yang and Chen, Haonan and Hu, Hexiang and Zhang, Ge and Fu, Jie and Ritter, Alan and Chen, Wenhu},
  booktitle={European Conference on Computer Vision},
  pages={387--404},
  year={2024},
  doi={10.1007/978-3-031-73021-4_23},
  url={https://doi.org/10.1007/978-3-031-73021-4_23},
  organization={Springer}
}

@article{xu2025llama,
  title={Llama nemoretriever colembed: Top-performing text-image retrieval model},
  author={Xu, Mengyao and Moreira, Gabriel and Ak, Ronay and Osmulski, Radek and Babakhin, Yauhen and Yu, Zhiding and Schifferer, Benedikt and Oldridge, Even},
  journal={arXiv preprint arXiv:2507.05513},
  year={2025}
}

@article{gunther2025jina,
  title={jina-embeddings-v4: Universal Embeddings for Multimodal Multilingual Retrieval},
  author={G{\"u}nther, Michael and Sturua, Saba and Akram, Mohammad Kalim and Mohr, Isabelle and Ungureanu, Andrei and Wang, Bo and Eslami, Sedigheh and Martens, Scott and Werk, Maximilian and Wang, Nan and others},
  journal={arXiv preprint arXiv:2506.18902},
  year={2025}
}

@article{xiao2025metaembed,
  title={Metaembed: Scaling multimodal retrieval at test-time with flexible late interaction},
  author={Xiao, Zilin and Ma, Qi and Gu, Mengting and Chen, Chun-cheng Jason and Chen, Xintao and Ordonez, Vicente and Mohan, Vijai},
  journal={arXiv preprint arXiv:2509.18095},
  year={2025}
}

@inproceedings{masry2025colflor,
  title={Colflor: Towards Bert-Size Vision-Language Document Retrieval Models},
  author={Masry, Ahmed and Hoque, Enamul},
  booktitle={2025 IEEE 35th International Workshop on Machine Learning for Signal Processing (MLSP)},
  pages={1--5},
  year={2025},
  doi={10.1109/MLSP62443.2025.11204231},
  url={https://doi.org/10.1109/MLSP62443.2025.11204231},
  organization={IEEE}
}

@article{teiletche2025modernvbert,
  title={ModernVBERT: Towards Smaller Visual Document Retrievers},
  author={Teiletche, Paul and Mac{\'e}, Quentin and Conti, Max and Loison, Antonio and Viaud, Gautier and Colombo, Pierre and Faysse, Manuel},
  journal={arXiv preprint arXiv:2510.01149},
  year={2025}
}

@inproceedings{sun2025unveil,
  title={Unveil: Unified Visual-Textual Integration and Distillation for Multi-modal Document Retrieval},
  author={Sun, Hao and Hou, Yingyan and Guo, Jiayan and Wang, Bo and Yang, Chunyu and Ni, Jinsong and Zhang, Yan},
  booktitle={Proceedings of the 63rd Annual Meeting of the Association for Computational Linguistics (Volume 1: Long Papers)},
  pages={23935--23945},
  year={2025},
  publisher={Association for Computational Linguistics},
  doi={10.18653/v1/2025.acl-long.1166},
  url={https://aclanthology.org/2025.acl-long.1166/}
}

@article{nussbaum2024nomic,
  title={Nomic embed vision: Expanding the latent space},
  author={Nussbaum, Zach and Duderstadt, Brandon and Mulyar, Andriy},
  journal={arXiv preprint arXiv:2406.18587},
  year={2024}
}

@article{li2025compression,
  title={Compressing then Matching: An Efficient Pre-training Paradigm for Multimodal Embedding},
  author={Li, Da and Luo, Yuxiao and Bi, Keping and Guo, Jiafeng and Yuan, Wei and Yang, Biao and Wang, Yan and Yang, Fan and Gao, Tingting and Zhou, Guorui},
  journal={arXiv preprint arXiv:2511.08480},
  year={2025}
}

@article{liu2025reasoning,
  title={Reasoning Guided Embeddings: Leveraging MLLM Reasoning for Improved Multimodal Retrieval},
  author={Liu, Chunxu and Yang, Jiyuan and Gao, Ruopeng and Zhu, Yuhan and Zhu, Feng and Zhao, Rui and Wang, Limin},
  journal={arXiv preprint arXiv:2511.16150},
  year={2025}
}

@inproceedings{ouali2025vladva,
  title={VladVA: Discriminative Fine-tuning of LVLMs},
  author={Ouali, Yassine and Bulat, Adrian and Xenos, Alexandros and Zaganidis, Anestis and Metaxas, Ioannis Maniadis and Martinez, Brais and Tzimiropoulos, Georgios},
  booktitle={Proceedings of the Computer Vision and Pattern Recognition Conference},
  pages={4101--4111},
  year={2025},
  doi={10.1109/CVPR52734.2025.00388},
  url={https://doi.org/10.1109/CVPR52734.2025.00388}
}

@inproceedings{liu2025lamra,
  title={Lamra: Large multimodal model as your advanced retrieval assistant},
  author={Liu, Yikun and Zhang, Yajie and Cai, Jiayin and Jiang, Xiaolong and Hu, Yao and Yao, Jiangchao and Wang, Yanfeng and Xie, Weidi},
  booktitle={Proceedings of the Computer Vision and Pattern Recognition Conference},
  pages={4015--4025},
  year={2025},
  doi={10.1109/CVPR52734.2025.00380},
  url={https://doi.org/10.1109/CVPR52734.2025.00380}
}

@article{thirukovalluru2025breaking,
  title={Breaking the Batch Barrier (B3) of Contrastive Learning via Smart Batch Mining},
  author={Thirukovalluru, Raghuveer and Meng, Rui and Liu, Ye and K, Karthikeyan and Su, Mingyi and Nie, Ping and Yavuz, Semih and Zhou, Yingbo and Chen, Wenhu and Dhingra, Bhuwan},
  journal={arXiv preprint arXiv:2505.11293},
  year={2025}
}

@article{lan2025ume,
  title={UME-R1: Exploring Reasoning-Driven Generative Multimodal Embeddings},
  author={Lan, Zhibin and Niu, Liqiang and Meng, Fandong and Zhou, Jie and Su, Jinsong},
  journal={arXiv preprint arXiv:2511.00405},
  year={2025}
}

@article{yu2025cafe,
  title={CAFe: Unifying Representation and Generation with Contrastive-Autoregressive Finetuning},
  author={Yu, Hao and Zhao, Zhuokai and Yan, Shen and Korycki, Lukasz and Wang, Jianyu and He, Baosheng and Liu, Jiayi and Zhang, Lizhu and Fan, Xiangjun and Yu, Hanchao},
  journal={arXiv preprint arXiv:2503.19900},
  year={2025}
}

@article{cui2025think,
  title={Think then embed: Generative context improves multimodal embedding},
  author={Cui, Xuanming and Cheng, Jianpeng and Chen, Hong-you and Shukla, Satya Narayan and Awasthi, Abhijeet and Pan, Xichen and Ahuja, Chaitanya and Mishra, Shlok Kumar and Yang, Yonghuan and Xiao, Jun and others},
  journal={arXiv preprint arXiv:2510.05014},
  year={2025}
}

@article{chen2025mme5,
  title={mme5: Improving multimodal multilingual embeddings via high-quality synthetic data},
  author={Chen, Haonan and Wang, Liang and Yang, Nan and Zhu, Yutao and Zhao, Ziliang and Wei, Furu and Dou, Zhicheng},
  journal={arXiv preprint arXiv:2502.08468},
  year={2025}
}

@article{lin2024mm,
  title={Mm-embed: Universal multimodal retrieval with multimodal llms},
  author={Lin, Sheng-Chieh and Lee, Chankyu and Shoeybi, Mohammad and Lin, Jimmy and Catanzaro, Bryan and Ping, Wei},
  journal={arXiv preprint arXiv:2411.02571},
  year={2024}
}

@article{chen2025moca,
  title={MoCa: Modality-aware Continual Pre-training Makes Better Bidirectional Multimodal Embeddings},
  author={Chen, Haonan and Liu, Hong and Luo, Yuping and Wang, Liang and Yang, Nan and Wei, Furu and Dou, Zhicheng},
  journal={arXiv preprint arXiv:2506.23115},
  year={2025}
}

@inproceedings{zhou2025megapairs,
  title={Megapairs: Massive data synthesis for universal multimodal retrieval},
  author={Zhou, Junjie and Xiong, Yongping and Liu, Zheng and Liu, Ze and Xiao, Shitao and Wang, Yueze and Zhao, Bo and Zhang, Chen Jason and Lian, Defu},
  booktitle={Proceedings of the 63rd Annual Meeting of the Association for Computational Linguistics (Volume 1: Long Papers)},
  pages={19076--19095},
  year={2025},
  publisher={Association for Computational Linguistics},
  doi={10.18653/v1/2025.acl-long.935},
  url={https://aclanthology.org/2025.acl-long.935/}
}

@article{zhu2025freeret,
  title={FreeRet: MLLMs as Training-Free Retrievers},
  author={Zhu, Yuhan and Zeng, Xiangyu and Wang, Chenting and Li, Xinhao and Liu, Chunxu and Xu, Yicheng and Yan, Ziang and Wang, Yi and Wang, Limin},
  journal={arXiv preprint arXiv:2509.24621},
  year={2025}
}

@article{jiang2024e5,
  title={E5-v: Universal embeddings with multimodal large language models},
  author={Jiang, Ting and Song, Minghui and Zhang, Zihan and Huang, Haizhen and Deng, Weiwei and Sun, Feng and Zhang, Qi and Wang, Deqing and Zhuang, Fuzhen},
  journal={arXiv preprint arXiv:2407.12580},
  year={2024}
}

@article{zhu2025retrvr1,
  title={Retrv-R1: A Reasoning-Driven MLLM Framework for Universal and Efficient Multimodal Retrieval},
  author={Zhu, Lanyun and Ji, Deyi and Chen, Tianrun and Wu, Haiyang and Wang, Shiqi},
  journal={Advances in Neural Information Processing Systems},
  year={2025},
  doi={10.52202/085713-5719},
  url={https://doi.org/10.52202/085713-5719}
}

@misc{gu2025unimev2,
      title={UniME-V2: MLLM-as-a-Judge for Universal Multimodal Embedding Learning}, 
      author={Tiancheng Gu and Kaicheng Yang and Kaichen Zhang and Xiang An and Ziyong Feng and Yueyi Zhang and Weidong Cai and Jiankang Deng and Lidong Bing},
      year={2025},
      eprint={2510.13515},
      archivePrefix={arXiv},
      primaryClass={cs.CV},
      url={https://arxiv.org/abs/2510.13515}, 
}

@misc{JinaAI2025rerankerm0,
  author       = {{Jina AI}},
  title        = {{jina-reranker-m0: Multilingual Multimodal Document Reranker}},
  year         = {2025},
  organization = {Jina AI},
  howpublished = {Hugging Face model card},
  url          = {https://huggingface.co/jinaai/jina-reranker-m0}
}

@article{wasserman2025docrerank,
  title={DocReRank: Single-Page Hard Negative Query Generation for Training Multi-Modal RAG Rerankers},
  author={Wasserman, Navve and Heinimann, Oliver and Golbari, Yuval and Zimbalist, Tal and Schwartz, Eli and Irani, Michal},
  journal={arXiv preprint arXiv:2505.22584},
  year={2025}
}

@article{chen2024mllm,
  title={Mllm is a strong reranker: Advancing multimodal retrieval-augmented generation via knowledge-enhanced reranking and noise-injected training},
  author={Chen, Zhanpeng and Xu, Chengjin and Qi, Yiyan and Guo, Jian},
  journal={arXiv preprint arXiv:2407.21439},
  year={2024}
}

@article{dai2025supervised,
  title={Supervised Fine-Tuning or Contrastive Learning? Towards Better Multimodal LLM Reranking},
  author={Dai, Ziqi and Zhang, Xin and Li, Mingxin and Zhang, Yanzhao and Long, Dingkun and Xie, Pengjun and Zhang, Meishan and Li, Wenjie and Zhang, Min},
  journal={arXiv preprint arXiv:2510.14824},
  year={2025}
}

@article{xu2025mm,
  title={MM-R5: MultiModal Reasoning-Enhanced ReRanker via Reinforcement Learning for Document Retrieval},
  author={Xu, Mingjun and Dong, Jinhan and Hou, Jue and Wang, Zehui and Li, Sihang and Gao, Zhifeng and Zhong, Renxin and Cai, Hengxing},
  journal={arXiv preprint arXiv:2506.12364},
  year={2025}
}

@misc{MonoQwen,
  title={MonoQwen: Visual Document Reranking},
  author={Chaffin, Antoine and Lac, Aurélien},
  year={2024},
  howpublished={Hugging Face model card},
  url={https://huggingface.co/lightonai/MonoQwen2-VL-v0.1}
}

@article{wang2025vragrl,
  title={VRAG-RL: Empower Vision-Perception-Based RAG for Visually Rich Information Understanding via Iterative Reasoning with Reinforcement Learning},
  author={Wang, Qiuchen and Ding, Ruixue and Zeng, Yu and Chen, Zehui and Chen, Lin and Wang, Shihang and Xie, Pengjun and Huang, Fei and Zhao, Feng},
  journal={arXiv preprint arXiv:2505.22019},
  year={2025}
}

@article{xie2025textlessrag,
  title={TextlessRAG: End-to-End Visual Document RAG by Speech Without Text},
  author={Xie, Peijin and Qian, Shun and Liu, Bingquan and Wang, Dexin and Sun, Lin and Zhang, Xiangzheng},
  journal={arXiv preprint arXiv:2509.07538},
  year={2025}
}

@article{cho2024m3docrag,
  title={M3docrag: Multi-modal retrieval is what you need for multi-page multi-document understanding},
  author={Cho, Jaemin and Mahata, Debanjan and Irsoy, Ozan and He, Yujie and Bansal, Mohit},
  journal={arXiv preprint arXiv:2411.04952},
  year={2024}
}

@article{han2025mdocagent,
  title={Mdocagent: A multi-modal multi-agent framework for document understanding},
  author={Han, Siwei and Xia, Peng and Zhang, Ruiyi and Sun, Tong and Li, Yun and Zhu, Hongtu and Yao, Huaxiu},
  journal={arXiv preprint arXiv:2503.13964},
  year={2025}
}

@article{tong2025hkrag,
  title={HKRAG: Holistic Knowledge Retrieval-Augmented Generation Over Visually-Rich Documents},
  author={Tong, Anyang and Niu, Xiang and Liu, ZhiPing and Tian, Chang and Wei, Yanyan and Shi, Zenglin and Wang, Meng},
  journal={arXiv preprint arXiv:2511.20227},
  year={2025}
}

@article{yu2025visual,
  title={Visual Document Understanding and Reasoning: A Multi-Agent Collaboration Framework with Agent-Wise Adaptive Test-Time Scaling},
  author={Yu, Xinlei and Xu, Chengming and Chen, Zhangquan and Zhang, Yudong and Lu, Shilin and Yang, Cheng and Zhang, Jiangning and Yan, Shuicheng and Hu, Xiaobin},
  journal={arXiv preprint arXiv:2508.03404},
  year={2025}
}

@inproceedings{wu2025molorag,
    title = "{M}o{L}o{RAG}: Bootstrapping Document Understanding via Multi-modal Logic-aware Retrieval",
    author = "Wu, Xixi  and
      Tan, Yanchao  and
      Hou, Nan  and
      Zhang, Ruiyang  and
      Cheng, Hong",
    editor = "Christodoulopoulos, Christos  and
      Chakraborty, Tanmoy  and
      Rose, Carolyn  and
      Peng, Violet",
    booktitle = "Proceedings of the 2025 Conference on Empirical Methods in Natural Language Processing",
    month = nov,
    year = "2025",
    address = "Suzhou, China",
    publisher = "Association for Computational Linguistics",
    url = "https://aclanthology.org/2025.emnlp-main.708/",
    doi = "10.18653/v1/2025.emnlp-main.708",
    pages = "14024--14045",
    ISBN = "979-8-89176-332-6"
}

@article{zhu2025doclens,
  title={Doclens: A tool-augmented multi-agent framework for long visual document understanding},
  author={Zhu, Dawei and Meng, Rui and Chen, Jiefeng and Li, Sujian and Pfister, Tomas and Yoon, Jinsung},
  journal={arXiv preprint arXiv:2511.11552},
  year={2025}
}

@article{gong2025mhier,
  title={MHier-RAG: Multi-Modal RAG for Visual-Rich Document Question-Answering via Hierarchical and Multi-Granularity Reasoning},
  author={Gong, Ziyu and Mai, Chengcheng and Huang, Yihua},
  journal={arXiv preprint arXiv:2508.00579},
  year={2025}
}

@article{sourati2025lad,
  title={LAD-RAG: Layout-aware Dynamic RAG for Visually-Rich Document Understanding},
  author={Sourati, Zhivar and Wang, Zheng and Liu, Marianne Menglin and Hu, Yazhe and Guo, Mengqing and Bharadwaj, Sujeeth and Han, Kyu and Sheng, Tao and Ravi, Sujith and Dehghani, Morteza and others},
  journal={arXiv preprint arXiv:2510.07233},
  year={2025}
}

@article{liu2025resolving,
  title={Resolving Evidence Sparsity: Agentic Context Engineering for Long-Document Understanding},
  author={Liu, Keliang and Chen, Zizhi and Li, Mingcheng and Tang, Jingqun and Yang, Dingkang and Zhang, Lihua},
  journal={arXiv preprint arXiv:2511.22850},
  year={2025}
}

@article{ashraf2025agent,
  title={Agent-X: Evaluating Deep Multimodal Reasoning in Vision-Centric Agentic Tasks},
  author={Ashraf, Tajamul and Saqib, Amal and Ghani, Hanan and AlMahri, Muhra and Li, Yuhao and Ahsan, Noor and Nawaz, Umair and Lahoud, Jean and Cholakkal, Hisham and Shah, Mubarak and others},
  journal={arXiv preprint arXiv:2505.24876},
  year={2025}
}

@article{zhang2025see,
  title={See, Think, Act: Online Shopper Behavior Simulation with VLM Agents},
  author={Zhang, Yimeng and Gesi, Jiri and Xue, Ran and Wang, Tian and Wang, Ziyi and Lu, Yuxuan and Zhan, Sinong and Zeng, Huimin and Cui, Qingjun and Guo, Yufan and others},
  journal={arXiv preprint arXiv:2510.19245},
  year={2025}
}

@article{yan2025mathagent,
  title={Mathagent: Leveraging a mixture-of-math-agent framework for real-world multimodal mathematical error detection},
  author={Yan, Yibo and Wang, Shen and Huo, Jiahao and Yu, Philip S and Hu, Xuming and Wen, Qingsong},
  journal={arXiv preprint arXiv:2503.18132},
  year={2025}
}

@article{su2025cafes,
  title={CAFES: A Collaborative Multi-Agent Framework for Multi-Granular Multimodal Essay Scoring},
  author={Su, Jiamin and Yan, Yibo and Gao, Zhuoran and Zhang, Han and Liu, Xiang and Hu, Xuming},
  journal={arXiv preprint arXiv:2505.13965},
  year={2025}
}

@article{cao2025toward,
  title={Toward generalizable evaluation in the llm era: A survey beyond benchmarks},
  author={Cao, Yixin and Hong, Shibo and Li, Xinze and Ying, Jiahao and Ma, Yubo and Liang, Haiyuan and Liu, Yantao and Yao, Zijun and Wang, Xiaozhi and Huang, Dan and others},
  journal={arXiv preprint arXiv:2504.18838},
  year={2025}
}

@inproceedings{zhang2025ocr,
  title={Ocr hinders rag: Evaluating the cascading impact of ocr on retrieval-augmented generation},
  author={Zhang, Junyuan and Zhang, Qintong and Wang, Bin and Ouyang, Linke and Wen, Zichen and Li, Ying and Chow, Ka-Ho and He, Conghui and Zhang, Wentao},
  booktitle={Proceedings of the IEEE/CVF International Conference on Computer Vision},
  pages={17443--17453},
  year={2025},
  doi={10.1109/ICCV51701.2025.01620},
  url={https://doi.org/10.1109/ICCV51701.2025.01620}
}

@inproceedings{most2025lost,
  title={Lost in ocr translation? vision-based approaches to robust document retrieval},
  author={Most, Alexander and Winjum, Joseph and Bhattarai, Manish and Jones, Shawn M. and Ranasinghe, Nishath Rajiv and Biswas, Ayan and O'Malley, Dan},
  booktitle={Proceedings of the 2025 ACM Symposium on Document Engineering},
  pages={13:1--13:10},
  year={2025},
  doi={10.1145/3704268.3742698},
  url={https://doi.org/10.1145/3704268.3742698}
}

@article{enevoldsen2025mmteb,
  title={Mmteb: Massive multilingual text embedding benchmark},
  author={Enevoldsen, Kenneth and Chung, Isaac and Kerboua, Imene and Kardos, M{\'a}rton and Mathur, Ashwin and Stap, David and Gala, Jay and Siblini, Wissam and Krzemi{\'n}ski, Dominik and Winata, Genta Indra and others},
  journal={arXiv preprint arXiv:2502.13595},
  year={2025}
}

@article{zhang2023miracl,
  title={Miracl: A multilingual retrieval dataset covering 18 diverse languages},
  author={Zhang, Xinyu and Thakur, Nandan and Ogundepo, Odunayo and Kamalloo, Ehsan and Alfonso-Hermelo, David and Li, Xiaoguang and Liu, Qun and Rezagholizadeh, Mehdi and Lin, Jimmy},
  journal={Transactions of the Association for Computational Linguistics},
  volume={11},
  pages={1114--1131},
  year={2023},
  doi={10.1162/tacl_a_00595},
  url={https://doi.org/10.1162/tacl_a_00595},
  publisher={MIT Press}
}

@article{su2024bright,
  title={Bright: A realistic and challenging benchmark for reasoning-intensive retrieval},
  author={Su, Hongjin and Yen, Howard and Xia, Mengzhou and Shi, Weijia and Muennighoff, Niklas and Wang, Han-yu and Liu, Haisu and Shi, Quan and Siegel, Zachary S and Tang, Michael and others},
  journal={arXiv preprint arXiv:2407.12883},
  year={2024}
}

@article{xiao2024rar,
  title={Rar-b: Reasoning as retrieval benchmark},
  author={Xiao, Chenghao and Hudson, G Thomas and Moubayed, Noura Al},
  journal={arXiv preprint arXiv:2404.06347},
  year={2024}
}

@article{shao2025reasonir,
  title={ReasonIR: Training Retrievers for Reasoning Tasks},
  author={Shao, Rulin and Qiao, Rui and Kishore, Varsha and Muennighoff, Niklas and Lin, Xi Victoria and Rus, Daniela and Low, Bryan Kian Hsiang and Min, Sewon and Yih, Wen-tau and Koh, Pang Wei and others},
  journal={arXiv preprint arXiv:2504.20595},
  year={2025}
}

@article{han2025ateb,
  title={ATEB: Evaluating and Improving Advanced NLP Tasks for Text Embedding Models},
  author={Han, Simeng and Gomez, Frank Palma and Vu, Tu and Li, Zefei and Cer, Daniel and Zeng, Hansi and Tar, Chris and Cohan, Arman and Abrego, Gustavo Hernandez},
  journal={arXiv preprint arXiv:2502.16766},
  year={2025}
}

@inproceedings{liu2025exploring,
  title={Exploring Reasoning-Infused Text Embedding with Large Language Models for Zero-Shot Dense Retrieval},
  author={Liu, Yuxiang and Wang, Tian and Kundu, Gourab and Cao, Tianyu and Cheng, Guang and Ge, Zhen and Chen, Jianshu and Cui, Qingjun and Chilimbi, Trishul},
  booktitle={Proceedings of the 34th ACM International Conference on Information and Knowledge Management},
  pages={4981--4985},
  year={2025},
  doi={10.1145/3746252.3760855},
  url={https://doi.org/10.1145/3746252.3760855}
}

@inproceedings{joshi2024reaper,
  title={Reaper: Reasoning based retrieval planning for complex rag systems},
  author={Joshi, Ashutosh and Sarwar, Sheikh Muhammad and Varshney, Samarth and Nag, Sreyashi and Agrawal, Shrivats and Naik, Juhi},
  booktitle={Proceedings of the 33rd ACM International Conference on Information and Knowledge Management},
  pages={4621--4628},
  year={2024},
  doi={10.1145/3627673.3680087},
  url={https://doi.org/10.1145/3627673.3680087}
}

@article{li2025r2med,
  title={R2MED: A Benchmark for Reasoning-Driven Medical Retrieval},
  author={Zhang, Xiangxu and Li, Lei and Zhou, Xiao and Liu, Zheng},
  journal={arXiv preprint arXiv:2505.14558},
  year={2025}
}

@article{chen2025reasonembed,
  title={ReasonEmbed: Enhanced Text Embeddings for Reasoning-Intensive Document Retrieval},
  author={Chen, Jianlyu and Lan, Junwei and Li, Chaofan and Lian, Defu and Liu, Zheng},
  journal={arXiv preprint arXiv:2510.08252},
  year={2025}
}

@inproceedings{das2025rader,
    title = "{R}a{D}e{R}: Reasoning-aware Dense Retrieval Models",
    author = "Das, Debrup  and
      O{'}Nuallain, Sam  and
      Rahimi, Razieh",
    editor = "Christodoulopoulos, Christos  and
      Chakraborty, Tanmoy  and
      Rose, Carolyn  and
      Peng, Violet",
    booktitle = "Proceedings of the 2025 Conference on Empirical Methods in Natural Language Processing",
    month = nov,
    year = "2025",
    address = "Suzhou, China",
    publisher = "Association for Computational Linguistics",
    url = "https://aclanthology.org/2025.emnlp-main.1011/",
    doi = "10.18653/v1/2025.emnlp-main.1011",
    pages = "19970--19997",
    ISBN = "979-8-89176-332-6"
}

@article{long2025diver,
  title={Diver: A multi-stage approach for reasoning-intensive information retrieval},
  author={Sun, Duolin and Long, Meixiu and Yang, Dan and Wang, Junjie and Luo, Yecheng and Shen, Yue and Wang, Jian and Zhou, Hualei and Guo, Chunxiao and Wei, Peng and Wang, Jiahai and Gu, Jinjie},
  journal={arXiv preprint arXiv:2508.07995},
  year={2025}
}

@article{yan2024errorradar,
  title={Errorradar: Benchmarking complex mathematical reasoning of multimodal large language models via error detection},
  author={Yan, Yibo and Wang, Shen and Huo, Jiahao and Li, Hang and Li, Boyan and Su, Jiamin and Gao, Xiong and Zhang, Yi-Fan and Xu, Tianlong and Chu, Zhendong and others},
  journal={arXiv preprint arXiv:2410.04509},
  year={2024}
}

@article{bi2025reasoning,
  title={Why reasoning matters? a survey of advancements in multimodal reasoning (v1)},
  author={Bi, Jing and Liang, Susan and Zhou, Xiaofei and Liu, Pinxin and Guo, Junjia and Tang, Yunlong and Song, Luchuan and Huang, Chao and Vosoughi, Ali and Sun, Guangyu and others},
  journal={arXiv preprint arXiv:2504.03151},
  year={2025}
}

@article{su2025thinking,
  title={Thinking with images for multimodal reasoning: Foundations, methods, and future frontiers},
  author={Su, Zhaochen and Xia, Peng and Guo, Hangyu and Liu, Zhenhua and Ma, Yan and Qu, Xiaoye and Liu, Jiaqi and Li, Yanshu and Zeng, Kaide and Yang, Zhengyuan and others},
  journal={arXiv preprint arXiv:2506.23918},
  year={2025}
}

@article{lin2025mind,
  title={Mind with eyes: from language reasoning to multimodal reasoning},
  author={Lin, Zhiyu and Gao, Yifei and Zhao, Xian and Yang, Yunfan and Sang, Jitao},
  journal={arXiv preprint arXiv:2503.18071},
  year={2025}
}

@inproceedings{duan2025docopilot,
  title={Docopilot: Improving Multimodal Models for Document-Level Understanding},
  author={Duan, Yuchen and Chen, Zhe and Hu, Yusong and Wang, Weiyun and Ye, Shenglong and Shi, Botian and Lu, Lewei and Hou, Qibin and Lu, Tong and Li, Hongsheng and others},
  booktitle={Proceedings of the Computer Vision and Pattern Recognition Conference},
  pages={4026--4037},
  year={2025},
  doi={10.1109/CVPR52734.2025.00381},
  url={https://doi.org/10.1109/CVPR52734.2025.00381}
}

@inproceedings{liao2025doclayllm,
  title={DocLayLLM: An Efficient Multi-modal Extension of Large Language Models for Text-rich Document Understanding},
  author={Liao, Wenhui and Wang, Jiapeng and Li, Hongliang and Wang, Chengyu and Huang, Jun and Jin, Lianwen},
  booktitle={Proceedings of the Computer Vision and Pattern Recognition Conference},
  pages={4038--4049},
  year={2025},
  doi={10.1109/CVPR52734.2025.00382},
  url={https://doi.org/10.1109/CVPR52734.2025.00382}
}

@inproceedings{nacson2025docvlm,
  title={Docvlm: Make your vlm an efficient reader},
  author={Nacson, Mor Shpigel and Aberdam, Aviad and Ganz, Roy and Ben-Avraham, Elad and Golts, Alona and Kittenplon, Yair and Mazor, Shai and Litman, Ron},
  booktitle={Proceedings of the Computer Vision and Pattern Recognition Conference},
  pages={29005--29015},
  year={2025},
  doi={10.1109/CVPR52734.2025.02701},
  url={https://doi.org/10.1109/CVPR52734.2025.02701}
}

@inproceedings{jiang2025docrelation,
  title={Relation-Rich Visual Document Generator for Visual Information Extraction},
  author={Jiang, Zi-Han and Lin, Chien-Wei and Li, Wei-Hua and Liu, Hsuan-Tung and Yeh, Yi-Ren and Chen, Chu-Song},
  booktitle={Proceedings of the Computer Vision and Pattern Recognition Conference},
  pages={14449--14459},
  year={2025},
  doi={10.1109/CVPR52734.2025.01347},
  url={https://doi.org/10.1109/CVPR52734.2025.01347}
}

@article{zhang2025imprag,
  title={ImpRAG: Retrieval-Augmented Generation with Implicit Queries},
  author={Zhang, Wenzheng and Lin, Xi Victoria and Stratos, Karl and Yih, Wen-tau and Chen, Mingda},
  journal={arXiv preprint arXiv:2506.02279},
  year={2025}
}

@article{wei2024instructrag,
  title={Instructrag: Instructing retrieval-augmented generation via self-synthesized rationales},
  author={Wei, Zhepei and Chen, Wei-Lin and Meng, Yu},
  journal={arXiv preprint arXiv:2406.13629},
  year={2024}
}

@inproceedings{zhu2024retrieval,
  title={Retrieval-augmented embodied agents},
  author={Zhu, Yichen and Ou, Zhicai and Mou, Xiaofeng and Tang, Jian},
  booktitle={Proceedings of the IEEE/CVF Conference on Computer Vision and Pattern Recognition},
  pages={17985--17995},
  year={2024},
  doi={10.1109/CVPR52733.2024.01703},
  url={https://doi.org/10.1109/CVPR52733.2024.01703}
}

@inproceedings{liu2025hm,
  title={Hm-rag: Hierarchical multi-agent multimodal retrieval augmented generation},
  author={Liu, Pei and Liu, Xin and Yao, Ruoyu and Liu, Junming and Meng, Siyuan and Wang, Ding and Ma, Jun},
  booktitle={Proceedings of the 33rd ACM International Conference on Multimedia},
  pages={2781--2790},
  year={2025},
  doi={10.1145/3746027.3754761},
  url={https://doi.org/10.1145/3746027.3754761}
}

@article{zhang2025toward,
  title={Toward agentic ai: generative information retrieval inspired intelligent communications and networking},
  author={Zhang, Ruichen and Tang, Shunpu and Liu, Yinqiu and Niyato, Dusit and Xiong, Zehui and Sun, Sumei and Mao, Shiwen and Han, Zhu},
  journal={arXiv preprint arXiv:2502.16866},
  year={2025}
}

@article{ji2024reasoningrank,
  title={Reasoningrank: Teaching student models to rank through reasoning-based knowledge distillation},
  author={Ji, Yuelyu and Li, Zhuochun and Meng, Rui and He, Daqing},
  journal={arXiv preprint arXiv:2410.05168},
  year={2024}
}

@inproceedings{samarinas2025distillation,
  title={Distillation and refinement of reasoning in small language models for document re-ranking},
  author={Samarinas, Chris and Zamani, Hamed},
  booktitle={Proceedings of the 2025 International ACM SIGIR Conference on Innovative Concepts and Theories in Information Retrieval (ICTIR)},
  pages={430--435},
  year={2025},
  doi={10.1145/3731120.3744613},
  url={https://doi.org/10.1145/3731120.3744613}
}

@article{weller2025rank1,
  title={Rank1: Test-time compute for reranking in information retrieval},
  author={Weller, Orion and Ricci, Kathryn and Yang, Eugene and Yates, Andrew and Lawrie, Dawn and Van Durme, Benjamin},
  journal={arXiv preprint arXiv:2502.18418},
  year={2025}
}

@article{fan2025tfrank,
  title={TFRank: Think-Free Reasoning Enables Practical Pointwise LLM Ranking},
  author={Fan, Yongqi and Chen, Xiaoyang and Ye, Dezhi and Liu, Jie and Liang, Haijin and Ma, Jin and He, Ben and Sun, Yingfei and Ruan, Tong},
  journal={arXiv preprint arXiv:2508.09539},
  year={2025}
}

@article{yang2025rank,
  title={Rank-k: Test-time reasoning for listwise reranking},
  author={Yang, Eugene and Yates, Andrew and Ricci, Kathryn and Weller, Orion and Chari, Vivek and Van Durme, Benjamin and Lawrie, Dawn},
  journal={arXiv preprint arXiv:2505.14432},
  year={2025}
}

@article{huang2025contextual,
  title={Contextual Relevance and Adaptive Sampling for LLM-Based Document Reranking},
  author={Huang, Jerry and Madala, Siddarth and Niu, Cheng and Hockenmaier, Julia and Zhang, Tong},
  journal={arXiv preprint arXiv:2511.01208},
  year={2025}
}

@article{seetharaman2025insertrank,
  title={InsertRank: LLMs can reason over BM25 scores to Improve Listwise Reranking},
  author={Seetharaman, Rahul and Dhole, Kaustubh D and Bansal, Aman},
  journal={arXiv preprint arXiv:2506.14086},
  year={2025}
}

@article{xu2025beyond,
  title={Beyond Sequential Reranking: Reranker-Guided Search Improves Reasoning Intensive Retrieval},
  author={Xu, Haike and Chen, Tong},
  journal={arXiv preprint arXiv:2509.07163},
  year={2025}
}

@misc{li2024coircomprehensivebenchmarkcode,
  archiveprefix = {arXiv},
  author = {Li, Xiangyang and Dong, Kuicai and Lee, Yi Quan and Xia, Wei and Zhang, Hao and Dai, Xinyi and Wang, Yasheng and Tang, Ruiming},
  eprint = {2407.02883},
  primaryclass = {cs.IR},
  title = {CoIR: A Comprehensive Benchmark for Code Information Retrieval Models},
  url = {https://arxiv.org/abs/2407.02883},
  year = {2024},
}

@misc{wang2024coderagbenchretrievalaugmentcode,
  archiveprefix = {arXiv},
  author = {Zora Zhiruo Wang and Akari Asai and Xinyan Velocity Yu and Frank F. Xu and Yiqing Xie and Graham Neubig and Daniel Fried},
  eprint = {2406.14497},
  primaryclass = {cs.SE},
  title = {CodeRAG-Bench: Can Retrieval Augment Code Generation?},
  url = {https://arxiv.org/abs/2406.14497},
  year = {2024},
}

@misc{weller2024followir,
  archiveprefix = {arXiv},
  author = {Orion Weller and Benjamin Chang and Sean MacAvaney and Kyle Lo and Arman Cohan and Benjamin Van Durme and Dawn Lawrie and Luca Soldaini},
  eprint = {2403.15246},
  primaryclass = {cs.IR},
  title = {FollowIR: Evaluating and Teaching Information Retrieval Models to Follow Instructions},
  year = {2024},
}

@article{zhu2024longembed,
  author = {Zhu, Dawei and Wang, Liang and Yang, Nan and Song, Yifan and Wu, Wenhao and Wei, Furu and Li, Sujian},
  journal = {arXiv preprint arXiv:2404.12096},
  title = {LongEmbed: Extending Embedding Models for Long Context Retrieval},
  year = {2024},
}

@inproceedings{weller2025mfollowir,
  title={mFollowIR: A Multilingual Benchmark for Instruction Following in Retrieval},
  author={Weller, Orion and Chang, Benjamin and Yang, Eugene and Yarmohammadi, Mahsa and Barham, Samuel and MacAvaney, Sean and Cohan, Arman and Soldaini, Luca and Van Durme, Benjamin and Lawrie, Dawn J.},
  booktitle={European Conference on Information Retrieval},
  pages={295--310},
  year={2025},
  doi={10.1007/978-3-031-88711-6_19},
  url={https://doi.org/10.1007/978-3-031-88711-6_19},
  organization={Springer}
}

@inproceedings{mathew2021docvqa,
  title={Docvqa: A dataset for vqa on document images},
  author={Mathew, Minesh and Karatzas, Dimosthenis and Jawahar, C. V.},
  booktitle={Proceedings of the IEEE/CVF winter conference on applications of computer vision},
  pages={2199--2208},
  year={2021},
  doi={10.1109/WACV48630.2021.00225},
  url={https://doi.org/10.1109/WACV48630.2021.00225}
}

@inproceedings{mathew2022infographicvqa,
  title={Infographicvqa},
  author={Mathew, Minesh and Bagal, Viraj and Tito, Rubèn and Karatzas, Dimosthenis and Valveny, Ernest and Jawahar, C. V.},
  booktitle={Proceedings of the IEEE/CVF Winter Conference on Applications of Computer Vision},
  pages={2582--2591},
  year={2022},
  doi={10.1109/WACV51458.2022.00264},
  url={https://doi.org/10.1109/WACV51458.2022.00264}
}

@inproceedings{li2024multimodalarxiv,
    title = "Multimodal {A}r{X}iv: A Dataset for Improving Scientific Comprehension of Large Vision-Language Models",
    author = "Li, Lei  and
      Wang, Yuqi  and
      Xu, Runxin  and
      Wang, Peiyi  and
      Feng, Xiachong  and
      Kong, Lingpeng  and
      Liu, Qi",
    editor = "Ku, Lun-Wei  and
      Martins, Andre  and
      Srikumar, Vivek",
    booktitle = "Proceedings of the 62nd Annual Meeting of the Association for Computational Linguistics (Volume 1: Long Papers)",
    month = aug,
    year = "2024",
    address = "Bangkok, Thailand",
    publisher = "Association for Computational Linguistics",
    url = "https://aclanthology.org/2024.acl-long.775/",
    doi = "10.18653/v1/2024.acl-long.775",
    pages = "14369--14387"
}

@inproceedings{zhu2022tatdqa,
  title={Towards complex document understanding by discrete reasoning},
  author={Zhu, Fengbin and Lei, Wenqiang and Feng, Fuli and Wang, Chao and Zhang, Haozhou and Chua, Tat-Seng},
  booktitle={Proceedings of the 30th ACM International Conference on Multimedia},
  pages={4857--4866},
  year={2022},
  doi={10.1145/3503161.3548422},
  url={https://doi.org/10.1145/3503161.3548422}
}

@inproceedings{khattab2020colbert,
  title={Colbert: Efficient and effective passage search via contextualized late interaction over bert},
  author={Khattab, Omar and Zaharia, Matei},
  booktitle={Proceedings of the 43rd International ACM SIGIR conference on research and development in Information Retrieval},
  pages={39--48},
  year={2020},
  eprint={2004.12832},
  archivePrefix={arXiv},
  doi={10.1145/3397271.3401075},
  url={https://doi.org/10.1145/3397271.3401075}
}

@article{oord2018representation,
  title={Representation learning with contrastive predictive coding},
  author={Oord, Aaron van den and Li, Yazhe and Vinyals, Oriol},
  journal={arXiv preprint arXiv:1807.03748},
  year={2018}
}

@article{ma2025towards,
  title={Towards Storage-Efficient Visual Document Retrieval: An Empirical Study on Reducing Patch-Level Embeddings},
  author={Ma, Yubo and Li, Jinsong and Zang, Yuhang and Wu, Xiaobao and Dong, Xiaoyi and Zhang, Pan and Cao, Yuhang and Duan, Haodong and Wang, Jiaqi and Cao, Yixin and others},
  journal={arXiv preprint arXiv:2506.04997},
  year={2025}
}

@article{yan2025docpruner,
  title={Docpruner: A storage-efficient framework for multi-vector visual document retrieval via adaptive patch-level embedding pruning},
  author={Yan, Yibo and Xu, Guangwei and Zou, Xin and Liu, Shuliang and Kwok, James and Hu, Xuming},
  journal={arXiv preprint arXiv:2509.23883},
  year={2025}
}

@article{zhang2025deep,
  title={Deep research: A survey of autonomous research agents},
  author={Zhang, Wenlin and Li, Xiaopeng and Zhang, Yingyi and Jia, Pengyue and Wang, Yichao and Guo, Huifeng and Liu, Yong and Zhao, Xiangyu},
  journal={arXiv preprint arXiv:2508.12752},
  year={2025}
}

@article{huang2025deep,
  title={Deep research agents: A systematic examination and roadmap},
  author={Huang, Yuxuan and Chen, Yihang and Zhang, Haozheng and Li, Kang and Zhou, Huichi and Fang, Meng and Yang, Linyi and Li, Xiaoguang and Shang, Lifeng and Xu, Songcen and others},
  journal={arXiv preprint arXiv:2506.18096},
  year={2025}
}

@article{li2025webthinker,
  title={Webthinker: Empowering large reasoning models with deep research capability},
  author={Li, Xiaoxi and Jin, Jiajie and Dong, Guanting and Qian, Hongjin and Wu, Yongkang and Wen, Ji-Rong and Zhu, Yutao and Dou, Zhicheng},
  journal={arXiv preprint arXiv:2504.21776},
  year={2025}
}

@article{zheng2025deepresearcher,
  title={Deepresearcher: Scaling deep research via reinforcement learning in real-world environments},
  author={Zheng, Yuxiang and Fu, Dayuan and Hu, Xiangkun and Cai, Xiaojie and Ye, Lyumanshan and Lu, Pengrui and Liu, Pengfei},
  journal={arXiv preprint arXiv:2504.03160},
  year={2025}
}

@article{java2025characterizing,
  title={Characterizing deep research: A benchmark and formal definition},
  author={Java, Abhinav and Khandelwal, Ashmit and Midigeshi, Sukruta and Halfaker, Aaron and Deshpande, Amit and Goyal, Navin and Gupta, Ankur and Natarajan, Nagarajan and Sharma, Amit},
  journal={arXiv preprint arXiv:2508.04183},
  year={2025}
}

@article{chen2025browsecomp,
  title={Browsecomp-plus: A more fair and transparent evaluation benchmark of deep-research agent},
  author={Chen, Zijian and Ma, Xueguang and Zhuang, Shengyao and Nie, Ping and Zou, Kai and Liu, Andrew and Green, Joshua and Patel, Kshama and Meng, Ruoxi and Su, Mingyi and others},
  journal={arXiv preprint arXiv:2508.06600},
  year={2025}
}

@article{fang2025cognitive,
  title={Cognitive kernel-pro: A framework for deep research agents and agent foundation models training},
  author={Fang, Tianqing and Zhang, Zhisong and Wang, Xiaoyang and Wang, Rui and Qin, Can and Wan, Yuxuan and Ma, Jun-Yu and Zhang, Ce and Chen, Jiaqi and Li, Xiyun and others},
  journal={arXiv preprint arXiv:2508.00414},
  year={2025}
}

@article{li2025webweaver,
  title={Webweaver: Structuring web-scale evidence with dynamic outlines for open-ended deep research},
  author={Li, Zijian and Guan, Xin and Zhang, Bo and Huang, Shen and Zhou, Houquan and Lai, Shaopeng and Yan, Ming and Jiang, Yong and Xie, Pengjun and Huang, Fei and others},
  journal={arXiv preprint arXiv:2509.13312},
  year={2025}
}

@article{team2025tongyi,
  title={Tongyi deepresearch technical report},
  author={Team, Tongyi DeepResearch and Li, Baixuan and Zhang, Bo and Zhang, Dingchu and Huang, Fei and Li, Guangyu and Chen, Guoxin and Yin, Huifeng and Wu, Jialong and Zhou, Jingren and others},
  journal={arXiv preprint arXiv:2510.24701},
  year={2025}
}

@article{qiao2025webresearcher,
  title={Webresearcher: Unleashing unbounded reasoning capability in long-horizon agents},
  author={Qiao, Zile and Chen, Guoxin and Chen, Xuanzhong and Yu, Donglei and Yin, Wenbiao and Wang, Xinyu and Zhang, Zhen and Li, Baixuan and Yin, Huifeng and Li, Kuan and others},
  journal={arXiv preprint arXiv:2509.13309},
  year={2025}
}

@article{deng2025atom,
  title={Atom-searcher: Enhancing agentic deep research via fine-grained atomic thought reward},
  author={Deng, Yong and Wang, Guoqing and Ying, Zhenzhe and Wu, Xiaofeng and Lin, Jinzhen and Xiong, Wenwen and Dai, Yuqin and Yang, Shuo and Zhang, Zhanwei and Wang, Qiwen and others},
  journal={arXiv preprint arXiv:2508.12800},
  year={2025}
}

@article{yang2025multimodal,
  title={Multimodal deepresearcher: Generating text-chart interleaved reports from scratch with agentic framework},
  author={Yang, Zhaorui and Pan, Bo and Wang, Han and Wang, Yiyao and Liu, Xingyu and Weng, Luoxuan and Feng, Yingchaojie and Feng, Haozhe and Zhu, Minfeng and Zhang, Bo and others},
  journal={arXiv preprint arXiv:2506.02454},
  year={2025}
}

@article{li2025deepagent,
  title={DeepAgent: A General Reasoning Agent with Scalable Toolsets},
  author={Li, Xiaoxi and Jiao, Wenxiang and Jin, Jiarui and Dong, Guanting and Jin, Jiajie and Wang, Yinuo and Wang, Hao and Zhu, Yutao and Wen, Ji-Rong and Lu, Yuan and others},
  journal={arXiv preprint arXiv:2510.21618},
  year={2025}
}

@article{apicella2025don,
  title={Don't Push the Button! Exploring Data Leakage Risks in Machine Learning and Transfer Learning},
  author={Apicella, Andrea and Isgr{\`o}, Francesco and Prevete, Roberto},
  journal={Artificial Intelligence Review},
  volume={58},
  number={11},
  pages={339},
  year={2025},
  doi={10.1007/s10462-025-11326-3},
  url={https://doi.org/10.1007/s10462-025-11326-3},
  publisher={Springer}
}

@inproceedings{lilja2024localization,
  title={Localization is all you evaluate: Data leakage in online mapping datasets and how to fix it},
  author={Lilja, Adam and Fu, Junsheng and Stenborg, Erik and Hammarstrand, Lars},
  booktitle={Proceedings of the IEEE/CVF Conference on Computer Vision and Pattern Recognition},
  pages={22150--22159},
  year={2024},
  doi={10.1109/CVPR52733.2024.02091},
  url={https://doi.org/10.1109/CVPR52733.2024.02091}
}

@article{tang2024we,
  title={Do we need domain-specific embedding models? An empirical investigation},
  author={Tang, Yixuan and Yang, Yi},
  journal={arXiv preprint arXiv:2409.18511},
  year={2024}
}

@article{cao2024recent,
  title={Recent advances in text embedding: A Comprehensive Review of Top-Performing Methods on the MTEB Benchmark},
  author={Cao, Hongliu},
  journal={arXiv preprint arXiv:2406.01607},
  year={2024}
}

@article{chung2025maintaining,
  title={Maintaining MTEB: Towards Long Term Usability and Reproducibility of Embedding Benchmarks},
  author={Chung, Isaac and Kerboua, Imene and Kardos, Marton and Solomatin, Roman and Enevoldsen, Kenneth},
  journal={arXiv preprint arXiv:2506.21182},
  year={2025}
}

@article{song2024both,
  title={Both Text and Images Leaked! A Systematic Analysis of Data Contamination in Multimodal {LLM}},
  author={Song, Dingjie and Lai, Sicheng and Wang, Mingxuan and Chen, Shunian and Sun, Lichao and Wang, Benyou},
  journal={arXiv preprint arXiv:2411.03823},
  year={2024}
}

@article{xu2024benchmark,
  title={Benchmark data contamination of large language models: A survey},
  author={Xu, Cheng and Guan, Shuhao and Greene, Derek and Kechadi, M-Tahar},
  journal={arXiv preprint arXiv:2406.04244},
  year={2024}
}

@inproceedings{samuel2025towards,
  title={Towards data contamination detection for modern large language models: Limitations, inconsistencies, and oracle challenges},
  author={Samuel, Vinay and Zhou, Yue and Zou, Henry Peng},
  booktitle={Proceedings of the 31st International Conference on Computational Linguistics},
  pages={5058--5070},
  year={2025},
  eprint={2409.09927},
  archivePrefix={arXiv},
  url={https://aclanthology.org/2025.coling-main.338/}
}

@article{tsai2025let,
  title={Let LLMs Speak Embedding Languages: Generative Text Embeddings via Iterative Contrastive Refinement},
  author={Tsai, Yu-Che and Chen, Kuan-Yu and Li, Yuan-Chi and Chen, Yuan-Hao and Tsai, Ching-Yu and Lin, Shou-De},
  journal={arXiv preprint arXiv:2509.24291},
  year={2025}
}

@article{peng2024answer,
  title={Answer is all you need: Instruction-following text embedding via answering the question},
  author={Peng, Letian and Zhang, Yuwei and Wang, Zilong and Srinivasa, Jayanth and Liu, Gaowen and Wang, Zihan and Shang, Jingbo},
  journal={arXiv preprint arXiv:2402.09642},
  year={2024}
}

@inproceedings{muennighoff2024generative,
  title={Generative representational instruction tuning},
  author={Muennighoff, Niklas and Su, Hongjin and Wang, Liang and Yang, Nan and Wei, Furu and Yu, Tao and Singh, Amanpreet and Kiela, Douwe},
  booktitle={The Thirteenth International Conference on Learning Representations},
  year={2025},
  eprint={2402.09906},
  archivePrefix={arXiv},
  url={https://openreview.net/forum?id=BC4lIvfSzv}
}

@article{li2025unimoe1,
  title={Uni-moe: Scaling unified multimodal llms with mixture of experts},
  author={Li, Yunxin and Jiang, Shenyuan and Hu, Baotian and Wang, Longyue and Zhong, Wanqi and Luo, Wenhan and Ma, Lin and Zhang, Min},
  journal={IEEE Transactions on Pattern Analysis and Machine Intelligence},
  year={2025},
  doi={10.1109/TPAMI.2025.3532688},
  url={https://doi.org/10.1109/TPAMI.2025.3532688},
  publisher={IEEE}
}

@article{li2025unimoe2,
  title={Uni-MoE-2.0-Omni: Scaling Language-Centric Omnimodal Large Model with Advanced MoE, Training and Data},
  author={Li, Yunxin and Chen, Xinyu and Jiang, Shenyuan and Shi, Haoyuan and Liu, Zhenyu and Zhang, Xuanyu and Deng, Nanhao and Xu, Zhenran and Ma, Yicheng and Zhang, Meishan and others},
  journal={arXiv preprint arXiv:2511.12609},
  year={2025}
}

@article{lei2024m3,
  title={{M3-JEPA}: Multimodal Alignment via Multi-gate {MoE} based on the Joint-Embedding Predictive Architecture},
  author={Lei, Hongyang and Cheng, Xiaolong and Qin, Qi and Wang, Dan and Fan, Kun and Huang, Huazhen and Gu, Qingqing and Wu, Yetao and Jiang, Zhonglin and Chen, Yong and Ji, Luo},
  journal={arXiv preprint arXiv:2409.05929},
  year={2024}
}

@article{li2024your,
  title={Your mixture-of-experts llm is secretly an embedding model for free},
  author={Li, Ziyue and Zhou, Tianyi},
  journal={arXiv preprint arXiv:2410.10814},
  year={2024}
}

@article{mischler2024contextual,
  title={Contextual feature extraction hierarchies converge in large language models and the brain},
  author={Mischler, Gavin and Li, Yinghao Aaron and Bickel, Stephan and Mehta, Ashesh D and Mesgarani, Nima},
  journal={Nature Machine Intelligence},
  volume={6},
  number={12},
  pages={1467--1477},
  year={2024},
  doi={10.1038/s42256-024-00925-4},
  url={https://doi.org/10.1038/s42256-024-00925-4},
  publisher={Nature Publishing Group UK London}
}

@article{jeong2024llm,
  title={Llm-select: Feature selection with large language models},
  author={Jeong, Daniel P and Lipton, Zachary C and Ravikumar, Pradeep},
  journal={arXiv preprint arXiv:2407.02694},
  year={2024}
}

@inproceedings{deng2025following,
    title = "Following the Autoregressive Nature of {LLM} Embeddings via Compression and Alignment",
    author = "Deng, Jingcheng  and
      Jiang, Zhongtao  and
      Pang, Liang  and
      Wei, Zihao  and
      Chen, Liwei  and
      Xu, Kun  and
      Song, Yang  and
      Shen, Huawei  and
      Cheng, Xueqi",
    editor = "Christodoulopoulos, Christos  and
      Chakraborty, Tanmoy  and
      Rose, Carolyn  and
      Peng, Violet",
    booktitle = "Proceedings of the 2025 Conference on Empirical Methods in Natural Language Processing",
    month = nov,
    year = "2025",
    address = "Suzhou, China",
    publisher = "Association for Computational Linguistics",
    url = "https://aclanthology.org/2025.emnlp-main.639/",
    doi = "10.18653/v1/2025.emnlp-main.639",
    pages = "12661--12677",
    ISBN = "979-8-89176-332-6"
}

@article{xiong2024autoregressive,
  title={Autoregressive models in vision: A survey},
  author={Xiong, Jing and Liu, Gongye and Huang, Lun and Wu, Chengyue and Wu, Taiqiang and Mu, Yao and Yao, Yuan and Shen, Hui and Wan, Zhongwei and Huang, Jinfa and others},
  journal={arXiv preprint arXiv:2411.05902},
  year={2024}
}

@article{gan2025mixture,
  title={Mixture of experts (moe): A big data perspective},
  author={Gan, Wensheng and Ning, Zhenyao and Qi, Zhenlian and Yu, Philip S},
  journal={arXiv preprint arXiv:2501.16352},
  year={2025}
}

@article{fan2024towards,
  title={Towards an empirical understanding of moe design choices},
  author={Fan, Dongyang and Messmer, Bettina and Jaggi, Martin},
  journal={arXiv preprint arXiv:2402.13089},
  year={2024}
}

@article{Liu2026LookIT,
  title={Structural Anchor Pruning: Training-Free Multi-Vector Compression for Visual Document Retrieval},
  author={Liu, Zhuchenyang and Hu, Ziyu and Zhang, Yao and Xiao, Yu},
  journal={arXiv preprint arXiv:2601.20107},
  year={2026},
  url={https://arxiv.org/abs/2601.20107}
}

@inproceedings{lo2025closer,
    title = "A Closer Look into Mixture-of-Experts in Large Language Models",
    author = "Lo, Ka Man  and
      Huang, Zeyu  and
      Qiu, Zihan  and
      Wang, Zili  and
      Fu, Jie",
    editor = "Chiruzzo, Luis  and
      Ritter, Alan  and
      Wang, Lu",
    booktitle = "Findings of the Association for Computational Linguistics: NAACL 2025",
    month = apr,
    year = "2025",
    address = "Albuquerque, New Mexico",
    publisher = "Association for Computational Linguistics",
    url = "https://aclanthology.org/2025.findings-naacl.251/",
    doi = "10.18653/v1/2025.findings-naacl.251",
    pages = "4427--4447",
    ISBN = "979-8-89176-195-7"
}

@inproceedings{lassance2022learned,
  title={Learned token pruning in contextualized late interaction over BERT (ColBERT)},
  author={Lassance, Carlos and Maachou, Maroua and Park, Joohee and Clinchant, St{\'e}phane},
  booktitle={Proceedings of the 45th International ACM SIGIR Conference on Research and Development in Information Retrieval},
  pages={2232--2236},
  year={2022},
  doi={10.1145/3477495.3531835},
  url={https://doi.org/10.1145/3477495.3531835}
}

@article{lassance2021study,
  title={A study on token pruning for colbert},
  author={Lassance, Carlos and Maachou, Maroua and Park, Joohee and Clinchant, St{\'e}phane},
  journal={arXiv preprint arXiv:2112.06540},
  year={2021}
}

@article{liu2024analysis,
  title={An analysis on matching mechanisms and token pruning for late-interaction models},
  author={Liu, Qi and Guo, Gang and Mao, Jiaxin and Dou, Zhicheng and Wen, Ji-Rong and Jiang, Hao and Zhang, Xinyu and Cao, Zhao},
  journal={ACM Transactions on Information Systems},
  volume={42},
  number={5},
  pages={1--28},
  year={2024},
  doi={10.1145/3639818},
  url={https://doi.org/10.1145/3639818},
  publisher={ACM New York, NY}
}

@article{kusupati2022matryoshka,
  title={Matryoshka representation learning},
  author={Kusupati, Aditya and Bhatt, Gantavya and Rege, Aniket and Wallingford, Matthew and Sinha, Aditya and Ramanujan, Vivek and Howard-Snyder, William and Chen, Kaifeng and Kakade, Sham and Jain, Prateek and others},
  journal={Advances in Neural Information Processing Systems},
  volume={35},
  pages={30233--30249},
  year={2022},
  doi={10.52202/068431-2192},
  url={https://doi.org/10.52202/068431-2192}
}

@article{zhang2025qwen3,
  title={Qwen3 Embedding: Advancing Text Embedding and Reranking Through Foundation Models},
  author={Zhang, Yanzhao and Li, Mingxin and Long, Dingkun and Zhang, Xin and Lin, Huan and Yang, Baosong and Xie, Pengjun and Yang, An and Liu, Dayiheng and Lin, Junyang and others},
  journal={arXiv preprint arXiv:2506.05176},
  year={2025}
}

@article{zhao2025kalm,
  title={Kalm-embedding-v2: Superior training techniques and data inspire A versatile embedding model},
  author={Zhao, Xinping and Hu, Xinshuo and Shan, Zifei and Huang, Shouzheng and Zhou, Yao and Zhang, Xin and Sun, Zetian and Liu, Zhenyu and Li, Dongfang and Wei, Xinyuan and others},
  journal={arXiv preprint arXiv:2506.20923},
  year={2025}
}

@article{vera2025embeddinggemma,
  title={Embeddinggemma: Powerful and lightweight text representations},
  author={Vera, Henrique Schechter and Dua, Sahil and Zhang, Biao and Salz, Daniel and Mullins, Ryan and Panyam, Sindhu Raghuram and Smoot, Sara and Naim, Iftekhar and Zou, Joe and Chen, Feiyang and others},
  journal={arXiv preprint arXiv:2509.20354},
  year={2025}
}

@article{lee2025gemini,
  title={Gemini embedding: Generalizable embeddings from gemini},
  author={Lee, Jinhyuk and Chen, Feiyang and Dua, Sahil and Cer, Daniel and Shanbhogue, Madhuri and Naim, Iftekhar and {\'A}brego, Gustavo Hern{\'a}ndez and Li, Zhe and Chen, Kaifeng and Vera, Henrique Schechter and others},
  journal={arXiv preprint arXiv:2503.07891},
  year={2025}
}

@article{sturua2024jina,
  title={jina-embeddings-v3: Multilingual embeddings with task lora},
  author={Sturua, Saba and Mohr, Isabelle and Akram, Mohammad Kalim and G{\"u}nther, Michael and Wang, Bo and Krimmel, Markus and Wang, Feng and Mastrapas, Georgios and Koukounas, Andreas and Wang, Nan and others},
  journal={arXiv preprint arXiv:2409.10173},
  year={2024}
}

@article{jha2024jina,
  title={Jina-colbert-v2: A general-purpose multilingual late interaction retriever},
  author={Jha, Rohan and Wang, Bo and G{\"u}nther, Michael and Mastrapas, Georgios and Sturua, Saba and Mohr, Isabelle and Koukounas, Andreas and Akram, Mohammad Kalim and Wang, Nan and Xiao, Han},
  journal={arXiv preprint arXiv:2408.16672},
  year={2024}
}

@article{bussmann2025learning,
  title={Learning multi-level features with matryoshka sparse autoencoders},
  author={Bussmann, Bart and Nabeshima, Noa and Karvonen, Adam and Nanda, Neel},
  journal={arXiv preprint arXiv:2503.17547},
  year={2025}
}

@article{cai2024matryoshka,
  title={Matryoshka multimodal models},
  author={Cai, Mu and Yang, Jianwei and Gao, Jianfeng and Lee, Yong Jae},
  journal={arXiv preprint arXiv:2405.17430},
  year={2024}
}

@inproceedings{yoon2024matryoshka,
    title = "Matryoshka-Adaptor: Unsupervised and Supervised Tuning for Smaller Embedding Dimensions",
    author = "Yoon, Jinsung  and
      Sinha, Rajarishi  and
      Arik, Sercan O  and
      Pfister, Tomas",
    editor = "Al-Onaizan, Yaser  and
      Bansal, Mohit  and
      Chen, Yun-Nung",
    booktitle = "Proceedings of the 2024 Conference on Empirical Methods in Natural Language Processing",
    month = nov,
    year = "2024",
    address = "Miami, Florida, USA",
    publisher = "Association for Computational Linguistics",
    url = "https://aclanthology.org/2024.emnlp-main.576/",
    doi = "10.18653/v1/2024.emnlp-main.576",
    pages = "10318--10336"
}

@article{li20242d,
  title={2d matryoshka sentence embeddings},
  author={Li, Xianming and Li, Zongxi and Li, Jing and Xie, Haoran and Li, Qing},
  journal={arXiv preprint arXiv:2402.14776},
  year={2024}
}

@article{hu2024matryoshka,
  title={Matryoshka query transformer for large vision-language models},
  author={Hu, Wenbo and Dou, Zi-Yi and Li, Liunian Harold and Kamath, Amita and Peng, Nanyun and Chang, Kai-Wei},
  journal={Advances in Neural Information Processing Systems},
  volume={37},
  pages={50168--50188},
  year={2024},
  doi={10.52202/079017-1588},
  url={https://doi.org/10.52202/079017-1588}
}

@article{nair2025matryoshka,
  title={Matryoshka quantization},
  author={Nair, Pranav and Datta, Puranjay and Dean, Jeff and Jain, Prateek and Kusupati, Aditya},
  journal={arXiv preprint arXiv:2502.06786},
  year={2025}
}

@article{shukla2024matmamba,
  title={Matmamba: A matryoshka state space model},
  author={Shukla, Abhinav and Vemprala, Sai and Kusupati, Aditya and Kapoor, Ashish},
  journal={arXiv preprint arXiv:2410.06718},
  year={2024}
}

@article{zhuang2024starbucks,
  title={Starbucks-v2: Improved Training for 2D Matryoshka Embeddings},
  author={Zhuang, Shengyao and Wang, Shuai and Zheng, Fabio and Koopman, Bevan and Zuccon, Guido},
  journal={arXiv preprint arXiv:2410.13230},
  year={2024}
}

@article{liu2025matryoshka,
  title={Matryoshka re-ranker: A flexible re-ranking architecture with configurable depth and width},
  author={Liu, Zheng and Li, Chaofan and Xiao, Shitao and Li, Chaozhuo and Lian, Defu and Shao, Yingxia},
  journal={arXiv preprint arXiv:2501.16302},
  year={2025}
}

@inproceedings{zhang2025smec,
    title = "{SMEC}:Rethinking Matryoshka Representation Learning for Retrieval Embedding Compression",
    author = "Zhang, Biao  and
      Chen, Lixin  and
      Liu, Tong  and
      Zheng, Bo",
    editor = "Christodoulopoulos, Christos  and
      Chakraborty, Tanmoy  and
      Rose, Carolyn  and
      Peng, Violet",
    booktitle = "Proceedings of the 2025 Conference on Empirical Methods in Natural Language Processing",
    month = nov,
    year = "2025",
    address = "Suzhou, China",
    publisher = "Association for Computational Linguistics",
    url = "https://aclanthology.org/2025.emnlp-main.1332/",
    doi = "10.18653/v1/2025.emnlp-main.1332",
    pages = "26209--26222",
    ISBN = "979-8-89176-332-6"
}

@article{lai2024matryoshka,
  title={Matryoshka representation learning for recommendation},
  author={Lai, Riwei and Chen, Li and Chen, Weixin and Chen, Rui},
  journal={arXiv preprint arXiv:2406.07432},
  year={2024}
}

@article{wang2024train,
  title={Train once, deploy anywhere: Matryoshka representation learning for multimodal recommendation},
  author={Wang, Yueqi and Yue, Zhenrui and Zeng, Huimin and Wang, Dong and McAuley, Julian},
  journal={arXiv preprint arXiv:2409.16627},
  year={2024}
}

@incollection{nacar2025enhancing,
  title={Enhancing semantic similarity understanding in arabic nlp with nested embedding learning},
  author={Nacar, Omer and Koubaa, Anis},
  booktitle={Generative AI and Large Language Models: Opportunities, Challenges, and Applications: Volume 1},
  pages={179--216},
  year={2025},
  doi={10.1007/978-3-031-90573-5_6},
  url={https://doi.org/10.1007/978-3-031-90573-5_6},
  publisher={Springer}
}

@inproceedings{hanley2025hierarchical,
    title = "Hierarchical Level-Wise News Article Clustering via Multilingual Matryoshka Embeddings",
    author = "Hanley, Hans William Alexander  and
      Durumeric, Zakir",
    editor = "Che, Wanxiang  and
      Nabende, Joyce  and
      Shutova, Ekaterina  and
      Pilehvar, Mohammad Taher",
    booktitle = "Proceedings of the 63rd Annual Meeting of the Association for Computational Linguistics (Volume 1: Long Papers)",
    month = jul,
    year = "2025",
    address = "Vienna, Austria",
    publisher = "Association for Computational Linguistics",
    url = "https://aclanthology.org/2025.acl-long.124/",
    doi = "10.18653/v1/2025.acl-long.124",
    pages = "2476--2492",
    ISBN = "979-8-89176-251-0"
}

@inproceedings{verma2025matryoshka,
  title={Matryoshka Model Learning for Improved Elastic Student Models},
  author={Verma, Chetan and Timmaraju, Aditya Srinivas and Hsieh, Cho-Jui and Damle, Suyash and Bui, Ngot and Zhang, Yang and Chen, Wen and Liu, Xin and Jain, Prateek and Dhillon, Inderjit S.},
  booktitle={Proceedings of the 31st ACM SIGKDD Conference on Knowledge Discovery and Data Mining V. 2},
  pages={4935--4944},
  year={2025},
  doi={10.1145/3711896.3737245},
  url={https://doi.org/10.1145/3711896.3737245}
}

@article{fu2025moon,
  title={MOON Embedding: Multimodal Representation Learning for E-commerce Search Advertising},
  author={Fu, Chenghan and Zhang, Daoze and Lin, Yukang and Nie, Zhanheng and Zhang, Xiang and Liu, Jianyu and Liu, Yueran and Guan, Wanxian and Wang, Pengjie and Xu, Jian and others},
  journal={arXiv preprint arXiv:2511.11305},
  year={2025}
}

@inproceedings{wang2024improving,
    title = "Improving Text Embeddings with Large Language Models",
    author = "Wang, Liang  and
      Yang, Nan  and
      Huang, Xiaolong  and
      Yang, Linjun  and
      Majumder, Rangan  and
      Wei, Furu",
    editor = "Ku, Lun-Wei  and
      Martins, Andre  and
      Srikumar, Vivek",
    booktitle = "Proceedings of the 62nd Annual Meeting of the Association for Computational Linguistics (Volume 1: Long Papers)",
    month = aug,
    year = "2024",
    address = "Bangkok, Thailand",
    publisher = "Association for Computational Linguistics",
    url = "https://aclanthology.org/2024.acl-long.642/",
    doi = "10.18653/v1/2024.acl-long.642",
    pages = "11897--11916"
}

@inproceedings{fang2024scaling,
  title={Scaling laws for dense retrieval},
  author={Fang, Yan and Zhan, Jingtao and Ai, Qingyao and Mao, Jiaxin and Su, Weihang and Chen, Jia and Liu, Yiqun},
  booktitle={Proceedings of the 47th International ACM SIGIR Conference on Research and Development in Information Retrieval},
  pages={1339--1349},
  year={2024},
  doi={10.1145/3626772.3657743},
  url={https://doi.org/10.1145/3626772.3657743}
}

@inproceedings{jiang2024scaling,
    title = "Scaling Sentence Embeddings with Large Language Models",
    author = "Jiang, Ting  and
      Huang, Shaohan  and
      Luan, Zhongzhi  and
      Wang, Deqing  and
      Zhuang, Fuzhen",
    editor = "Al-Onaizan, Yaser  and
      Bansal, Mohit  and
      Chen, Yun-Nung",
    booktitle = "Findings of the Association for Computational Linguistics: EMNLP 2024",
    month = nov,
    year = "2024",
    address = "Miami, Florida, USA",
    publisher = "Association for Computational Linguistics",
    url = "https://aclanthology.org/2024.findings-emnlp.181/",
    doi = "10.18653/v1/2024.findings-emnlp.181",
    pages = "3182--3196"
}

@article{alabdulmohsin2022revisiting,
  title={Revisiting neural scaling laws in language and vision},
  author={Alabdulmohsin, Ibrahim M and Neyshabur, Behnam and Zhai, Xiaohua},
  journal={Advances in Neural Information Processing Systems},
  volume={35},
  pages={22300--22312},
  year={2022},
  doi={10.52202/068431-1620},
  url={https://doi.org/10.52202/068431-1620}
}

@article{zhang2024scaling,
  title={When scaling meets llm finetuning: The effect of data, model and finetuning method},
  author={Zhang, Biao and Liu, Zhongtao and Cherry, Colin and Firat, Orhan},
  journal={arXiv preprint arXiv:2402.17193},
  year={2024}
}

@article{xiao2025densing,
  title={Densing law of llms},
  author={Xiao, Chaojun and Cai, Jie and Zhao, Weilin and Lin, Biyuan and Zeng, Guoyang and Zhou, Jie and Zheng, Zhi and Han, Xu and Liu, Zhiyuan and Sun, Maosong},
  journal={Nature Machine Intelligence},
  volume={7},
  number={11},
  pages={1823--1833},
  year={2025},
  doi={10.1038/s42256-025-01137-0},
  url={https://doi.org/10.1038/s42256-025-01137-0},
  publisher={Nature Publishing Group UK London}
}

@article{pearce2024scaling,
  title={Scaling laws for pre-training agents and world models},
  author={Pearce, Tim and Rashid, Tabish and Bignell, Dave and Georgescu, Raluca and Devlin, Sam and Hofmann, Katja},
  journal={arXiv preprint arXiv:2411.04434},
  year={2024}
}

@article{beyer2024paligemma,
  title={Paligemma: A versatile 3b vlm for transfer},
  author={Beyer, Lucas and Steiner, Andreas and Pinto, Andr{\'e} Susano and Kolesnikov, Alexander and Wang, Xiao and Salz, Daniel and Neumann, Maxim and Alabdulmohsin, Ibrahim and Tschannen, Michael and Bugliarello, Emanuele and others},
  journal={arXiv preprint arXiv:2407.07726},
  year={2024}
}

@article{bai2025qwen2,
  title={Qwen2. 5-vl technical report},
  author={Bai, Shuai and Chen, Keqin and Liu, Xuejing and Wang, Jialin and Ge, Wenbin and Song, Sibo and Dang, Kai and Wang, Peng and Wang, Shijie and Tang, Jun and others},
  journal={arXiv preprint arXiv:2502.13923},
  year={2025}
}

@misc{laurençon2024building,
      title={Building and better understanding vision-language models: insights and future directions.}, 
      author={Hugo Laurençon and Andrés Marafioti and Victor Sanh and Léo Tronchon},
      year={2024},
      eprint={2408.12637},
      archivePrefix={arXiv},
      primaryClass={cs.CV}
}

@article{huo2026causalembed,
  title={CausalEmbed: Auto-Regressive Multi-Vector Generation in Latent Space for Visual Document Embedding},
  author={Huo, Jiahao and Huang, Yu and Yan, Yibo and Pan, Ye and Zheng, Kening and Huang, Wei-Chieh and Cao, Yi and Ou, Mingdong and Yu, Philip S. and Hu, Xuming},
  journal={arXiv preprint arXiv:2601.21262},
  year={2026}
}

@article{li2026qwen3vlembed,
  title={Qwen3-VL-Embedding and Qwen3-VL-Reranker: A Unified Framework for State-of-the-Art Multimodal Retrieval and Ranking},
  author={Li, Mingxin and Zhang, Yanzhao and Long, Dingkun and Chen, Keqin and Song, Sibo and Bai, Shuai and Yang, Zhibo and Xie, Pengjun and Yang, An and Liu, Dayiheng and others},
  journal={arXiv preprint arXiv:2601.04720},
  year={2026}
}

@article{cha2026reinpool,
  title={ReinPool: Reinforcement Learning Pooling Multi-Vector Embeddings for Retrieval System},
  author={Cha, Sungguk and Kim, DongWook and Kim, Mintae and Han, Youngsub and Jeon, Byoung-Ki and Lee, Sangyeob},
  journal={arXiv preprint arXiv:2601.07125},
  year={2026}
}

@article{pony2026colbandit,
  title={Col-Bandit: Query-Time Top-$K$ Estimation for Late-Interaction Retrieval},
  author={Pony, Roi and Goldfarb, Adi Raz and Naparstek, Oshri and Friedman, Idan and Barzelay, Udi and Schwartz, Eli},
  journal={arXiv preprint arXiv:2602.02827},
  year={2026}
}

@article{yan2026sculpting,
  title={Sculpting the Vector Space: Towards Efficient Multi-Vector Visual Document Retrieval via Prune-then-Merge Framework},
  author={Yan, Yibo and Ou, Mingdong and Cao, Yi and Zou, Xin and Huo, Jiahao and Liu, Shuliang and Kwok, James and Hu, Xuming},
  journal={arXiv preprint arXiv:2602.19549},
  year={2026}
}

@article{guo2026csrv2,
  title={CSRv2: Unlocking Ultra-Sparse Embeddings},
  author={Guo, Lixuan and Wang, Yifei and Wen, Tiansheng and Wang, Yifan and Feng, Aosong and Chen, Bo and Jegelka, Stefanie and You, Chenyu},
  journal={arXiv preprint arXiv:2602.05735},
  year={2026}
}

@article{wen2025beyond,
  title={Beyond matryoshka: Revisiting sparse coding for adaptive representation},
  author={Wen, Tiansheng and Wang, Yifei and Zeng, Zequn and Peng, Zhong and Su, Yudi and Liu, Xinyang and Chen, Bo and Liu, Hongwei and Jegelka, Stefanie and You, Chenyu},
  journal={arXiv preprint arXiv:2503.01776},
  year={2025}
}

@article{zhang2026llms,
  title={LLMs Meet Isolation Kernel: Lightweight, Learning-free Binary Embeddings for Fast Retrieval},
  author={Zhang, Zhibo and Xu, Yang and Ting, Kai Ming and Nguyen, Cam-Tu},
  journal={arXiv preprint arXiv:2601.09159},
  year={2026}
}

@article{qin2026multi,
  title={Multi-Vector Index Compression in Any Modality},
  author={Qin, Hanxiang and Martin, Alexander and Jha, Rohan and Zuo, Chunsheng and Kriz, Reno and Van Durme, Benjamin},
  journal={arXiv preprint arXiv:2602.21202},
  year={2026}
}

@article{jiang2026embedrl,
  title={Embed-RL: Reinforcement Learning for Reasoning-Driven Multimodal Embeddings},
  author={Jiang, Haonan and Wang, Yuji and Zhu, Yongjie and Lu, Xin and Qin, Wenyu and Wang, Meng and Wan, Pengfei and Tang, Yansong},
  journal={arXiv preprint arXiv:2602.13823},
  year={2026}
}

@article{moreira2026nemotron,
  title={Nemotron ColEmbed V2: Top-Performing Late Interaction embedding models for Visual Document Retrieval},
  author={Moreira, Gabriel de Souza P and Ak, Ronay and Xu, Mengyao and Holworthy, Oliver and Schifferer, Benedikt and Yu, Zhiding and Babakhin, Yauhen and Osmulski, Radek and Cai, Jiarui and Chesler, Ryan and others},
  journal={arXiv preprint arXiv:2602.03992},
  year={2026}
}

@inproceedings{zhang2025roles,
    title = "Roles of {MLLM}s in Visually Rich Document Retrieval for {RAG}: A Survey",
    author = "Zhang, Xiantao",
    editor = "Inui, Kentaro  and
      Sakti, Sakriani  and
      Wang, Haofen  and
      Wong, Derek F.  and
      Bhattacharyya, Pushpak  and
      Banerjee, Biplab  and
      Ekbal, Asif  and
      Chakraborty, Tanmoy  and
      Singh, Dhirendra Pratap",
    booktitle = "Proceedings of the 14th International Joint Conference on Natural Language Processing and the 4th Conference of the Asia-Pacific Chapter of the Association for Computational Linguistics",
    month = dec,
    year = "2025",
    address = "Mumbai, India",
    publisher = "The Asian Federation of Natural Language Processing and The Association for Computational Linguistics",
    url = "https://aclanthology.org/2025.ijcnlp-long.2/",
    doi = "10.18653/v1/2025.ijcnlp-long.2",
    pages = "19--36",
    ISBN = "979-8-89176-298-5"
}

@article{Cai2026WhenVM,
  title={When Vision Meets Texts in Listwise Reranking},
  author={Cai, Hongyi},
  journal={arXiv preprint arXiv:2601.20623},
  year={2026}
}

@article{cui2025reason,
  title={Reason to Contrast: A Cascaded Multimodal Retrieval Framework},
  author={Cui, Xuanming and Chen, Hong-You and Yu, Hao and Yuan, Hao and Wang, Zihao and Mishra, Shlok Kumar and Yu, Hanchao and Yang, Yonghuan and Xiao, Jun and Lim, Ser-Nam and others},
  journal={arXiv preprint arXiv:2602.23369},
  year={2025}
}

@article{yan2026beyond,
  title={Beyond the Grid: Layout-Informed Multi-Vector Retrieval with Parsed Visual Document Representations},
  author={Yan, Yibo and Ou, Mingdong and Cao, Yi and Zou, Xin and Liu, Shuliang and Huo, Jiahao and Huang, Yu and Kwok, James and Hu, Xuming},
  journal={arXiv preprint arXiv:2603.01666},
  year={2026}
}

@article{wu2026tsembed,
  title={TSEmbed: Unlocking Task Scaling in Universal Multimodal Embeddings},
  author={Wu, Yebo and Liu, Feng and Xie, Ziwei and Liu, Zhiyuan and Zhang, Changwang and Wang, Jun and Li, Li},
  journal={arXiv preprint arXiv:2603.04772},
  year={2026}
}

@article{liu2026nanovdr,
  title={NanoVDR: Distilling a 2B Vision-Language Retriever into a 70M Text-Only Encoder for Visual Document Retrieval},
  author={Liu, Zhuchenyang and Zhang, Yao and Xiao, Yu},
  journal={arXiv preprint arXiv:2603.12824},
  year={2026}
}

@article{zhu2026mure,
  title={MURE: Hierarchical Multi-Resolution Encoding via Vision-Language Models for Visual Document Retrieval},
  author={Zhu, Fengbin and Cai, Zijing and Wang, Yuzhe and Shao, Pengyang and Wang, Wenjie and Feng, Fuli and Hong, Richang and Chua, Tat-Seng},
  journal={arXiv preprint arXiv:2603.13349},
  year={2026}
}

@article{yan2026colchunk,
  title={Visual Late Chunking: An Empirical Study of Contextual Chunking for Efficient Visual Document Retrieval},
  author={Yan, Yibo and Ou, Mingdong and Cao, Yi and Huo, Jiahao and Zou, Xin and Liu, Shuliang and Kwok, James and Hu, Xuming},
  journal={arXiv preprint arXiv:2604.10167},
  year={2026}
}

@article{Wang2026MMEmbR1RM,
  title={MMEmb-R1: Reasoning-Enhanced Multimodal Embedding with Pair-Aware Selection and Adaptive Control},
  author={Wang, Yuchi and Yang, Dingkang and Yu, Haiyang and Bian, Weikang and Long, Jiefeng and Liang, Xiao and Feng, Chao and Li, Hongsheng},
  year={2026},
  journal={arXiv preprint arXiv:2604.06156},
  url={https://arxiv.org/abs/2604.06156}
}

@article{He2026PLUMELR,
  title={PLUME: Latent Reasoning Based Universal Multimodal Embedding},
  author={He, Chenwei and Hao, Xiangzhao and Yang, Tianyu and Ma, Yuxiang and Jia, Yuheng and Wu, Lingxiang and Zhao, Chaoyang and Guo, Haiyun and Wang, Jinqiao},
  year={2026},
  journal={arXiv preprint arXiv:2604.02073}
}

@article{Li2025MIRAGERS,
  title={MIRAGE: Runtime Scheduling for Multi-Vector Image Retrieval with Hierarchical Decomposition},
  author={Li, Maoliang and Li, Ke and Liu, Yaoyang and Chen, Jiayu and Zheng, Zihao and Wu, Yinjun and Liu, Chenchen and Chen, Xiang},
  year={2025},
  journal={arXiv preprint arXiv:2510.08976},
  url={https://arxiv.org/abs/2510.08976}
}

@article{Yang2026UnifiedAE,
  title={Unified and Efficient Approach for Multi-Vector Similarity Search},
  author={Yang, Binhan and Zeng, Yuxiang and Zhang, Hengxin and Zheng, Zhuanglin and Chi, Yunzhen and Tong, Yongxin and Xu, Ke},
  year={2026},
  journal={arXiv preprint arXiv:2604.02815},
  url={https://arxiv.org/abs/2604.02815}
}

@article{Chaffin2026ColBERTZeroTP,
  title={ColBERT-Zero: To Pre-train Or Not To Pre-train ColBERT models},
  author={Chaffin, Antoine and Arnaboldi, Luca and Chatelain, Am{\'e}lie and Krzakala, Florent},
  journal={arXiv preprint arXiv:2602.16609},
  year={2026},
  url={https://arxiv.org/abs/2602.16609}
}

@inproceedings{kanagalingam2026invoice,
  title={Invoice Haystack: Benchmarking Document Retrieval and Visual Question Answering Under Strong Visual Homogeneity},
  author={Kanagalingam, Heethanjan and Pathmanathan, Thenukan and Vathanakumar, Mokeeshan and Azam, Basim and Erfani, Sarah Monazam and Akhtar, Naveed},
  booktitle={Proceedings of the European Conference on Computer Vision (ECCV)},
  year={2026},
  note={arXiv:2606.25343},
  url={https://arxiv.org/abs/2606.25343}
}

@article{hu2026visdocagentbench,
  title={{VisDocAgentBench}: Benchmarking Agents for Visually Rich Document Retrieval},
  author={Hu, Lexiang and Zhang, Yanzhao and Li, Mingxin and Long, Dingkun and Li, Yikang and Zhang, Fuwei and Wang, Yisen and Lin, Zhouchen},
  journal={arXiv preprint arXiv:2608.17889},
  year={2026},
  url={https://arxiv.org/abs/2608.17889}
}

@article{choi2026kovidore,
  title={{KoViDoRe}: Korean Visual Document Retrieval},
  author={Choi, Yongbin and Song, Yongwoo and Sung, Mujeen},
  journal={arXiv preprint arXiv:2608.20840},
  year={2026},
  url={https://arxiv.org/abs/2608.20840}
}

@article{abdallah2026argus,
  title={Argus-Retriever: {Vision-LLM} Late-Interaction Retrieval with Region-Aware Query-Conditioned {MoE} for Visual Document Retrieval},
  author={Abdallah, Abdelrahman and Abdalla, Mahmoud and Ali, Mohammed and Jatowt, Adam},
  journal={arXiv preprint arXiv:2606.04300},
  year={2026},
  url={https://arxiv.org/abs/2606.04300}
}

@article{xiang2026mmmatryoshka,
  title={{MM-Matryoshka}: Towards Budget-Elastic Visual Document Retrieval via a 2D Multimodal Matryoshka Training Framework},
  author={Xiang, Haowen and Yan, Yibo and Huo, Jiahao and Huang, Yu and Cao, Yi and Ou, Mingdong and Hu, Xuming},
  journal={arXiv preprint arXiv:2606.07654},
  year={2026},
  url={https://arxiv.org/abs/2606.07654}
}

@article{wang2026varsdoc,
  title={{VaRS-Doc}: Interpretation-Aware Variant Representations via Latent Self-Probing for Visual Document Retrieval},
  author={Wang, Haocheng and Guan, Tongkun and Shen, Wei and Yang, Xiaokang},
  journal={arXiv preprint arXiv:2608.01211},
  year={2026},
  url={https://arxiv.org/abs/2608.01211}
}

@inproceedings{sun2026ziprerank,
  title={Very Efficient Listwise Multimodal Reranking for Long Documents},
  author={Sun, Yiqun and Wei, Pengfei and Hsieh, Lawrence B.},
  booktitle={Proceedings of the International Conference on Machine Learning (ICML)},
  year={2026},
  note={arXiv:2605.11864},
  url={https://arxiv.org/abs/2605.11864}
}

@article{fan2026minireranker,
  title={miniReranker: Efficient Multimodal Reranking through Visual Cache Reuse and Interaction Sparsity},
  author={Fan, Yingqi and Lu, Xuan and Zhao, Anhao and Tong, Junlong and Nie, Ping and Zou, Kai and Ma, Yunpu and Zhang, Wei and Shen, Xiaoyu},
  journal={arXiv preprint arXiv:2606.10759},
  year={2026},
  url={https://arxiv.org/abs/2606.10759}
}

@inproceedings{hu2026docretriever,
  title={{DocRetriever}: A Plug-and-Play Framework for Multimodal Document Retrieval with Comprehensive Benchmark},
  author={Hu, Ruofan and Zhu, Menghui and Zhu, Jieming and Chen, Bo and Xu, Shengyang and Hong, Minjie and Yang, Xiaoda and Zhou, Sashuai and Tang, Li and Jin, Tao and Zhao, Zhou},
  booktitle={Proceedings of the 32nd ACM SIGKDD Conference on Knowledge Discovery and Data Mining (KDD)},
  year={2026},
  note={arXiv:2605.30027},
  url={https://arxiv.org/abs/2605.30027}
}

@article{tilli2026beyond,
  title={Beyond Bag-of-Patches: Learning Global Layout via Textual Supervision for Late-Interaction Visual Document Retrieval},
  author={Tilli, Pascal and Mesgar, Mohsen},
  journal={arXiv preprint arXiv:2605.08421},
  year={2026},
  url={https://arxiv.org/abs/2605.08421}
}

@article{liu2026distilvdr,
  title={{DistilVDR}: A Compact End-to-End Visual Document Retriever via Dual-Student Distillation},
  author={Liu, Zhuchenyang and Wang, Ziyi and Zhang, Yao and Xiao, Yu},
  journal={arXiv preprint arXiv:2608.10636},
  year={2026},
  url={https://arxiv.org/abs/2608.10636}
}

@article{choi2026kovre,
  title={{KoVRE}: Training an Efficient Embedding Model for Korean Visual Document Retrieval},
  author={Choi, Yongbin and Shim, Gyuho and Jang, Youngjoon},
  journal={arXiv preprint arXiv:2608.01389},
  year={2026},
  url={https://arxiv.org/abs/2608.01389}
}

@article{josef2026learn2pool,
  title={Learn to Pool: Lightweight Fine-Tuning for Flexible Multi-Vector Compression},
  author={Josef, Stefan},
  journal={arXiv preprint arXiv:2607.06036},
  year={2026},
  url={https://arxiv.org/abs/2607.06036}
}

@article{mahdizadeh2026marginmerge,
  title={Coverage Matters: {MarginMerge} for Compressing Multi-Vector Visual Document Retrievers},
  author={Mahdizadeh, Ailar and Salari, Aria and Rajabi, Sohail and Mirabbasi, Shahriar and Nasiopoulos, Panos and Morsali, Alireza},
  journal={arXiv preprint arXiv:2608.02969},
  year={2026},
  url={https://arxiv.org/abs/2608.02969}
}

@article{zuo2026anchorfold,
  title={{AnchorFold}: A Focus-Then-Fold Framework via Recursive Attention Propagation for Efficient Multi-Vector Visual Document Retrieval},
  author={Zuo, Haoyu and Yan, Yibo and Zou, Xin and Liu, Shuliang and Cao, Yi and Ou, Mingdong and Hu, Xuming},
  journal={arXiv preprint arXiv:2608.08732},
  year={2026},
  url={https://arxiv.org/abs/2608.08732}
}

@article{salla2026crossq,
  title={{CrossQ}: Task-Aligned Cross-Token Conditional Quantization for Late Interaction Retrieval},
  author={Salla, Rohit Kumar and Saravanan, Manoj and Amancherla, Ramya Manasa},
  journal={arXiv preprint arXiv:2608.19204},
  year={2026},
  url={https://arxiv.org/abs/2608.19204}
}
